\documentclass[9pt,twocolumn,letterpaper]{article}
\usepackage[a4paper,left=2cm,right=2cm,top=2cm,bottom=2cm]{geometry}
\usepackage{dsfont}
\usepackage[english]{babel}
\usepackage{amsmath,amsfonts}
\usepackage{mathtools}
\usepackage{graphicx}
\usepackage[table, svgnames, dvipsnames]{xcolor}
\usepackage{caption}
\usepackage{subcaption}
\usepackage{longtable}
\usepackage{booktabs}
\usepackage{multirow}
\usepackage{pifont}\newcommand{\xmark}{\ding{55}}
\usepackage{hyperref}
\usepackage{cleveref}
\usepackage{rotating}
\usepackage{balance}
\usepackage{url}
\usepackage{ragged2e}

\DeclarePairedDelimiter\abs{\lvert}{\rvert}
\usepackage{mathtools}

\DeclarePairedDelimiter\floor{\lfloor}{\rfloor}
\def\method{DiMoDif}
\date{}

 \title{\method: 
 \underline{Di}scourse \underline{Mo}dality-information \underline{Dif}ferentiation
 for Audio-visual Deepfake Detection and Localization}

\author{Christos Koutlis,
Symeon Papadopoulos\\
Information Technologies Institute @ CERTH\\
Thessaloniki, Greece\\
{\texttt{\{ckoutlis,papadop\}@iti.gr}}
}

\begin{document}
\maketitle
\begin{abstract}
Deepfake technology has rapidly advanced and poses significant threats to information integrity and trust in online multimedia. While significant progress has been made in detecting deepfakes, the simultaneous manipulation of audio and visual modalities, sometimes at small parts or in subtle ways, presents highly challenging detection scenarios. To address these challenges, we present \method\ (\cref{fig:concept}), an audio-visual deepfake detection framework that leverages the \emph{inter-modality differences in machine perception of speech}, based on the assumption that in real samples --  in contrast to deepfakes -- visual and audio signals coincide in terms of information. \method\ leverages features from deep networks that specialize in visual and audio speech recognition to spot frame-level cross-modal incongruities, and in that way to temporally localize the deepfake forgery. To this end, we devise a hierarchical cross-modal fusion network, integrating adaptive temporal alignment modules and a learned discrepancy mapping layer to explicitly model the subtle differences between visual and audio representations. Then, the detection model is optimized through a composite loss function accounting for frame-level detections and fake intervals localization. \method\ outperforms the state-of-the-art on the Deepfake Detection task by 30.5 AUC on the highly challenging AV-Deepfake1M, while it performs exceptionally on FakeAVCeleb and LAV-DF. On the Temporal Forgery Localization task, it outperforms the state-of-the-art by 47.88 AP@0.75 on AV-Deepfake1M, and performs on-par on LAV-DF. Code available at \url{https://github.com/mever-team/dimodif}.
\end{abstract}

\section{Introduction}\label{sec:intro}
\begin{figure}
     \centering
     \includegraphics[trim={1.25cm 19.55cm 27.7cm 0.55cm},clip,width=\linewidth]{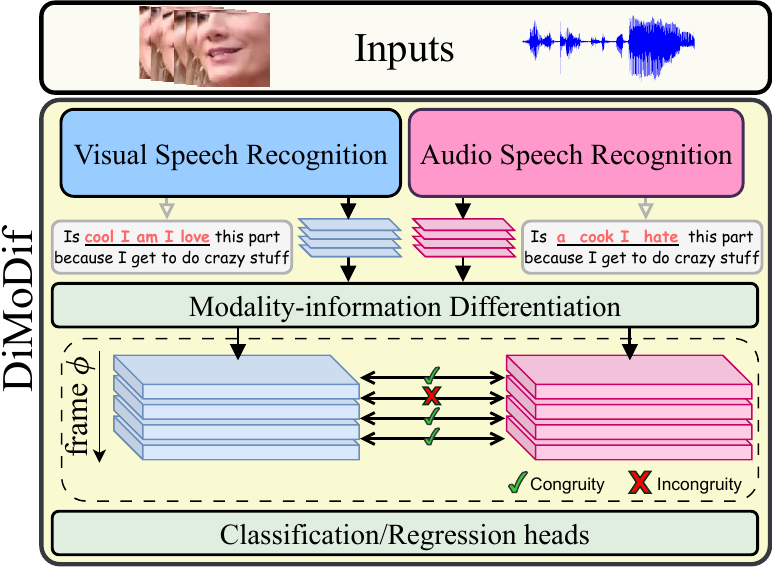}
     \caption{Partial audio-visual manipulation leads to different visual and audio speech predictions. \method\ detects and localizes the fake part based on feature space incongruity.}
     \label{fig:concept}
\end{figure}
Audio-visual deepfakes are AI-generated multimedia involving manipulations to one or both modalities, sometimes in small parts, with the intention to deceive \cite{masood2023deepfakes}. Deepfake content can spread widely online, and mislead viewers, contributing to the growing issue of information disorder. A revealing feature of audio-visual deepfakes is the presence of incongruities between the visual and audio signals, which are, however, increasingly hard to detect due to the quality and realism of current deepfakes. This necessitates the development of robust AI-based deepfake detection methods.

Deepfake detection is a growing research area \cite{pei2024deepfake}. Pertinent methods focus on pixel-level (gradient \cite{tan2023learning}, color \cite{he2019detection}, and artifact analysis \cite{cao2022end}) and feature-based (facial landmarks \cite{yang2019exposing}, temporal coherence \cite{gu2022delving}) inconsistencies.
An expanding area of study involves the detection and localization of deepfakes in audio-visual data, which is especially hard due to the way sound and visuals are intertwined. Key approaches analyze the consistency between raw visual and auditory cues \cite{gu2021deepfake,cai2023glitch,chugh2020not}, consider feature reconstruction learning modules \cite{zhang2023ummaformer}, and follow self-supervised approaches learning synchronization patterns from real videos only \cite{haliassos2022leveraging,cozzolino2023audio}.

\begin{figure*}[!ht]
    \centering
    \begin{subfigure}[b]{0.9\textwidth}
        \centering
        \includegraphics[trim={0cm 21cm 11cm 1cm},clip,width=\textwidth]{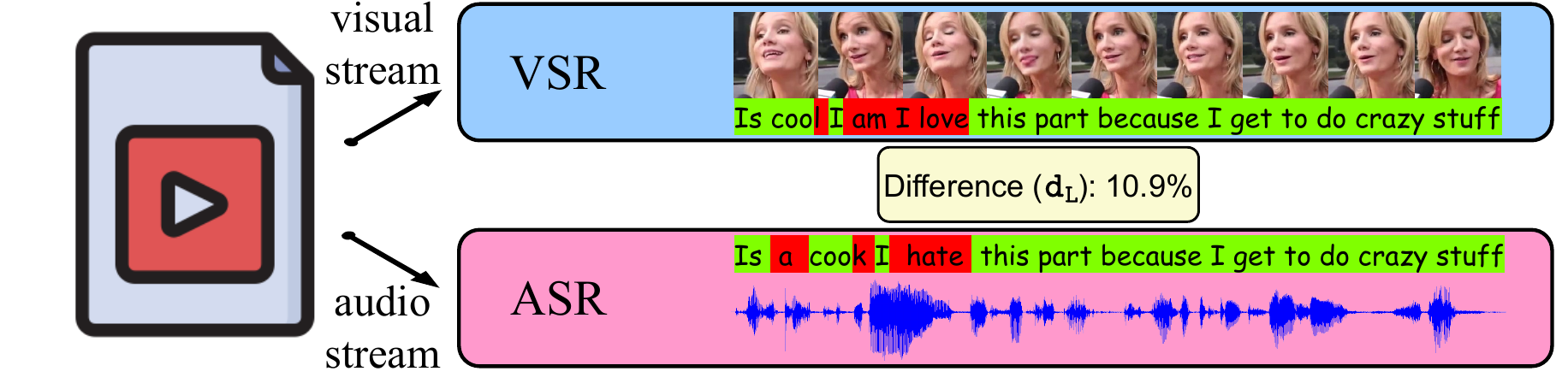}
        \caption{Example of differences between Visual Speech Recognition (VSR) and Audio Speech Recognition (ASR) outputs. \\
        File path: \texttt{AV-Deepfake1M/val/id01003/9nmM17wsxGU/00013/fake\_video\_fake\_audio.mp4}}\label{subfig:motivation_example}
    \end{subfigure}
    \begin{subfigure}[b]{0.32\textwidth}
        \centering
        \includegraphics[width=\textwidth]{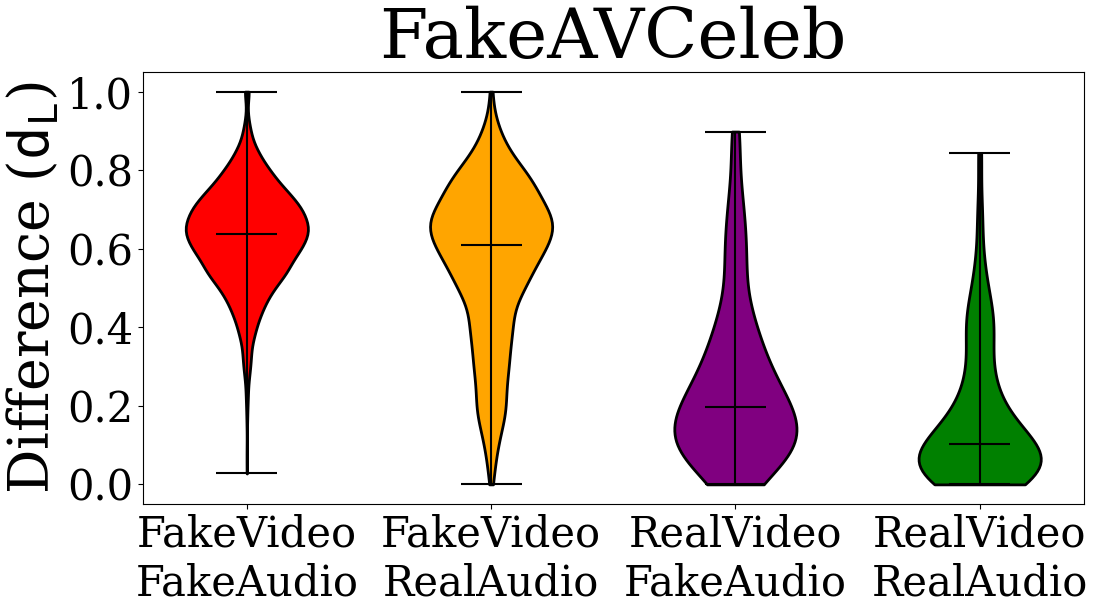}
        \caption{}\label{subfig:violin_favc}
    \end{subfigure}
    \hfill
    \begin{subfigure}[b]{0.32\textwidth}
        \centering
        \includegraphics[width=\textwidth]{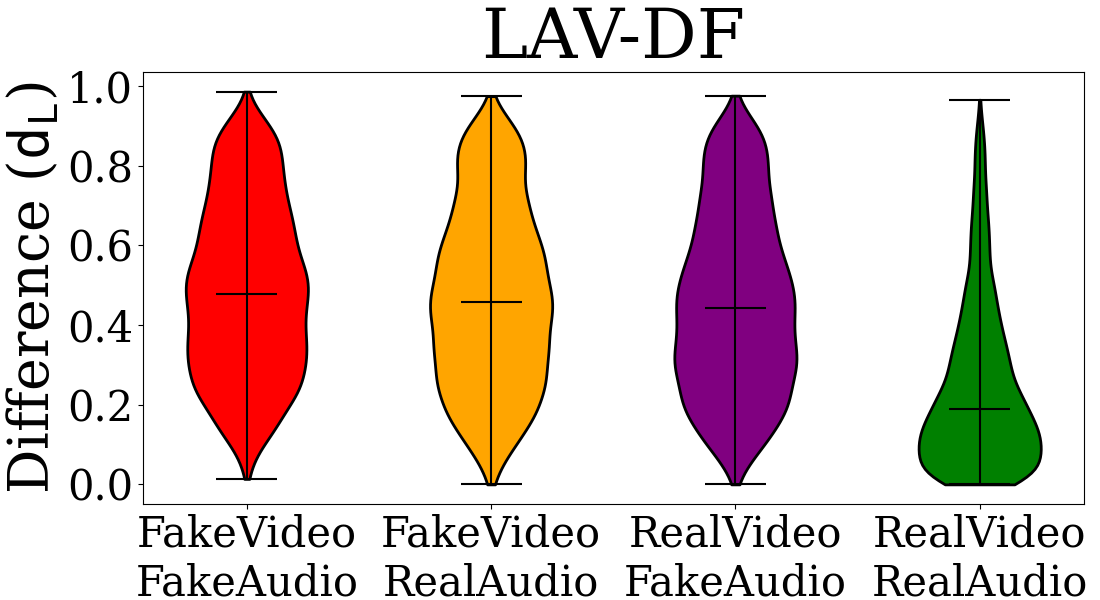}
        \caption{}\label{subfig:violin_lavdf}
    \end{subfigure}
    \hfill
    \begin{subfigure}[b]{0.32\textwidth}
        \centering
        \includegraphics[width=\textwidth]{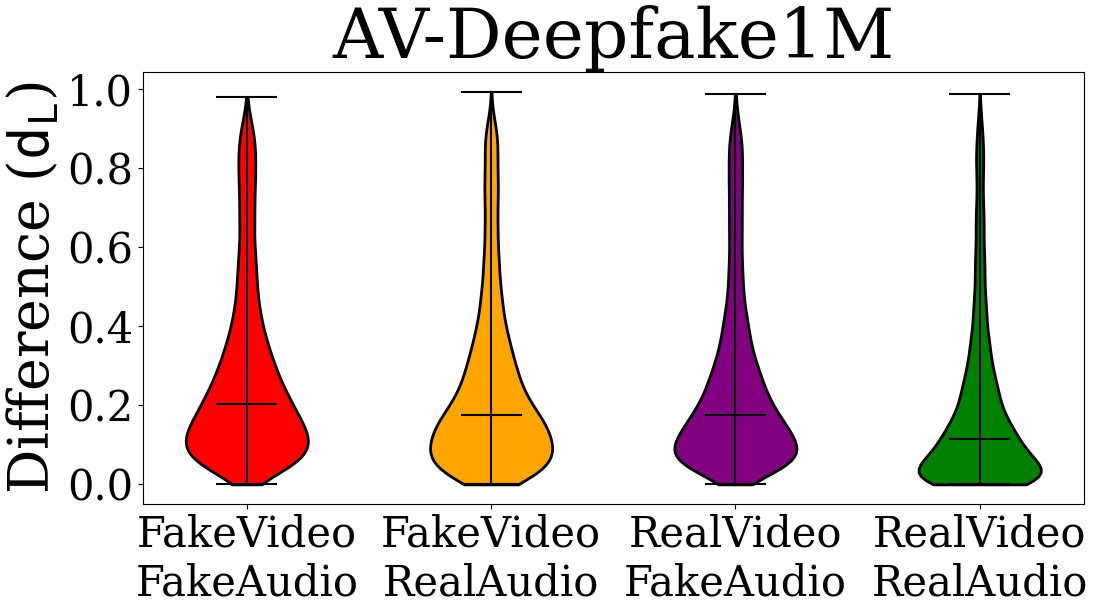}
        \caption{}\label{subfig:violin_avdf1m}
    \end{subfigure}
    \caption{Identifying machine perception discrepancies between visual and audio speech for deepfake detection. In (a), a video's visual and audio streams are separately processed by VSR and ASR models, then the outputs' normalized Levenshtein distance $\mathtt{d_L}$ is calculated. In (b,c,d) the $\mathtt{d_L}$ distributions are illustrated for FakeAVCeleb \cite{khalid2021fakeavceleb}, LAV-DF \cite{cai2022you}, and AV-Deepfake1M \cite{cai2024av}.}
    \label{fig:motivation}
\end{figure*}

Humans integrate multimodal sensory information, such as visual and auditory input, to extract meaningful features and perform a wide range of recognition tasks. Cognitive neuroscience research has extensively documented the interplay between video and audio \cite{stacey2016contribution,sumby1954visual}, often resulting in the alteration of perceptual content \cite{williams2022you,chen2018you} or even the induction of perceptual illusions \cite{o2008seeing,shams2002visual}. In the context of speech perception, the brain actively constructs content representations by combining visual, auditory, and contextual cues to predict ongoing and future utterances \cite{chandrasekaran2011when}. When visual information conflicts with auditory input, as in the McGurk effect \cite{nath2012neural} or in the case of deepfakes, increased prediction errors are observed, accompanied by significant changes in brain activity, which often manifest as higher-frequency neural oscillations and localized activation patterns, indicative of heightened cognitive effort \cite{arnal2011transitions}.

We illustrate how state-of-the-art AI analysis tools could emulate this process in \cref{fig:motivation}. In \cref{subfig:motivation_example}, an input video is decomposed into its visual and audio streams feeding state-of-the-art models for Visual and Audio Speech Recognition (VSR, ASR)\footnote{Also known as lip-reading and speech-to-text, respectively.} \cite{ma2022visual}. 
The difference between the two textual outputs is quantified using the normalized Levenshtein distance ($\mathtt{d_L}$), calculated by dividing the edit distance (\texttt{delta})\footnote{Determined through Python's \texttt{difflib}.} by the combined length of the texts, resulting in a percentage from 0\% (identical) to 100\% (completely different).
\Cref{subfig:violin_favc,subfig:violin_lavdf,subfig:violin_avdf1m} illustrate aggregate results computed on the evaluation sets of three popular audio-visual deepfake detection benchmarks, FakeAVCeleb \cite{khalid2021fakeavceleb}, LAV-DF \cite{cai2022you}, and AV-Deepfake1M \cite{cai2024av}. 
Real videos, on average, showed lower difference scores $\mathtt{d_L}$ than fake videos across all datasets, confirming the score's ability to measure incongruity.
However, even though these differences are significant on FakeAVCeleb, they are smaller on LAV-DF, and much smaller on AV-Deepfake1M, which renders \textit{naive} thresholding unsuitable for robust detection. This is due to the fact that FakeAVCeleb's videos are fully manipulated, while LAV-DF's samples are manipulated only in parts, and samples from AV-Deepfake1M are manipulated in even smaller parts (on average half the length of those in LAV-DF).

To leverage machine perception discrepancies between visual and audio speech, we introduce Discourse Modality-information Differentiation (\method). The proposed architecture decomposes the input video into its visual and audio streams, and extracts features via the Discourse-related Feature Extraction (DiFE) module, which uses the pre-trained VSR and ASR models proposed in \cite{ma2022visual}. Subsequently, the Modality-information Differentiation (MiD) module, implemented via a Transformer encoder with local cross-modal attention and feature pyramids, identifies frame-level inconsistencies and utilizes them for deepfake detection and localization.
Note that low difference scores $\mathtt{d_L}$ frequently occur in partly manipulated fake videos (e.g. \cref{subfig:motivation_example}); however, \method\ manages to classify them as fake by capturing audio-visual divergence at the frame level. In addition, high difference scores $\mathtt{d_L}$ occasionally occur in real videos; however, \method\ leverages alternative features, e.g., hard \textit{phonemes} \cite{bear2017phoneme}, to classify them as real (cf. \cref{subsec:results_dfd}). A composite loss optimizes frame-level detection and fake interval prediction (overlap and boundaries). We evaluate \method\ on Deepfake detection (DFD) and Temporal Forgery Localization (TFL) tasks using three audio-visual deepfake detection benchmarks, FakeAVCeleb \cite{khalid2021fakeavceleb}, LAV-DF \cite{cai2022you}, and AV-Deepfake1M \cite{cai2024av}. \method\ outperforms the state-of-the-art by a considerable margin on the challenging AV-Deepfake1M (+30.5\% AUC for DFD and +47.88\% AP@0.75 for TFL), while it maintains top performance on FakeAVCeleb and LAV-DF. 
The main contributions of this paper are:
\begin{enumerate}
    \item A novel audio-visual deepfake detection methodology leveraging speech representations from both modalities to identify local cross-modal incongruities, called \method.
    \item An extensive evaluation analysis of ablation, comparative, robustness, generalization, and in-the-wild experiments demonstrating \method's effectiveness on deepfake detection and localization.
    \item An inherent capability for interpreting DFD outputs through \method's cross-modal representation incongruity.
\end{enumerate}

\section{Related Work}\label{sec:related_works}
\subsection{Deepfake Detection Approaches}
Deepfake detection approaches considering visual-only manipulations \cite{shiohara2022detecting,rossler2019faceforensics++,zheng2021exploring,li2020face,afchar2018mesonet,coccomini2022combining,qian2020thinking,shuai2023locate,li2023spatio}, while valuable, overlook the audio component of videos and its interplay with the visual content. We compare \method\ with a representative set of these approaches following previous practices.

Recent deepfake detection approaches employ multimodal techniques like fusion, contrastive learning, and self-supervised learning to identify discrepancies between visual and audio streams. While these methods effectively capture audio-visual inconsistencies, they neglect the speech component of videos. For instance, attention-based frameworks analyze synchronization patterns \cite{zhou2021joint}, spatio-temporal models assess inter- and intra-modal disharmonies \cite{yang2023avoid}, and forgery-aware adaptation is applied to pre-trained ViTs \cite{nie2024frade}. Similarly, \cite{raza2023multimodaltrace,ilyas2023avfakenet,zou2024cross,wang2024building} propose joint representation learning frameworks that consider inter- and intra-modal encoding mechanisms. UMMAFormer \cite{zhang2023ummaformer} 
adopts feature reconstruction and cross-reconstruction attention combined with an enhanced feature pyramid network, while  BA-TFD \& BA-TFD+ \cite{cai2022you,cai2023glitch} adopt 3DCNN and multi-scale vision Transformer backbones guided by contrastive, frame classification, and boundary matching objectives. Contrastive learning approaches compare inter-modal representations within a video \cite{chugh2020not,cheng2023voice} or between videos \cite{zhang2024joint}. \cite{gu2021deepfake} identifies audio-visual inconsistency by focusing on feature similarity, and \cite{shahzad2022lip} by comparing the video's lip movements with audio-derived lip sequences. Self-supervised learning addresses the generalization issue of deepfake detection by exploiting natural audio-visual synchronization in real videos \cite{zhang2023video,grill2020bootstrap,haliassos2022leveraging,feng2023self,cozzolino2023audio,oorloff2024avff,yu2023pvass,esser2021taming}. Representation learning has also been tailored to modality emotion consistency \cite{mittal2020emotions} and articulatory/lip movement consistency \cite{wang2024audio}. Finally, works \cite{lin2019bmn,nawhal2021activity,shi2023tridet,zhang2022actionformer,bagchi2021hear} dealing with Temporal Action Localization (TAL) are considered relevant by the deepfake detection literature.

Related works are presented in \cite{bohacek2024lost,li24ta_interspeech}. These rely on a simple threshold on the divergence between VSR and ASR model outputs. However, as shown in \cref{fig:motivation} this simple technique is insufficient on benchmarks with harder samples than FakeAVCeleb, such as LAV-DF and AV-Deepfake1M. In contrast, our approach extracts VSR and ASR content representations that are then processed by a Transformer-based fusion mechanism capable of identifying discriminative speech features on top of speech divergence ones. Pre-trained VSR models have been utilized as feature extractors by Haliassos et al. \cite{haliassos2021lips} for visual-only deepfake detection but neither audio speech features nor cross-modal incongruities were considered. Finally, \cite{reiss2023detecting} sets a threshold on cross-modal feature similarity deriving from an audio-visual speech self-supervised representation learning model.

\subsection{Deepfake Detection Datasets}\label{subsec:dfd_datasets}

\begin{table*}[t]
    \centering
    % \resizebox{\linewidth}{!}{
    \begin{tabular}{llllllllllll}
        \toprule
        &&\multicolumn{2}{c}{content}&&\multicolumn{2}{c}{manip.}&&\multicolumn{3}{c}{target}\\
        \cline{3-4}\cline{6-7}\cline{9-11}
        \\
        dataset&size&A&V&&A&V&&A&V&joint&task\\
        \midrule
        DFDC \cite{dolhansky2020deepfake} &128,154&\checkmark&\checkmark&&\checkmark&\checkmark&&\xmark&\xmark&\checkmark&DFD\\
        KoDF \cite{kwon2021kodf}&237,942&\checkmark&\checkmark&&\xmark&\checkmark&&\xmark&\checkmark&\xmark&DFD\\
        FakeAVCeleb \cite{khalid2021fakeavceleb}&21,544&\checkmark&\checkmark&&\checkmark&\checkmark&&\checkmark&\checkmark&\xmark&DFD\\
        LAV-DF \cite{cai2022you}&136,304&\checkmark&\checkmark&&\checkmark&\checkmark&&\checkmark&\checkmark&\xmark&DFD,TFL\\
        AV-Deepfake1M \cite{cai2024av}&1,146,760&\checkmark&\checkmark&&\checkmark&\checkmark&&\checkmark&\checkmark&\xmark&DFD,TFL\\
        \bottomrule
    \end{tabular}
    % }
    \caption{Deepfake detection datasets providing both video and audio. Size determines the number of samples, content specifies the provided modalities, manipulation specifies which modality has been manipulated, target specifies for which modality the dataset provides ground truth, while task is reported in the last column.}
    \label{tab:datasets}
\end{table*}

During the past years, several datasets have been proposed for video deepfake detection. Most of them focus solely on visual manipulations, thus including only the visual modality in their samples, e.g., DF-TIMIT \cite{korshunov2018deepfakes}, FaceForensics++ \cite{rossler2019faceforensics++}, DeeperForensics \cite{jiang2020deeperforensics}, Celeb-DF \cite{li2020celeb}, WildDeepfake \cite{zi2020wilddeepfake}, and DF-Platter \cite{narayan2023df}.

Our focus is on the detection of audio-visual deepfakes, which require content, manipulations and annotations in both modalities. \Cref{tab:datasets} presents deepfake detection datasets that provide both visual and audio streams. DFDC \cite{dolhansky2020deepfake} is a seminal large-scale benchmark in the field that contains approximately 128K samples. Although it provides audio-visual content and contains manipulations on both modalities, (i) it does not provide separate manipulation annotations per modality, (ii) fake audio parts appear in only 4\% of the total number of videos, while fake visual parts in 84\%\footnote{Based on a later estimation: \url{https://www.kaggle.com/datasets/basharallabadi/dfdc-video-audio-labels}.}, and (iii) it contains videos where the depicted individual does not talk or the speaking person is not depicted and the depicted one does not talk. These features render DFDC inappropriate for audio-visual deepfake detection. KoDF \cite{kwon2021kodf} is another large-scale benchmark containing 237K videos of Korean subjects. Although it provides both the visual and audio components of its samples, it contains only visual manipulations, which makes it unsuitable for our problem. FakeAVCeleb \cite{khalid2021fakeavceleb}, contains 21K manipulated and 0.5K real videos with manipulations on both modalities and corresponding annotations. Due to its size it is commonly used as an evaluation benchmark; here, we consider it for training and evaluation purposes with real video additions from the VoxCeleb2 dataset \cite{chung2018voxceleb2}. LAV-DF \cite{cai2022you} is a recently introduced dataset that contains 136K real and fake videos with manipulations and targets on both modalities for two tasks, Deepfake Detection (DFD) and Temporal Forgery Localization (TFL). Finally, AV-Deepfake1M \cite{cai2024av} is the largest audio-visual deepfake detection benchmark to date containing over 1.1M real and fake videos with manipulations and targets on both modalities, for both the DFD and TFL tasks. 

\section{Methodology}
\begin{figure*}[th]
    \centering
    \includegraphics[trim={3.7cm 16.3cm 15.4cm 3.45cm},clip,width=\textwidth]{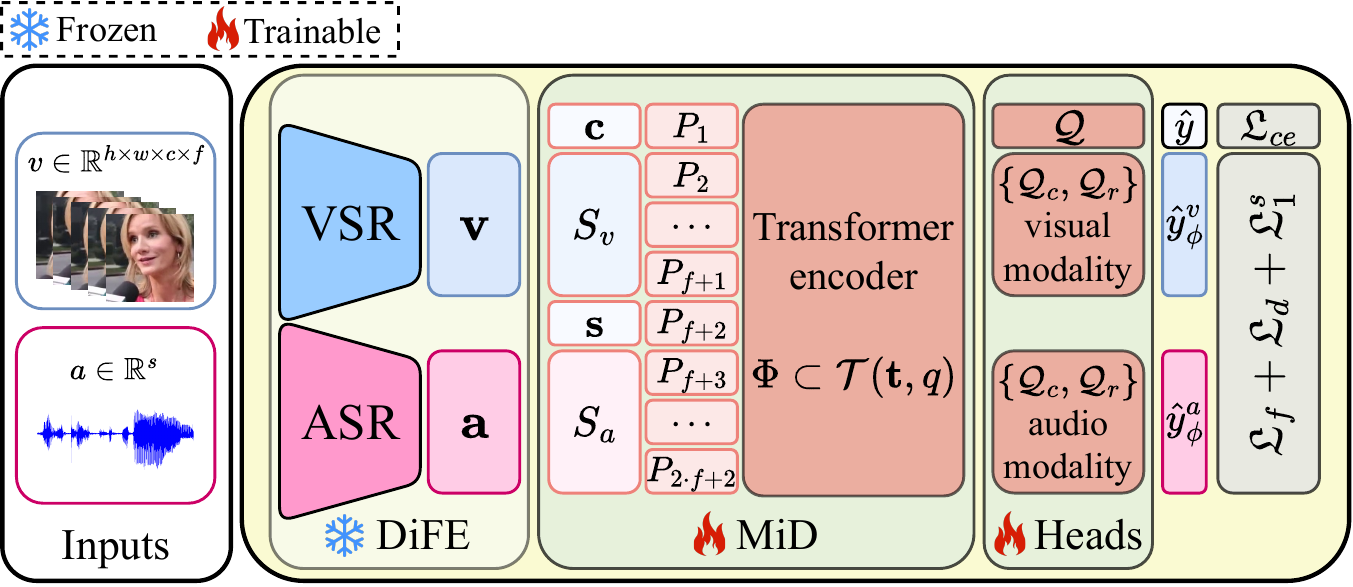}
    \caption{The \method\ architecture.}
    \label{fig:architecture}
\end{figure*}

In this section we elaborate on the main components of \method, namely Discourse-related Feature Extraction (DiFE), Modality-information Differentiation (MiD), classification \& regression heads, and objective function. \method's architecture is illustrated in \Cref{fig:architecture}.

\subsection{Problem Formulation}
Consider $(\mathfrak{v}, y, \mathbf{y})\in\mathfrak{D}$ where $\mathfrak{v}=\{v,a\}$ denotes a video of dataset $\mathfrak{D}$, containing the visual $v\in\mathbb{R}^{h\times w\times c \times f}$ and the audio $a\in\mathbb{R}^{s}$ signal\footnote{Utterances mainly comprise monophonic signals; conversion to single-channel is straightforward if required.}, with $h$ denoting the height, $w$ the width, $c$ the number of channels, $f$ the number of video frames, and $s$ the number of audio samples. $y\in\{0,1\}^2$ denotes the Deepfake Detection (DFD) target per modality, while $\mathbf{y}=\{y^v_{1},\dots, y^v_{f},y^a_{1},\dots, y^a_{f}\}$, with $y^m_{\phi}=(d^{m,s}_{\phi},d^{m,e}_{\phi},a^m_{\phi})$, denotes the Temporal Forgery Localization (TFL) target, assigning a forgery ground truth $a^m_{\phi}\in\{0,1\}$ to each modality $m\in\{v,a\}$ and frame $\phi$, along with the corresponding distances $d^{m,s}_{\phi},d^{m,e}_{\phi}\in\mathbb{R}$ between frame $\phi$ and the onset $s^m_{\phi}$ and offset $e^m_{\phi}$ of a fake interval containing $\phi$.  We train deep networks $\mathcal{F}$ on the DFD task $\hat{y}=\mathcal{F}(\mathfrak{v})$ and the TFL task $\hat{\mathbf{y}}=\mathcal{F}(\mathfrak{v})$ separately, while predictions $\hat{\mathbf{y}}$ are decoded by:
\begin{equation}
    \hat{a}^m_{\phi}>0.5\text{, }\hat{s}^m_{\phi}=\phi-\hat{d}^{m,s}_{\phi}\text{, }\hat{e}^m_{\phi}=\phi-\hat{d}^{m,e}_{\phi}
\end{equation}
$\hat{\cdot}$ denotes prediction while its absence ground truth.

\subsection{Discourse-related Feature Extraction (DiFE)}
\label{subsec:feature_extraction}
Discourse modality-information differentiation (\method) is learned by utilizing speech features extracted with disjoint-only encoding of visual and audio streams. Otherwise, e.g., by utilizing Audio-Visual Speech Recognition (AVSR) encoders, information missing from one modality due to deepfake artifacts is retrieved from the other resulting in indistinguishable cross-modal differences between real and fake samples. From each video $\mathfrak{v}=\{v,a\}$, we extract the visual $\mathbf{v}\in\mathbb{R}^{f\times d_0}$ and audio $\mathbf{a}\in\mathbb{R}^{f\times d_0}$ features corresponding to its visual $v$ and audio $a$ components, denoting with $d_0$ the embedding dimension, based on the distinct VSR and ASR models proposed in \cite{ma2022visual}. Then, $\mathbf{v}$ and $\mathbf{a}$ are projected to $d$ dimensions using a lightweight 1-D convolutional network with two ReLU-activated layers. Under ideal conditions, in real videos, $\mathbf{v}$ and $\mathbf{a}$ contain the same information corresponding to the discourse, while in deepfakes that contain manipulated parts, differentiation is expected due to artifacts. 

\subsection{Modality-information Differentiation (MiD)}
Trainable sequence encoding $S_v,S_a\in\mathbb{R}^d$ is added to $\mathbf{v}$ and $\mathbf{a}$, while a separation token $\mathbf{s}\in\mathbb{R}^{d}$ is employed to encode modality information. An additional classification token $\mathbf{c}\in\mathbb{R}^{d}$ is only employed in DFD experiments, and position encoding $P_\tau\in\mathbb{R}^d$, with $\tau=1,\dots,2\cdot f+e$, is applied on all tokens, resulting in tensor $\mathbf{t}=\oplus\{\mathbf{c}, \mathbf{v}, \mathbf{s}, \mathbf{a}\}\in\mathbb{R}^{(2\cdot f+e)\times d}$, with $\oplus$ denoting concatenation, being the input to the Transformer encoder $\mathcal{T}$ \cite{vaswani2017attention}. $e$ depends on the task, being equal to 2 in DFD experiments and equal to 1 in TFL experiments where $\mathbf{c}$ is absent. Given that short-term inter-modality inconsistency detection is key to our tasks, multimodal local attention is preferable in contrast to global; besides, it is computationally more efficient. Thus, we limit the Transformer's attention to a multimodal local window of $2\cdot q$ tokens $\tau\in[1+\phi-\floor*{q/2},1+\phi+\floor*{q/2}]\cup[f+2+\phi-\floor*{q/2},f+2+\phi+\floor*{q/2}]$ for frames $\phi$, while $\mathbf{c}$ and $\mathbf{s}$ attend to all tokens. Then, we extract $d$-dimensional feature pyramids corresponding to $\mathbf{c}$ (if applicable), $\mathbf{v}$, and $\mathbf{a}$ tokens deriving from the $l$ layers of the Transformer $\mathcal{T}$, denoted by $\Phi\subset\mathcal{T}(\mathbf{t},q)$, with $\Phi\in\mathbb{R}^{(2\cdot f+e-1)\times l\times d}$. To do so, $\mathcal{T}$ processes $\mathbf{t}$ as:
\begin{align}
    \tilde{\mathbf{z}}_\lambda&=\text{MSA}(\text{LN}(\mathbf{z}_{\lambda-1}))\\
    \mathbf{z}_{\lambda}&=\text{MLP}(\text{LN}(\tilde{\mathbf{z}}_\lambda))+\tilde{\mathbf{z}}_\lambda
\end{align}
where $\lambda=1,\dots,l$ denotes the layer index, $\mathbf{z}_0=\mathbf{t}$ is the input, MSA denotes Multi-head Softmax Attention \cite{vaswani2017attention} with $r$ number of heads, LN denotes Layer Normalization \cite{lei2016layer}, and MLP denotes a Multi-Layer Perceptron with internal dimensionality $d\cdot u$. Then, $\Phi=\oplus\{\mathbf{z}_\lambda\}_{\lambda=1}^l$, omitting the separation token $\mathbf{s}$. 

\subsection{Classification and Regression Heads}
In DFD experiments, predictions $\hat{y}$ are made by a feed-forward classification head $\mathcal{Q}$ that takes as input the classification tokens of $\Phi$, denoted by $\Psi\in\mathbb{R}^{l\times d}$, and consists of three linear layers, the first two of which are followed by Layer Normalization (LN) and ReLU activation:
\begin{align}
    \dot{\Psi}&=\text{ReLU}\big(\text{LN}(\Psi\cdot \mathrm{W}_1+\mathrm{b}_1)\big)\\
    \ddot{\Psi}&=\text{ReLU}\big(\text{LN}(\dot{\Psi}\cdot \mathrm{W}_2+\mathrm{b}_2)\big)\\
    \hat{y}&=\ddot{\Psi}\cdot \mathrm{W}_3+\mathrm{b}_3
\end{align}
where W$_1$,W$_2\in\mathbb{R}^{d\times d}$, b$_1$,b$_2\in\mathbb{R}^{d}$, W$_3\in\mathbb{R}^{d\times 2}$, b$_3\in\mathbb{R}^{2}$ are learnable parameters, and $\hat{y}\in\mathbb{R}^{l\times 2}$ are the logits for the visual and audio modalities per Transformer layer. In TFL experiments, predictions $\hat{\mathbf{y}}$ are made by two lightweight 1-D convolutional networks, namely the classification $\mathcal{Q}_c$ and regression $\mathcal{Q}_r$ heads (with input $\Phi$ omitting $\mathbf{c}$), each consisting of three convolutional layers, the first two of which are followed by Layer Normalization (LN) and ReLU activation. The classification head predicts $l$ probabilities $\hat{a}^m_{\phi}\in\mathbb{R}^l$ for frame $\phi$ of modality $m$ to be fake, while the regression head predicts $l$ distance pairs ($\hat{d}^{m,s}_\phi$, $\hat{d}^{m,e}_\phi$)$\in\mathbb{R}^{l\times 2}$, respectively.

\subsection{Objective Function}
For the DFD task, we consider the binary cross-entropy loss $\mathfrak{L}_{ce}$ \cite{good1952rational}, optimizing the real vs. fake objective. For the TFL task, we propose a composite loss as the combination of three loss functions, namely (1) the focal loss $\mathfrak{L}_{f}$ \cite{lin2017focal} to optimize the classification objective, while accounting for class imbalance being prevalent in each video's frames, (2) the DIoU loss $\mathfrak{L}_{d}$ \cite{zheng2020distance} to optimize the regression objective by maximizing the overlap between predicted and ground truth fake time intervals, and (3) the smooth L1 loss $\mathfrak{L}_{1}^s$ \cite{girshick2015fast} to minimize the distance between predicted and actual boundaries of fake time intervals. Detailed mathematical expressions of loss function computation can be found in the supplementary material.

\section{Experimental Setup}
\subsection{Datasets}
We consider FakeAVCeleb \cite{khalid2021fakeavceleb}, LAV-DF \cite{cai2022you}, and AV-Deepfake1M \cite{cai2024av} datasets for training and evaluation. FakeAVCeleb contains 21K fake and 500 real samples which we split in 70\% for training and 30\% for testing, following \cite{yang2023avoid,oorloff2024avff}\footnote{To balance classes during training only, we add real samples from VoxCeleb2 \cite{chung2018voxceleb2} including equal number of male and female speakers of several ethnicities talking in English. Language identification via \url{https://github.com/speechbrain/speechbrain}.}, 
while we keep 5\% of the training set for validation. LAV-DF has been released with standard splits of 78K training, 31K validation, and 26K test samples. AV-Deepfake1M has 746K training, 57K validation, and 343K test samples; the metadata of the latter have not been released. Supplementary material provides further details on class and split sizes.

\subsection{Evaluation}
For DFD, we use accuracy (ACC), Average Precision (AP), and area under ROC curve (AUC). For TFL, we use Average Precision at $p$ (AP@$p$), and Average Recall at $n$ (AR@$n$). We report metrics per dataset in accordance to previous works for comparability. On AV-Deepfake1M we evaluate through Codabench platform\footnote{\url{https://deepfakes1m.github.io/evaluation}}. Competing methods are audio-visual deepfake \& face manipulation detectors, and temporal action localization models (cf. \Cref{sec:related_works}). Also for DFD the \textit{naive} thresholding on $\mathtt{d_L}$ approach (cf. \cref{fig:motivation}).

\subsection{Implementation Details}
We train \method\ for 100 epochs with early stopping patience 10 and checkpoint based on validation metrics, AUC for DFD and the sum of AP@$p$ and AR@$n$ for TFL. Batch size is set to 64, initial learning rate to 0.001 (reduced on plateau)
with Adam optimizer, local attention window size $q$ to 15, and focal loss' hyperparameters $\alpha=0.98$ and $\gamma=2$. We set a maximum sequence length $f$ to 600 and zero-pad the smaller. We consider a grid for $\mathcal{T}$'s size, with $(d,r,u)\in\{(32,2,1),(64,2,1),(64,4,1),(128,4,2),(256,8,2)\}$ and number of layers $l\in\{1,3,5\}$. Grid search for AV-Deepfake1M is performed with 200K training samples to reduce resource requirements, but retrain the best configuration on the whole set.
We spent $\sim$3900 GPU hours for training on NVIDIA GeForce RTX 3090 Ti GPUs.

\section{Results}
\subsection{Ablation Study}
\begin{figure}[t]
     \centering
     \begin{subfigure}[b]{0.45\textwidth}
         \centering
         \includegraphics[width=\textwidth]{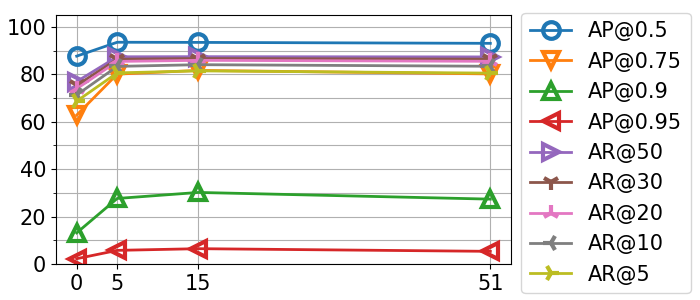}
         \caption{AV-Deepfake1M: Window size $q$}
         \label{fig:ablation_tfl_avdeepfake1m_win_size}
     \end{subfigure}

     \begin{subfigure}[b]{0.45\textwidth}
         \centering
         \includegraphics[width=\textwidth]{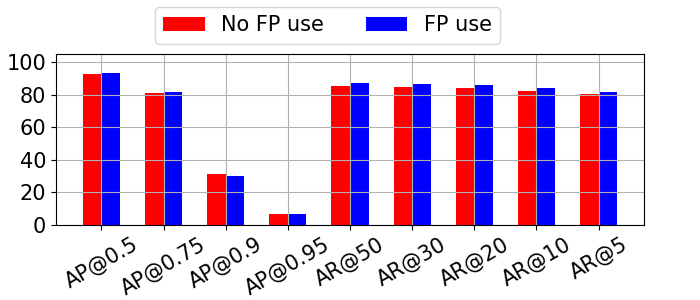}
         \caption{AV-Deepfake1M: Feature pyramid (FP) use}
         \label{fig:ablation_tfl_avdeepfake1m_feature_pyramid}
     \end{subfigure}

     \begin{subfigure}[b]{0.45\textwidth}
         \centering
         \includegraphics[width=\textwidth]{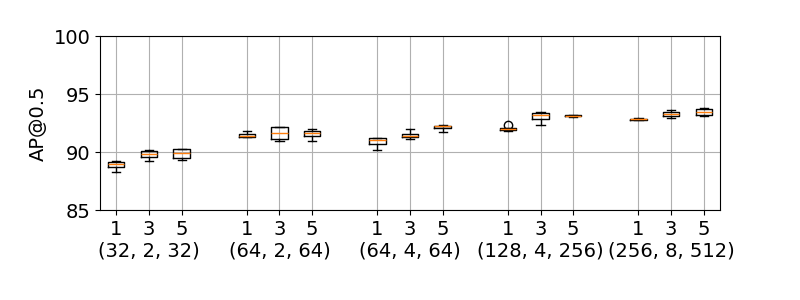}
         \caption{AV-Deepfake1M: $\mathcal{T}$ size; $l$ \& $(d,r,d\cdot u)$}
         \label{fig:ablation_hp_tfl_avdeepfake1m}
     \end{subfigure}

    \caption{Ablation and hyperparameter tuning analysis.}
    \label{fig:ablation}
\end{figure}

Other speech-related pre-trained models are equivalent for use as backbones in \method, e.g., AV-Hubert \cite{shi2022learning} (cf. \cref{subsec:feature_extraction}). To study their impact, we train \method\ on FakeAVCeleb and LAV-DF w/ AV-Hubert features and present the results in \Cref{tab:avhubert}. This variant performs slightly worse on FakeAVCeleb and LAV-DF (DFD), better wrt AP@0.5 and worse wrt AR@100 on LAV-DF (TFL). Henceforth, we opt for the better performing and more intuitive speech prediction variant (Ma 2022) \cite{ma2022visual}. \Cref{fig:ablation} presents ablations on AV-Deepfake1M\footnote{We report validation performance as conducting the ablations through Codabench would be impractical.}. 
\Cref{fig:ablation_tfl_avdeepfake1m_win_size} indicates that a small window size $q=15$ for Transformer's $\mathcal{T}$ local attention is optimal, especially in contrast to attending on all tokens determined by $q=0$. Also, the feature pyramid scheme provides a small performance increase as indicated by \Cref{fig:ablation_tfl_avdeepfake1m_feature_pyramid}. The size of $\mathcal{T}$ is important for achieving maximum performance as shown in 
\Cref{fig:ablation_hp_tfl_avdeepfake1m}, both in terms of number of layers $l$ and layer size determined by input dimension $d$, number of attention heads $r$, and internal dimension $d\cdot u$: the larger the model the higher its performance. Supplementary material provides further TFL task ablations on AV-Deepfake1M wrt hyperparameter $\alpha$ and learning rate scheduler, the same analysis on LAV-DF, and a pertinent DFD task ablation analysis.

\begin{table}[t]
    \centering
    \resizebox{\linewidth}{!}{
    \begin{tabular}{ccccc}
    \toprule
    & FAVC & L-DF (DFD) & L-DF (TFL) \\
    \midrule
    AV-Hubert \cite{shi2022learning} & 99.98$^\ddagger$, 99.35* & 99.79$^\ddagger$, 99.78* & \textbf{98.00}$^+$, 93.00$^\dagger$ \\
    (Ma 2022) \cite{ma2022visual} & \textbf{99.99}$^\ddagger$, \textbf{99.71}* & \textbf{99.94}$^\ddagger$, \textbf{99.84}* & 95.49$^+$, \textbf{94.17}$^\dagger$\\
    \bottomrule
    \end{tabular}
    }
    \caption{Speech-related pre-trained backbone ablation analysis. $^\ddagger$AP, *AUC, $^+$AP@0.5, $^\dagger$AR@100.}
    \label{tab:avhubert}
\end{table}

\subsection{Deepfake Detection}\label{subsec:results_dfd}
\Cref{tab:lavdf_dfd,tab:avdeepfake1m_dfd,tab:favc_dfd}, present comparative results in terms of video-level Deepfake Detection (DFD) on FakeAVCeleb \cite{khalid2021fakeavceleb}, LAV-DF \cite{cai2022you}, and AV-Deepfake1M \cite{cai2024av}. On FakeAVCeleb, \method\ outperforms competitive methods achieving +0.8 ACC, and +0.6 AUC. On LAV-DF, \method\ performs slightly better than UMMAFormer \cite{zhang2023ummaformer}, achieving 99.84 AUC. On AV-Deepfake1M dataset, \method\ outperforms the state-of-the-art by a significant absolute performance increase of +30.5 AUC\footnote{Codabench does not provide ACC for the test samples, thus we report the validation accuracy, which, even though is not comparable, still provides a sense on the level of uncalibrated threshold performance increase.}. \Cref{tab:favc_cross} presents cross-manipulation performance on FakeAVCeleb in which \method\ achieves 100.0 AVG-FV AP yet 66.4 AP on RVFA. Notably, \method\ presents perfect detection rate in face-swaps (100.0 AUC). Mouth in face-swaps moves occasionally unnaturally yielding inaccurate lip-reading predictions, while audio is unaffected yielding correct speech-to-text predictions. \method\ translates this inconsistency into a deepfake detection. The \textit{naive} classifier achieves high performance on FakeAVCeleb but moderate performance on LAV-DF and AV-Deepfake1M; however, it outperforms several methods. Finally, challenging real samples with 
$\mathtt{d_L}$ over 0.8 (cf. \cref{sec:intro}), are scored by \method\ with an average of 0.19, and 0.16 on LAV-DF, and AV-Deepfake1M, respectively, attesting to \method's advantage wrt \textit{naive} distance-thresholding.

\begin{table*}[!t]
    \centering
    \begin{minipage}{0.32\textwidth}
    \centering
    \resizebox{\textwidth}{!}{
    \begin{tabular}{llll}
    \toprule
    Method&Modality&ACC&AUC\\
    \midrule
    \textit{naive}&$\mathcal{AV}$&97.7&93.3\\
    \midrule
    Xception \cite{rossler2019faceforensics++}& $\mathcal{V}$ &67.9& 70.5\\
    LipForensics \cite{haliassos2021lips}& $\mathcal{V}$ & 80.1 &82.4\\
    FTCN \cite{zheng2021exploring} &$\mathcal{V}$ & 64.9 &84.0\\
    MDS \cite{chugh2020not}& $\mathcal{AV}$& 82.8 &86.5\\
    AVoiD-DF \cite{yang2023avoid}& $\mathcal{AV}$& 83.7 &89.2\\
    ART-AVDF \cite{wang2024audio}&$\mathcal{AV}$&96.4&98.2\\
    AVFF \cite{oorloff2024avff}& $\mathcal{AV}$& \underline{98.6} &\underline{99.1}\\
    \midrule
    \method\ (ours)&$\mathcal{AV}$&\textbf{99.4}&\textbf{99.7}\\
    \bottomrule
    \end{tabular}
    }
    \caption{In-dataset perfromance on FakeAVCeleb \cite{khalid2021fakeavceleb}. Modality denotes the model's input type with $\mathcal{V}$ being visual and $\mathcal{A}$ audio. \textbf{Bold} indicates best and \underline{underline} second to best performance.}
    \label{tab:favc_dfd}
    \end{minipage}\hfill
    \begin{minipage}{0.28\textwidth}
    \centering
    \resizebox{\textwidth}{!}{
    \begin{tabular}{lll}
    \toprule
        Method & Modality&AUC\\
    \midrule
    \textit{naive}&$\mathcal{AV}$&77.8\\
    \midrule
    F\textsuperscript{3}-Net \cite{qian2020thinking}&$\mathcal{V}$&52.0\\
    MDS \cite{chugh2020not}&$\mathcal{AV}$&82.8\\
    EfficientViT \cite{coccomini2022combining}&$\mathcal{V}$&96.5\\
    BA-TFD \cite{cai2022you}&$\mathcal{AV}$&99.0\\
    UMMAFormer \cite{zhang2023ummaformer}&$\mathcal{AV}$&\underline{99.8}\\
    \midrule
        \method\ (ours) &$\mathcal{AV}$& \textbf{99.84}\\
    \bottomrule
    \end{tabular}
    }
    \caption{Deepfake detection results on LAV-DF \cite{cai2022you}. Modality denotes the model's input type with $\mathcal{V}$ being visual and $\mathcal{A}$ audio. \textbf{Bold} indicates best and \underline{underline} second to best performance.}
    \label{tab:lavdf_dfd}
    \end{minipage}
\hfill
\begin{minipage}{0.38\textwidth}
    \centering
    \resizebox{\textwidth}{!}{
    \begin{tabular}{llll}
    \toprule
        Method & Modality&AUC & ACC\\
    \midrule
    \textit{naive}&$\mathcal{AV}$&\textcolor{gray}{62.8*}&\textcolor{gray}{77.1*}\\
    \midrule
    Video-LLaMA (13B) E5 \cite{zhang2023video}& $\mathcal{AV}$& 50.7& 25.1\\
    LipForensics \cite{haliassos2021lips}&$\mathcal{V}$&51.6& 68.8\\
    Face X-Ray \cite{li2020face}& $\mathcal{V}$& 61.5& 73.8\\
    Meso4 \cite{afchar2018mesonet}&$\mathcal{V}$&50.2& 75.0\\
    MesoInception4 \cite{afchar2018mesonet}&$\mathcal{V}$&50.1& 75.0\\
    SBI \cite{shiohara2022detecting}&$\mathcal{V}$&\underline{65.8}& 69.0\\
    MDS \cite{chugh2020not}& $\mathcal{AV}$& 56.6& 59.4\\
    \midrule
        \method\ (ours) &$\mathcal{AV}$& \textbf{96.3}&\textcolor{gray}{96.3*}\\
    \bottomrule
    \end{tabular}
    }
    \caption{Deepfake detection results on AV-Deepfake1M \cite{cai2024av}. Modality denotes the model's input type with $\mathcal{V}$ being visual and $\mathcal{A}$ audio. \textbf{Bold} indicates best and \underline{underline} second to best performance. *Computed on validation set.}
    \label{tab:avdeepfake1m_dfd}
    \end{minipage}
\end{table*}

\begin{table*}
    \centering
    \resizebox{\textwidth}{!}{
    \begin{tabular}{llllllllllllllll}
    \toprule
        &&&&\multicolumn{12}{c}{Category}\\
        \cmidrule{5-16}
        &Method &Modality&Pre-training&  \multicolumn{2}{c}{RVFA}&  \multicolumn{2}{c}{FVRA-WL}&   \multicolumn{2}{c}{FVFA-FS}&  \multicolumn{2}{c}{FVFA-GAN}&\multicolumn{2}{c}{FVFA-WL}&\multicolumn{2}{c}{AVG-FV}\\
        \cmidrule(lr){5-6}\cmidrule(lr){7-8}\cmidrule(lr){9-10}\cmidrule(lr){11-12}\cmidrule(lr){13-14}\cmidrule(lr){15-16}
        &&&& AP&AUC& AP&AUC& AP&AUC& AP&AUC& AP&AUC& AP&AUC\\
    \midrule
    &\textit{naive}&$\mathcal{AV}$&-&61.2 & 65.6&99.2 & 96.7&98.3 & 95.6&98.8 & 95.7&99.3 & 96.8&98.9 & 96.2\\
    \cmidrule{2-16}
    \multirow{5}{*}{\rotatebox[origin=c]{90}{Unsupervised}}&AVBYOL \cite{grill2020bootstrap}&$\mathcal{AV}$&LRW \cite{chung2017lip}&50.0& 50.0& 73.4& 61.3& 88.7& 80.8& 60.2& 33.8& 73.2& 61.0& 73.9& 59.2\\[4pt]
    &VQ-GAN \cite{esser2021taming}&$\mathcal{V}$&LRS2 \cite{afouras2018deep}&-&-&50.3& 49.3& 57.5& 53.0& 49.6& 48.0& 62.4& 56.9& 55.0& 51.8\\[4pt]
    &AVAD \cite{feng2023self}&$\mathcal{AV}$&LRS2 \cite{afouras2018deep}& 62.4& 71.6& 93.6& 93.7& 95.3& 95.8& 94.1& 94.3& 93.8& 94.1& 94.2& 94.5\\[4pt]
    &AVAD \cite{feng2023self}&$\mathcal{AV}$&LRS3 \cite{afouras2018lrs3}&70.7& \underline{80.5}& 91.1& 93.0& 91.0& 92.3& 91.6& 92.7& 91.4& 93.1& 91.3& 92.8\\
    \midrule
    \multirow{7}{*}{\rotatebox[origin=c]{90}{Supervised}}&Xception \cite{rossler2019faceforensics++}&$\mathcal{V}$&ImageNet \cite{deng2009imagenet}&-& -& 88.2& 88.3& 92.3& 93.5& 67.6& 68.5& 91.0& 91.0& 84.8& 85.3\\
    &LipForensics \cite{haliassos2021lips}&$\mathcal{V}$&LRW \cite{chung2017lip}& -& -& \underline{97.8}& 97.7& \underline{99.9}& \underline{99.9}& 61.5& 68.1& 98.6& 98.7& 89.4& 91.1\\
    &AD DFD \cite{zhou2021joint}&$\mathcal{AV}$&Kinetics \cite{kay2017kinetics}&\underline{74.9}& 73.3& 97.0& 97.4& 99.6& 99.7& 58.4& 55.4& \textbf{100.}& \textbf{100.}& 88.8& 88.1\\
    &FTCN \cite{zheng2021exploring}&$\mathcal{V}$&-&-&-& 96.2& 97.4& \textbf{100.}& \textbf{100.}& 77.4& 78.3& 95.6& 96.5& 92.3& 93.1\\
    &RealForensics \cite{haliassos2022leveraging}&$\mathcal{V}$&LRW \cite{chung2017lip}&-&-&88.8& 93.0& 99.3& 99.1& 99.8&\underline{99.8}& 93.4& 96.7& 95.3& 97.1\\
    &AVFF \cite{oorloff2024avff}& $\mathcal{AV}$&LRS3 \cite{afouras2018lrs3}& \textbf{93.3} &\textbf{92.4}& 94.8& \underline{98.2}& \textbf{100.}& \textbf{100.}& \underline{99.9}& \textbf{100.}& \underline{99.4}& \underline{99.8}& \underline{98.5}& \underline{99.5}\\
    \cmidrule{2-16}
    &\method\ (ours)&$\mathcal{AV}$&-&66.4&51.6&\textbf{100.}&\textbf{99.8}&\textbf{100.}&\textbf{100.}&\textbf{100.}&\textbf{100.}&\textbf{100.}&\textbf{100.}&\textbf{100.}&\textbf{99.9}\\
    \bottomrule
    \end{tabular}
    }
    \caption{Cross-manipulation performance on FakeAVCeleb \cite{khalid2021fakeavceleb}. Modality denotes the model's input type with $\mathcal{V}$ being visual and $\mathcal{A}$ audio. \textbf{Bold} indicates best and \underline{underline} second to best performance. Supervised methods that use pre-training are fine-tuned on FakeAVCeleb, while unsupervised methods are not trained with labels and fake examples.}
    \label{tab:favc_cross}
\end{table*}

\subsection{Temporal Forgery Localization}
\begin{table*}[ht]
    \centering
    % \resizebox{\textwidth}{!}{
    \begin{tabular}{lllllllll}
    \toprule
        Method&Modality &  AP@0.5 & AP@0.75 & AP@0.95 & AR@100 &AR@50&AR@20&AR@10\\
    \midrule
        MDS \cite{chugh2020not} &$\mathcal{AV}$& 12.8& 1.6& 0.0& 37.9& 36.7& 34.4& 32.2\\
        AGT \cite{nawhal2021activity} &$\mathcal{V}$& 17.9& 9.4& 0.1& 43.2& 34.2& 24.6& 16.7\\
        BMN (I3D) \cite{lin2019bmn} &$\mathcal{V}$& 10.6& 1.7& 0.0& 48.5& 44.4& 37.1& 31.6\\
        AVFusion \cite{bagchi2021hear} &$\mathcal{AV}$& 65.4& 23.9& 0.1& 63.0& 59.3& 54.8& 52.1\\
        ActionFormer \cite{zhang2022actionformer} &$\mathcal{V}$& 95.3& 90.2& 23.7& 88.4& 89.6& 90.3& 90.4\\
        BA-TFD \cite{cai2022you}&$\mathcal{AV}$&76.9&38.5&0.3&66.9 &64.1 &60.8 &58.4\\
        BA-TFD+ \cite{cai2023glitch}&$\mathcal{AV}$&96.3&85.0&4.4&81.6 &80.5 &79.4 &78.8\\
        UMMAFormer \cite{zhang2023ummaformer}&$\mathcal{AV}$&\textbf{98.8}&\textbf{95.5}&\underline{37.6}&92.4&92.5&92.5&\textbf{92.1} \\
        MMMS-BA \cite{katamneni2024contextual}&$\mathcal{AV}$& \underline{97.5}& \underline{95.2}& \textbf{39.0}&\underline{94.0}&\underline{93.4}&\textbf{95.9}&89.4\\
    \midrule
        \method\ (ours)&$\mathcal{AV}$&95.5&87.9&20.6&\textbf{94.2}&\textbf{93.7}&\underline{92.7}&\underline{91.4}\\
    \bottomrule
    \end{tabular}
    % }
    \caption{Temporal forgery localization results on LAV-DF \cite{cai2022you}. Modality denotes the model's input type with $\mathcal{V}$ being visual and $\mathcal{A}$ audio. \textbf{Bold} indicates best and \underline{underline} second to best performance.}
    \label{tab:lavdf_tfl}
\end{table*}
\begin{table*}[ht]
    \centering
    \resizebox{\textwidth}{!}{
    \begin{tabular}{lllllllllll}
    \toprule
        Method&Modality &  AP@0.5 & AP@0.75 & AP@0.9 & AP@0.95 & AR@50 &AR@30&AR@20&AR@10&AR@5\\
    \midrule
        PyAnnote (Zero-Shot) \cite{plaquet2023powerset}&$\mathcal{A}$ &00.03 &00.00& 00.00& 00.00& 00.67& 00.67& 00.67& 00.67& 00.67\\
        Meso4 \cite{afchar2018mesonet}& $\mathcal{V}$& 09.86& 06.05& 02.22& 00.59& 38.92& 38.91& 38.81& 36.47& 26.91\\
        MesoInception4 \cite{afchar2018mesonet}& $\mathcal{V}$& 08.50& 05.16& 01.89& 00.50& 39.27& 39.22& 39.00& 35.78& 24.59\\
        EfficientViT \cite{coccomini2022combining}& $\mathcal{V}$& 14.71& 02.42& 00.13& 00.01& 27.04& 26.99& 26.43& 23.90& 20.31\\
        TriDet+VideoMAEv2 \cite{shi2023tridet,wang2023videomae}& $\mathcal{V}$& 21.67& 05.83& 00.54& 00.06& 20.27& 20.23& 20.12& 19.50& 18.18\\
        ActionFormer+VideoMAEv2 \cite{zhang2022actionformer,wang2023videomae}& $\mathcal{V}$& 20.24& 05.73& 00.57& 00.07& 19.97& 19.93& 19.81& 19.11& 17.80\\
        BA-TFD \cite{cai2022you}&$\mathcal{AV}$ &37.37 &6.34 &0.19 &0.02 &45.55 &40.37 &35.95 &30.66 &26.82 \\
        BA-TFD+ \cite{cai2023glitch}&$\mathcal{AV}$&44.42 &13.64 &0.48 &0.03 &\underline{48.86} &\underline{44.51} &40.37 &34.67 &29.88 \\
        UMMAFormer \cite{zhang2023ummaformer}&$\mathcal{AV}$ &\underline{51.64}&\underline{28.07}&\underline{7.65} &\underline{1.58} &44.07 &43.93 &\underline{43.45} &\underline{42.09} &\underline{40.27} \\
        MMMS-BA \cite{katamneni2024contextual}&$\mathcal{AV}$& \textcolor{gray}{62.75*}& \textcolor{gray}{35.87*}&-& \textcolor{gray}{18.37*}& \textcolor{gray}{57.49*}&-&\textcolor{gray}{55.94*}&\textcolor{gray}{54.28*}&-\\
    \midrule
        \method\ (ours)&$\mathcal{AV}$&\textbf{86.93}&\textbf{75.95}&\textbf{28.72}&\textbf{5.43}&\textbf{81.57}&\textbf{80.85}&\textbf{80.25}&\textbf{78.84}&\textbf{76.64}\\
    \bottomrule
    \end{tabular}
    }
    \caption{Temporal forgery localization results on AV-Deepfake1M \cite{cai2024av}. Modality denotes the model's input type with $\mathcal{V}$ being visual and $\mathcal{A}$ audio. \textbf{Bold} indicates best and \underline{underline} second to best performance. *Computed on validation set.}
    \label{tab:avdf1m_tfl}
\end{table*}

\Cref{tab:lavdf_tfl,tab:avdf1m_tfl} present the performance of \method\ in comparison to state-of-the-art, in terms of Temporal Forgery Localization (TFL), on LAV-DF \cite{cai2022you} and AV-Deepfake1M \cite{cai2024av}, respectively. On LAV-DF, \method\ outperforms all competitive approaches wrt AR@\{100,50\}, is second to best wrt AR@\{20,10\}, and performs below state-of-the-art wrt AP@\{0.5,0.75,0.95\}. MMMS-BA \cite{katamneni2024contextual}, UMMAFormer \cite{zhang2023ummaformer}, BA-TFD+ \cite{cai2023glitch} and ActionFormer \cite{zhang2022actionformer} exhibit similar performance, while the remaining models have lower localization ability. On AV-Deepfake1M, \method\ outperforms all competitive approaches wrt all metrics exhibiting significant performance increase. The proposed method's AR is almost double compared with the second to best, while it exhibits a notable absolute increase of +35.29\% and +47.88\% in AP at 0.5 and 0.75, respectively.

\subsection{Interpretability}

\begin{figure}[t]
    \centering
    \includegraphics[trim={0.9cm 0.7cm 1cm 1cm},clip,width=0.9\linewidth]{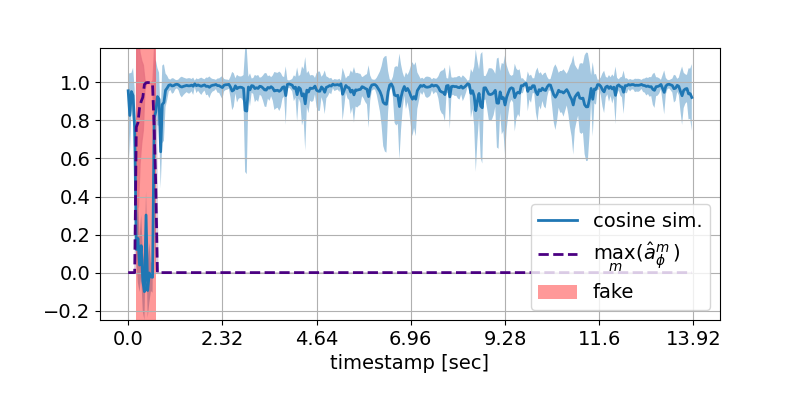}
    \caption{Fake video sample (manipulated in red). Avg. and std. (across layers $\lambda$) of cross-modal similarity and frame-level fake probability shown in blue and purple, respectively.}
    \label{fig:interpretability}
\end{figure}

\method\ is inherently interpretable as its frame-level cross-modal representations' similarity reflects the level of congruity between visual and audio speech. \Cref{fig:interpretability} illustrates a fake video example from LAV-DF\footnote{LAV-DF/test/000010.mp4} with one small manipulated part during which cross-modal similarity is significantly reduced indicating audio-visual incongruity while during real parts similarity is close to 1.0. Fake frames are perfectly identified by \method\ with $\max_{m}(\hat{a}^m_{\phi})>>0.5$. More real and fake examples are provided in suppl. material.

\subsection{Robustness}

\begin{figure}[t]
    \centering
    \includegraphics[width=0.75\linewidth]{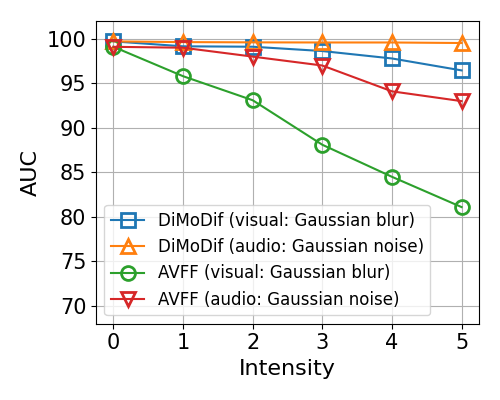}
    \caption{Robustness to visual (Gaussian blur) and audio (Gaussian noise) distortions.}
    \label{fig:robustness}
\end{figure}

\Cref{fig:robustness} illustrates \method's robustness to visual (Gaussian blurring) and audio (Gaussian noise) distortions at 5 levels of intensity. The analysis is conducted on FakeAVCeleb test set, following the evaluation protocol of \cite{oorloff2024avff,feng2023self,haliassos2021lips}\footnote{Perturbations' implementation directly sourced from \url{https://github.com/EndlessSora/DeeperForensics-1.0}.}. \method\ is not affected by the audio perturbations, outperforming AVFF\footnote{Code unavailable. Value approximations from \cite{oorloff2024avff}'s supplementary material.} \cite{oorloff2024avff} by approx. + 6.5 AUC at intensity level 5. Robustness to visual perturbations is gradually reduced to 96.5 AUC, approx. +25.0 AUC compared to AVFF. Supplementary material shows that \method\ is also more robust than AVFF under \texttt{saturation}, \texttt{contrast}, \texttt{block-wise}, and \texttt{JPEG compression} visual perturbations, yet less robust under \texttt{Gaussian noise} and \texttt{video compression}. Also, \method\ is more robust than AVFF under \texttt{pitch shift}, \texttt{reverberence}, and \texttt{audio compression} audio perturbations, achieving an unprecedented 99.3 AUC on average at intensity level 5 indicating the high robustness of speech prediction features on such distortions.

\subsection{Generalization}\label{subsec:generalization}
This paper is the first to conduct generalization experiments on the considered datasets. \Cref{tab:generalization_dfd} presents \method's cross-dataset performance on DFD. The reported performance, although high, indicates that the training data affect generalization due to scale, fake part duration, and generator quality. A small pool of low quality easier fake data (FakeAVCeleb) makes the model prone to overfitting reducing its generalization ability to partially manipulated data generated through more complex approaches (LAV-DF, AV-Deepfake1M). In contrast, large pools of high-quality and harder deepfakes (LAV-DF, AV-Deepfake1M) result in richer and more robust representations, and less overfitting. Supplementary material provides further generalization evaluations.
\begin{table}[t]
    \centering
    \resizebox{0.45\textwidth}{!}{
    \begin{tabular}{llllllll}
    \toprule
    &&\multicolumn{6}{c}{Test dataset}\\
    \cmidrule{3-8}
    &&\multicolumn{2}{c}{FakeAVCeleb}&\multicolumn{2}{c}{LAV-DF}&\multicolumn{2}{c}{AVD1M*}\\
    \cmidrule(lr){3-4}\cmidrule(lr){5-6}\cmidrule(lr){7-8}
    &&AP&AUC&AP&AUC&AP&AUC\\
    \midrule
\multirow{3}{*}{\rotatebox[origin=c]{90}{\shortstack[c]{Training\\dataset}}}&FakeAVCeleb&\textcolor{gray}{99.99}&\textcolor{gray}{99.71}&93.10&84.47&77.91&54.00\\[2pt]
&LAV-DF&99.69&90.25&\textcolor{gray}{99.94}&\textcolor{gray}{99.84}&88.46&70.40\\[2pt]
&AV-Deepfake1M&99.69&90.69&94.98&86.30&\textcolor{gray}{99.72}&\textcolor{gray}{99.18}\\
\bottomrule
\end{tabular}
}
\caption{Generalization on DFD. * Validation set.}
\label{tab:generalization_dfd}
\end{table}

\subsection{Real-world Analysis}
\begin{figure}[t]
    \centering
    \includegraphics[width=\linewidth]{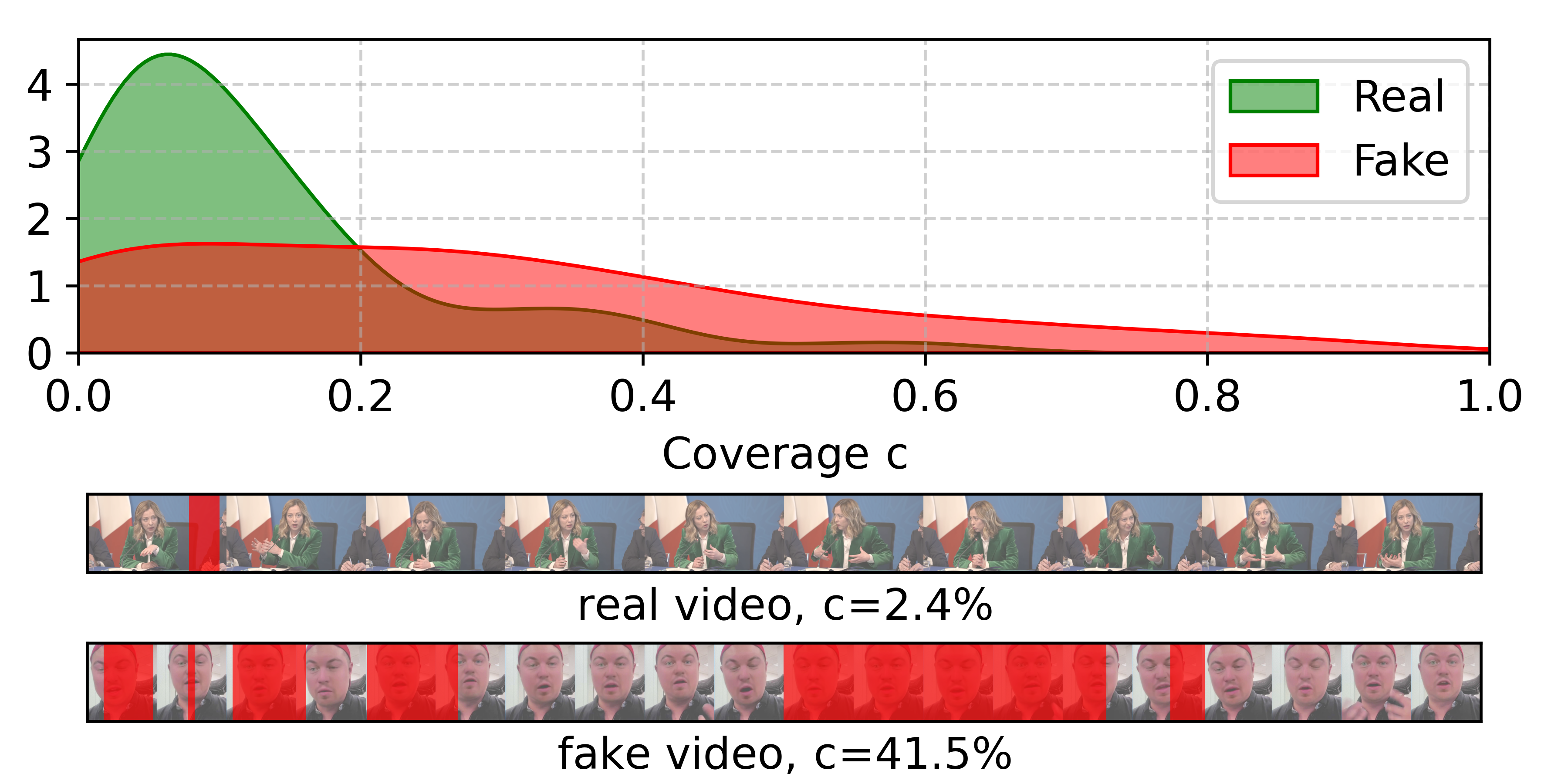}
    \caption{Distribution of coverage $\mathtt{c}$ for in-the-wild real vs. fake and examples of decisions. Red parts are flagged as fake.}
    \label{fig:itw}
\end{figure}

We also present an analysis of \method's performance in the wild. Initially, we compiled a collection of 204 Internet videos, across several languages (incl. Out-Of-Distribution $\mathtt{OOD}$) and varying lengths ($\mathtt{max}$ 46$^\prime$, $\mathtt{median}$ 1$^\prime$), with 75 of them being real and 129 of them being fake. Fakes were selected for being popular deepfakes, supporting conspiracy theories, or spreading political misinformation, and real videos for containing discourse (e.g., news anchoring). Next, we developed a pre-processing pipeline to address model's limitations. Chunking is considered for accepting any-length videos, while frame groups with small faces (e.g. depicting crowds or wide shots), no faces, or small duration (<2s) are discarded. We utilize \method\ trained on AV-Deepfake1M (TFL), and evaluate the ratio of fake to valid video parts, $\mathtt{c}=\mathtt{fake}/\mathtt{valid}$, as the model occasionally flags small parts of authentic videos as deepfakes. \method\ achieves 82.1 AP, indicating robust detection of real-world cases, with linear processing complexity $\mathcal{O}(\mathtt{[sec]})$. \Cref{fig:itw} illustrates the distribution of $\mathtt{c}$ for real vs. fake and decision examples. For reference, RealForensics \cite{haliassos2022leveraging} achieves 76.4 AP on the same set. Additionally, although the developed pipeline is tailored to real-world cases it excels on evaluation benchmarks, i.e., achieves 99.1 AP on FakeAVCeleb, 93.5 AP on LAV-DF*, and 90.6 AP on AV-Deepfake1M* (*1K random unseen samples). Further details are provided in supplementary material.

\section{Conclusions}
In this work, we propose an audio-visual deepfake detection and localization framework
that leverages cross-modal differences in machine perception of speech. It hinges on the assumption that the visual and audio signals of videos with real discourse coincide wrt information, in contrast to deepfakes that exhibit cross-modal incongruities. It considers a visual and audio speech recognition feature extraction stage, and a hierarchical cross-modal fusion learning stage along with a combination of three loss functions that optimize frame-level detection, predicted interval overlap, and boundaries. An extensive evaluation study indicates the effectiveness of our framework exhibiting state-of-the-art performance. 

\textbf{Limitations}: \method's language coverage is constrained by its speech recognition backbones although it was found to generalize well to Korean (cf. supplementary material) and in-the-wild $\mathtt{OOD}$ examples, while performance may be degraded by modality misalignment in real-world samples. Additionally, the trained model assumes a single, consistently visible speaker, sufficient video quality for facial landmark detection, and the availability of both visual and audio modalities. Development for the real-world analysis has addressed in part some of these limitations through video pre-processing. We aim to further improve in-the-wild applicability and extend functionality to multi-speaker scenarios.

\section{Acknowledgments}
This work has been partially funded by the Horizon Europe projects AI4Trust (GA No. 101070190) and vera.ai (GA No. 101070093).

\balance
\bibliographystyle{plain}
\bibliography{main}

\clearpage
\appendix
\section{Objective function details}\label{sec:loss_details}
For the DFD task, we consider the binary cross-entropy loss \cite{good1952rational}, directly optimizing the real vs. fake objective:
\begin{multline}
        \mathfrak{L}_{DFD}=\mathfrak{L}_{ce}=-\frac{1}{2l}\sum_{m\in\{v,a\}}\sum_{\lambda=1}^{l} y[m]\text{log}\hat{y}[\lambda,m]+\\(1-y[m])\cdot\text{log}(1-\hat{y}[\lambda,m])
\end{multline}
where $\lambda$ is layer index, $m$ is modality index, and $x[\cdot]$ is used to denote vector or matrix element access operation (array indexing).

For the TFL task, we consider a combination of three loss functions. For clarity we omit to also include the symbols for the averaging across $l$ layers, which however takes place as well. Specifically, we use:
\begin{enumerate}
    \item The focal loss \cite{lin2017focal} to optimize the classification objective, while accounting for class imbalance being prevalent among each video's frames:
    \begin{equation}
        \mathfrak{L}_{f}=\sum_{m\in\{v,a\}}\sum_{\phi=1}^{f} -\alpha_t^{m,\phi}(1-p_t^{m,\phi})^\gamma\text{log}(p_t^{m,\phi})
    \end{equation}
    in which:
    \begin{equation}
        \alpha_t^{m,\phi}=\begin{cases}
            \alpha & \text{if } y^m_\phi=1\\
            1-\alpha & \text{otherwise}
        \end{cases}
    \end{equation}
    \begin{equation}
        p_t^{m,\phi}=\begin{cases}
            \hat{y}^m_\phi & \text{if } y^m_\phi=1\\
            1-\hat{y}^m_\phi & \text{otherwise}
        \end{cases}
    \end{equation}
    while $\alpha$ and $\gamma$ are hyperparameters.
    
    \item The DIoU loss \cite{zheng2020distance} to optimize the regression objective by maximizing the overlap between predicted and ground truth fake time intervals:
    \begin{multline}
        \mathfrak{L}_{d}=\sum_{m\in\{v,a\}}\sum_{\phi=1}^{f} \Bigg(1.0-\text{IoU}(\hat{\mathbf{b}}^m_\phi,\mathbf{b}^m_\phi)+\\\frac{\rho^2(\hat{b}^m_\phi,b^m_\phi)}{\kappa^2}\Bigg)\cdot \mathds{1}_{m,\phi}
    \end{multline}
    where $\phi$ is frame index, $m$ is modality index, IoU denotes intersection over union, $\mathbf{b}^m_\phi=[s^m_\phi,e^m_\phi]$ denotes a time interval, $b^m_\phi=0.5\cdot (e^m_\phi+s^m_\phi)$ denotes the interval's center, $\rho$ denotes the Euclidean distance, $\kappa$ denotes the length of the smallest enclosing interval covering both predicted and ground truth intervals, and $\mathds{1}_{m,\phi}$ is an indicator function that denotes if frame $\phi$ of modality $m$ is fake.
    
    \item The smooth L1 loss \cite{girshick2015fast} to minimize the distance between predicted and actual boundaries of fake time intervals:
    \begin{multline}
    \mathfrak{L}_1^{s}=\sum_{m\in\{v,a\}}\sum_{\phi=1}^{f}\frac{1}{2}\bigg(h(s^m_\phi-\hat{s}^m_\phi)+\\h(e^m_\phi-\hat{e}^m_\phi)\bigg)\cdot \mathds{1}_{m,\phi}
    \end{multline}
    in which $\mathds{1}_{m,\phi}$ is an indicator function that denotes if frame $\phi$ of modality $m$ is fake, and $h$ is defined as:
    \begin{equation}
        h(x)=\begin{cases}
            0.5x^2 & \text{if } \abs{x} < 1 \\
            \abs{x}-0.5 & \text{otherwise}
        \end{cases}
    \end{equation}
\end{enumerate}
We combine the three loss functions with addition divided by the total number of fake visual and audio frames $p=\sum_{m\in\{v,a\}}\sum_{\phi=1}^{f}\mathds{1}_{m,\phi}$:
\begin{equation}   \mathfrak{L}_{TFL}=\big(\mathfrak{L}_{f}+\mathfrak{L}_{d}+\mathfrak{L}_1^{s}\big)/p
\end{equation}
We finally average both $\mathfrak{L}_{DFD}$ and $\mathfrak{L}_{TFL}$ across batch samples during training.

\section{Dataset class and split sizes}
\Cref{tab:splits} provides details on dataset class and split sizes. Although, FakeAVCeleb paper \cite{khalid2021fakeavceleb} reports 10,000 FVFA and 9,000 FVRA samples, the download from the corresponding homepage \url{https://sites.google.com/view/fakeavcelebdash-lab} contains 10,835 and 9,709 respectively, resulting in 21.5K total number of samples (incl. 500 RVFA \& 500 RVRA) instead of the commonly reported 20,000 size for this dataset. Documentation therein aslo claims \textit{``Since we apply the manual screening process on synthesized videos, the final video count is more than 20,000.''}. Test metadata of AV-Deepfake1M have not been released but evaluation is enabled through Codabench.

\section{Further Ablations}
\label{sec:supp_ablation_dfd}
\begin{figure*}[h]
     \centering
     \begin{subfigure}[b]{0.45\textwidth}
         \centering
         \includegraphics[width=\textwidth]{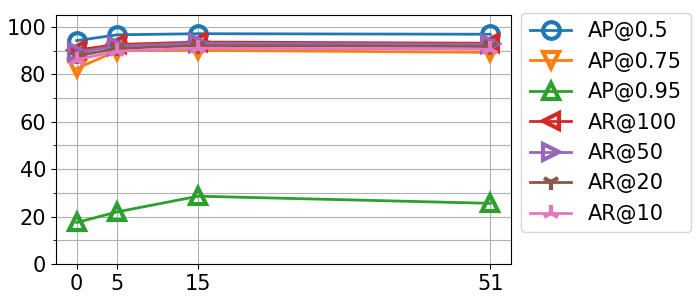}
         \caption{LAV-DF: Window size $q$}
         \label{fig:ablation_tfl_lavdf_win_size}
     \end{subfigure}
    \hfill
     \begin{subfigure}[b]{0.45\textwidth}
         \centering
         \includegraphics[width=\textwidth]{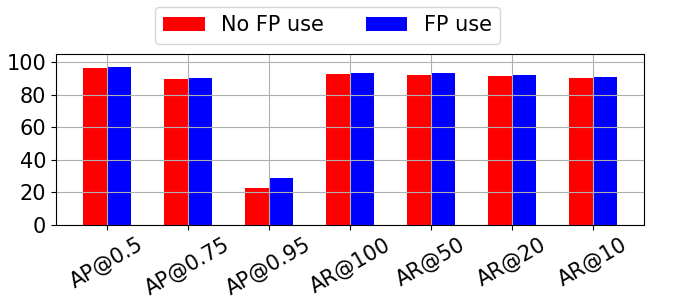}
         \caption{LAV-DF: Feature Pyramid (FP) use}
         \label{fig:ablation_tfl_lavdf_feature_pyramid}
     \end{subfigure}
     \hfill
     \begin{subfigure}[b]{0.55\textwidth}
         \centering
         \includegraphics[width=\textwidth]{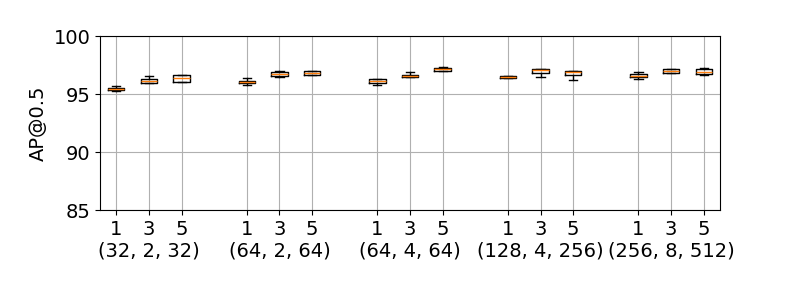}
         \caption{LAV-DF: $\mathcal{T}$'s size; $l$ \& $(d,r,d\cdot u)$}
         \label{fig:ablation_hp_tfl_lavdf}
     \end{subfigure}

    \caption{Ablation and hyperparameter tuning analysis on LAV-DF (TFL).}
    \label{fig:ablation_lavdf_corresponding}
\end{figure*}

\begin{figure*}
     \centering

     \begin{subfigure}[b]{0.49\textwidth}
         \centering
         \includegraphics[width=\textwidth]{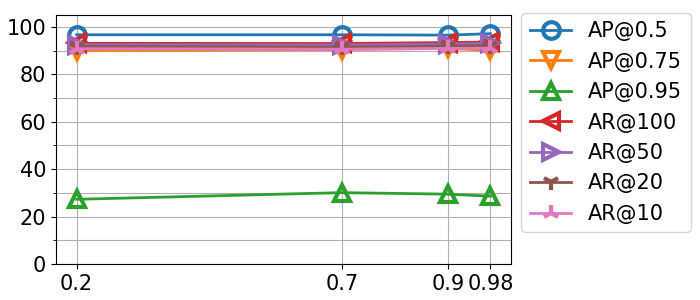}
         \caption{LAV-DF: Focal loss hyperparmeter $\alpha$}
         \label{fig:ablation_tfl_lavdf_alpha}
     \end{subfigure}
     \hfill
     \begin{subfigure}[b]{0.49\textwidth}
         \centering
         \includegraphics[width=\textwidth]{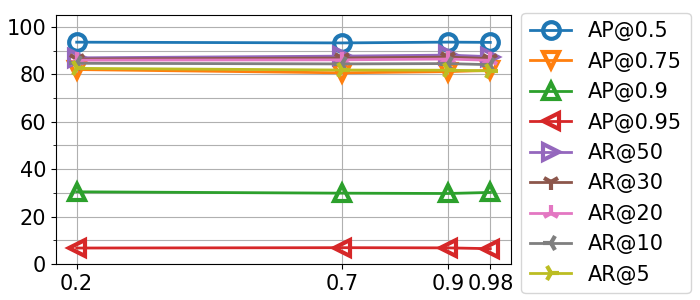}
         \caption{AV-Deepfake1M: Focal loss hyperparmeter $\alpha$}
         \label{fig:ablation_tfl_avdeepfake1m_alpha}
     \end{subfigure}

     \begin{subfigure}[b]{0.49\textwidth}
         \centering
         \includegraphics[width=\textwidth]{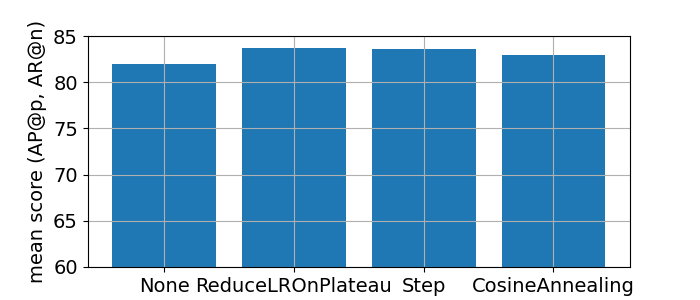}
         \caption{LAV-DF: Learning rate scheduler}
         \label{fig:ablation_schedulers_tfl_lavdf}
     \end{subfigure}
     \hfill
     \begin{subfigure}[b]{0.49\textwidth}
         \centering
         \includegraphics[width=\textwidth]{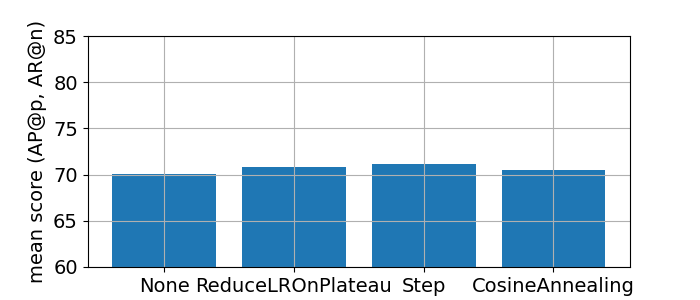}
         \caption{AV-Deepfake1M: Learning rate scheduler}
         \label{fig:ablation_schedulers_tfl_avdeepfake1m}
     \end{subfigure}
    \caption{Ablation and hyperparameter tuning analysis on LAV-DF and AV-Deepfake1M (TFL).}
    \label{fig:ablation_tfl}
\end{figure*}

\begin{figure*}
     \centering
     \begin{subfigure}[b]{0.32\textwidth}
         \centering
         \includegraphics[width=\textwidth]{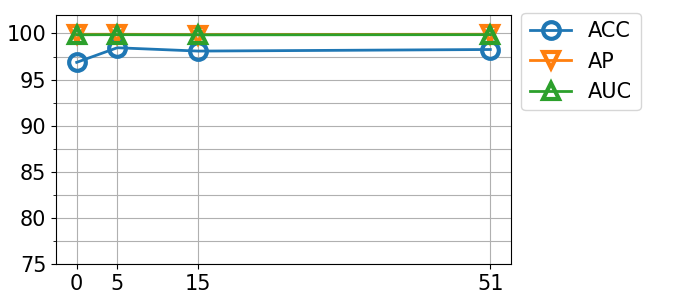}
         \caption{FakeAVCeleb: Window size $q$}
         \label{fig:ablation_dfd_fakeavceleb_win_size}
     \end{subfigure}
     \hfill
     \begin{subfigure}[b]{0.32\textwidth}
         \centering
         \includegraphics[width=\textwidth]{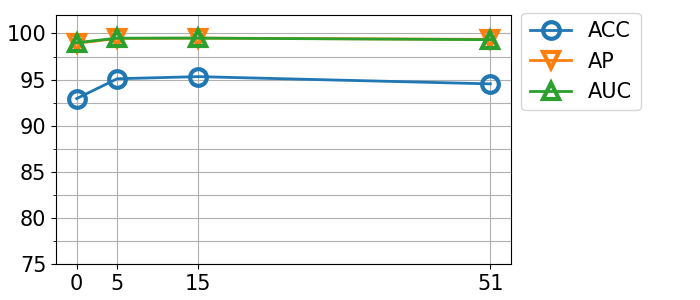}
         \caption{LAV-DF: Window size $q$}
         \label{fig:ablation_dfd_lavdf_win_size}
     \end{subfigure}
     \hfill
     \begin{subfigure}[b]{0.32\textwidth}
         \centering
         \includegraphics[width=\textwidth]{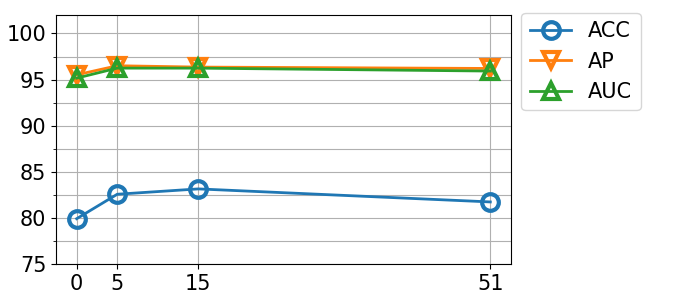}
         \caption{AV-Deepfake1M: Window size $q$}
         \label{fig:ablation_dfd_avdeepfake1m_win_size}
     \end{subfigure}

     \begin{subfigure}[b]{0.32\textwidth}
         \centering
         \includegraphics[width=\textwidth]{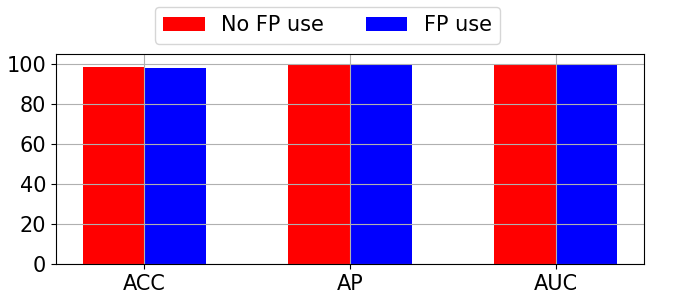}
         \caption{FakeAVCeleb: Feature Pyramid (FP) use}
         \label{fig:ablation_dfd_fakeavceleb_feature_pyramid}
     \end{subfigure}
     \hfill
     \begin{subfigure}[b]{0.32\textwidth}
         \centering
         \includegraphics[width=\textwidth]{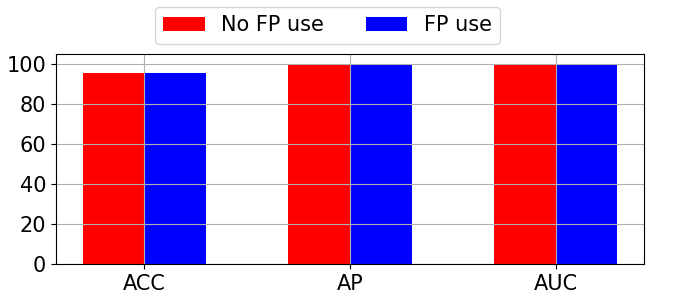}
         \caption{LAV-DF: Feature Pyramid (FP) use}
         \label{fig:ablation_dfd_lavdf_feature_pyramid}
     \end{subfigure}
     \hfill
     \begin{subfigure}[b]{0.32\textwidth}
         \centering
         \includegraphics[width=\textwidth]{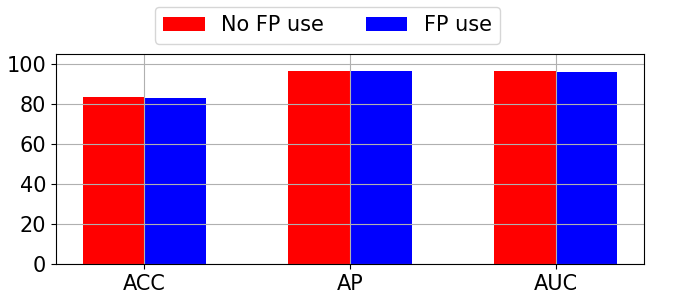}
         \caption{AV-Deepfake1M: Feature Pyramid (FP) use}
         \label{fig:ablation_dfd_avdeepfake1m_feature_pyramid}
     \end{subfigure}

     \begin{subfigure}[b]{0.32\textwidth}
         \centering
         \includegraphics[width=\textwidth]{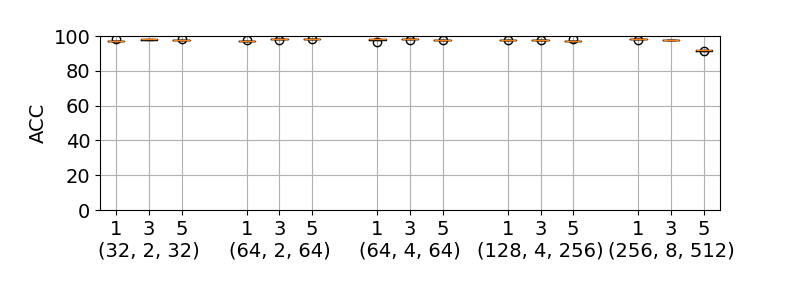}
         \caption{FakeAVCeleb: $\mathcal{T}$'s size; $l$ \& $(d,r,d\cdot u)$}
         \label{fig:ablation_hp_dfd_fakeavceleb}
     \end{subfigure}
     \hfill
     \begin{subfigure}[b]{0.32\textwidth}
         \centering
         \includegraphics[width=\textwidth]{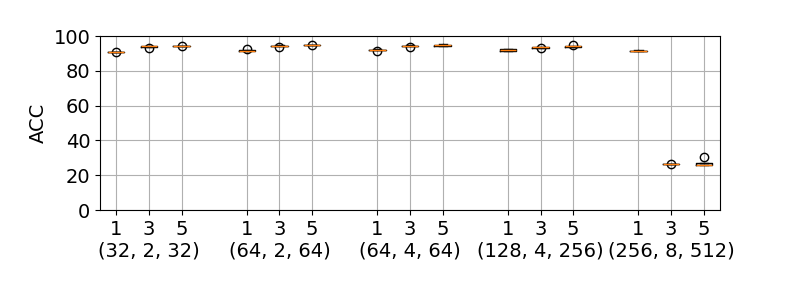}
         \caption{LAV-DF: $\mathcal{T}$'s size; $l$ \& $(d,r,d\cdot u)$}
         \label{fig:ablation_hp_dfd_lavdf}
     \end{subfigure}
     \hfill
     \begin{subfigure}[b]{0.32\textwidth}
         \centering
         \includegraphics[width=\textwidth]{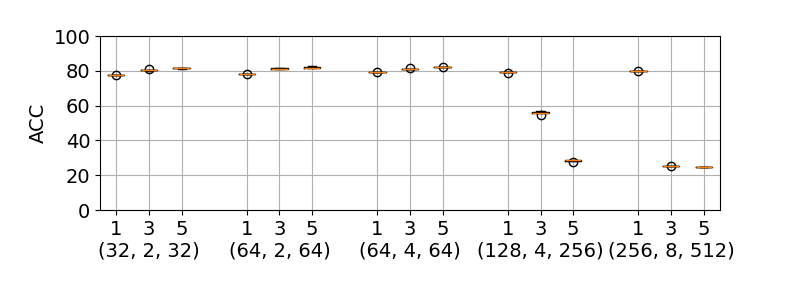}
         \caption{AV-Deepfake1M: $\mathcal{T}$'s size; $l$ \& $(d,r,d\cdot u)$}
         \label{fig:ablation_hp_dfd_avdeepfake1m}
     \end{subfigure}

     \begin{subfigure}[b]{0.32\textwidth}
         \centering
         \includegraphics[width=\textwidth]{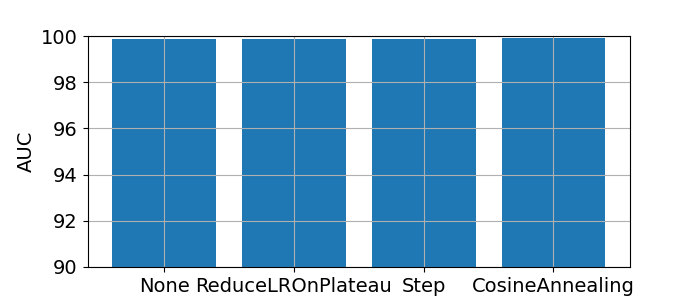}
         \caption{FakeAVCeleb: Learning rate scheduler}
         \label{fig:ablation_schedulers_dfd_fakeavceleb}
     \end{subfigure}
     \hfill
     \begin{subfigure}[b]{0.32\textwidth}
         \centering
         \includegraphics[width=\textwidth]{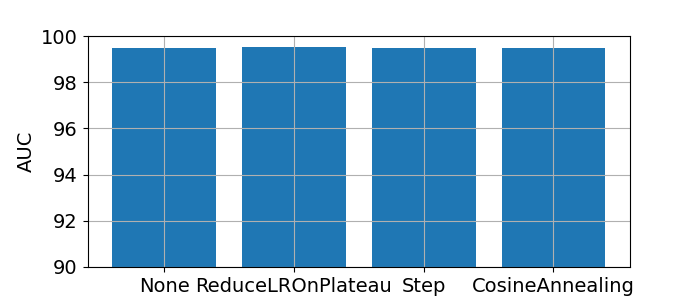}
         \caption{LAV-DF: Learning rate scheduler}
         \label{fig:ablation_schedulers_dfd_lavdf}
     \end{subfigure}
     \hfill
     \begin{subfigure}[b]{0.32\textwidth}
         \centering
         \includegraphics[width=\textwidth]{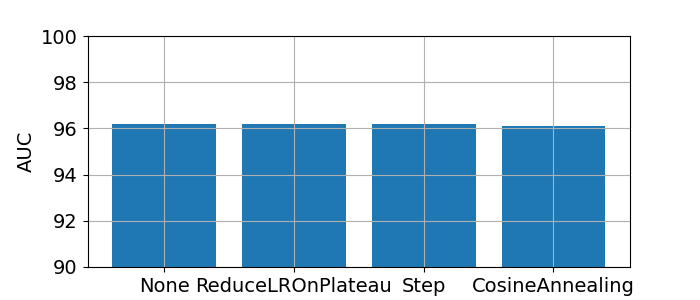}
         \caption{AV-Deepfake1M: Learning rate scheduler}
         \label{fig:ablation_schedulers_dfd_avdeepfake1m}
     \end{subfigure}
    \caption{Ablation and hyperparameter tuning analysis on FakeAVCeleb, LAV-DF,  and AV-Deepfake1M (DFD).}
    \label{fig:ablation_dfd}
\end{figure*}

\Cref{fig:ablation_lavdf_corresponding} presents the ablations on LAV-DF (TFL) corresponding to the ablations presented in the main manuscript on AV-Deepfake1M, namely for window size $q$, feature pyramid use, and Transformer $\mathcal{T}$ size. The conclusions drawn from these ablations on both datasets agree. \Cref{fig:ablation_tfl} then presents further TFL task ablations on both LAV-DF and AV-Deepfake1M. Specifically, 
\Cref{fig:ablation_tfl_lavdf_alpha,fig:ablation_tfl_avdeepfake1m_alpha} indicate little to no sensitivity to the focal loss' hyperparameter $\alpha$. \Cref{fig:ablation_schedulers_tfl_lavdf,fig:ablation_schedulers_tfl_avdeepfake1m} indicate that the use of a learning rate scheduler is beneficial however the type of scheduler is not significantly affecting the results. Next, \Cref{fig:ablation_dfd} presents ablations on FakeAVCeleb, LAV-DF, and AV-Deepfake1M wrt the DFD task. The results indicate that on DFD a window size of $q=15$ is optimal, but performance is not increased with feature pyramids nor with different learning rate schedules. Also, medium-sized models achieve maximum performance.

\section{Interpretability}
\begin{figure*}[htbp]
    \centering
    \begin{subfigure}{0.5\textwidth}
        \centering
        \includegraphics[width=\textwidth]{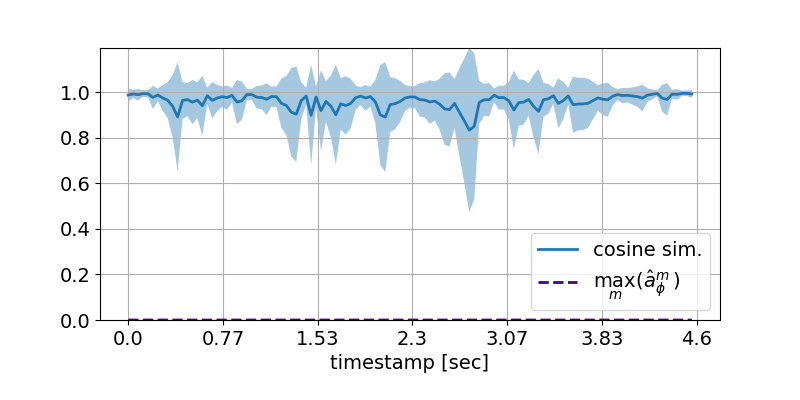}
        \caption{Real \#1: LAV-DF/test/000000.mp4}
    \end{subfigure}\hfill
    \begin{subfigure}{0.5\textwidth}
        \centering
        \includegraphics[width=\textwidth]{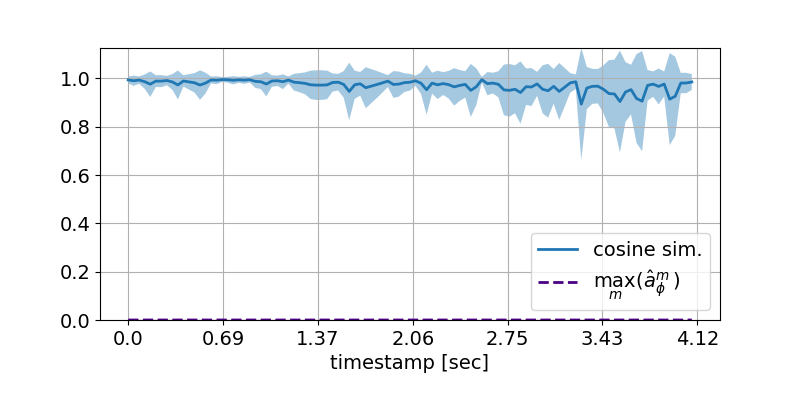}
        \caption{Real \#2: LAV-DF/test/000001.mp4}
    \end{subfigure}\hfill
    \begin{subfigure}{0.5\textwidth}
        \centering
        \includegraphics[width=\textwidth]{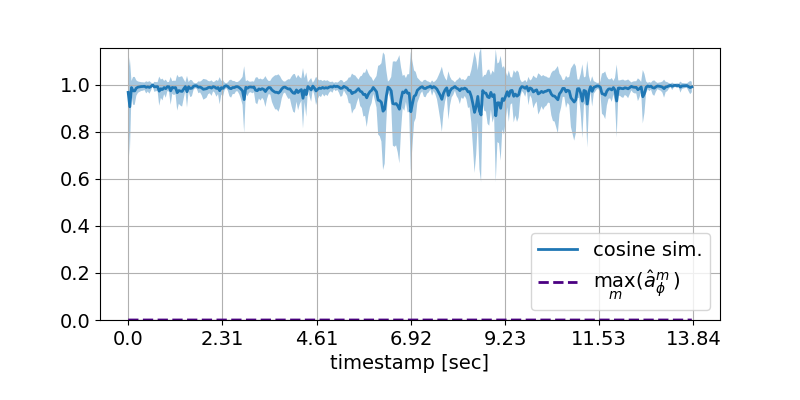}
        \caption{Real \#3: LAV-DF/test/000005.mp4}
    \end{subfigure}\hfill
    \begin{subfigure}{0.5\textwidth}
        \centering
        \includegraphics[width=\textwidth]{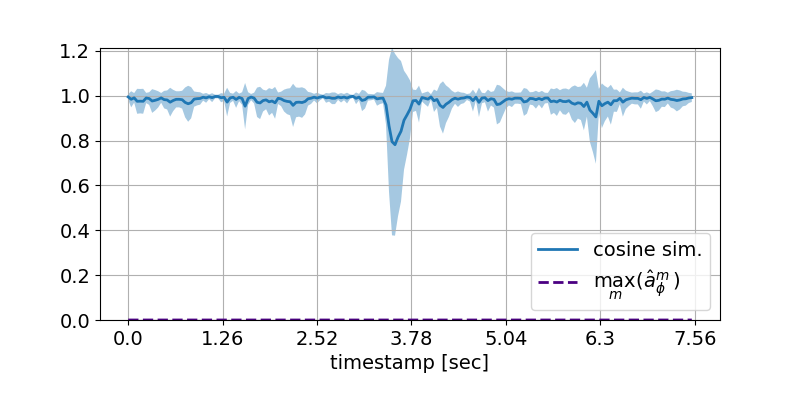}
        \caption{Real \#4: LAV-DF/test/000012.mp4}
    \end{subfigure}\hfill
    \begin{subfigure}{0.5\textwidth}
        \centering
        \includegraphics[width=\textwidth]{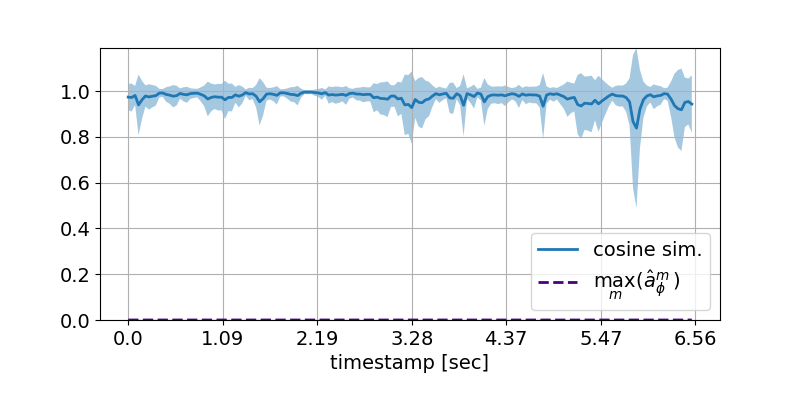}
        \caption{Real \#5: LAV-DF/test/000016.mp4}
    \end{subfigure}\hfill
    \begin{subfigure}{0.5\textwidth}
        \centering
        \includegraphics[width=\textwidth]{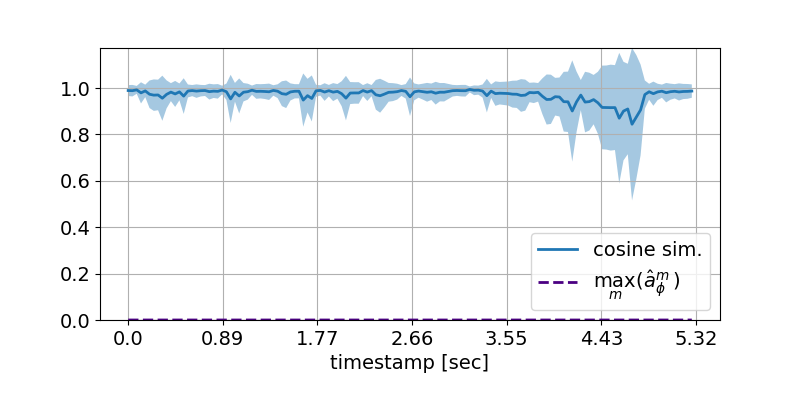}
        \caption{Real \#6: LAV-DF/test/000032.mp4}
    \end{subfigure}\hfill
    \begin{subfigure}{0.5\textwidth}
        \centering
        \includegraphics[width=\textwidth]{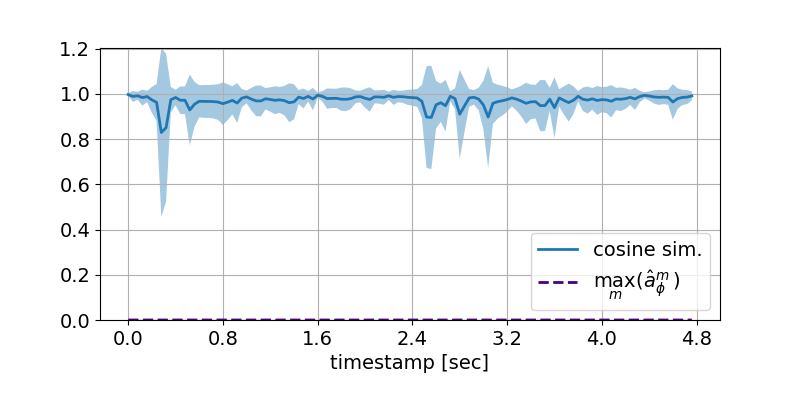}
        \caption{Real \#7: LAV-DF/test/000036.mp4}
    \end{subfigure}\hfill
    \begin{subfigure}{0.5\textwidth}
        \centering
        \includegraphics[width=\textwidth]{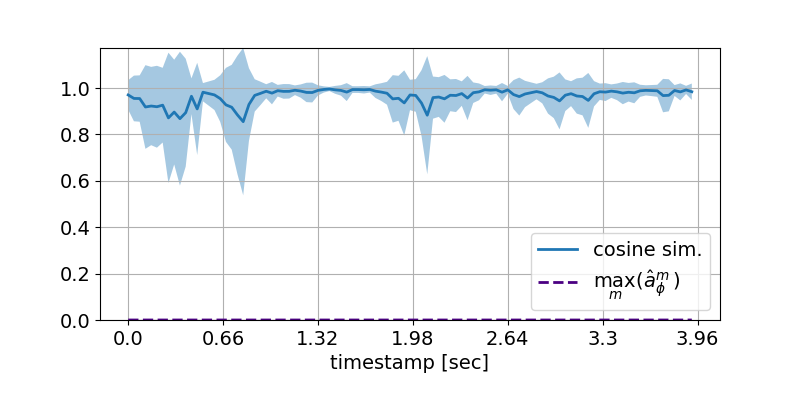}
        \caption{Real \#8: LAV-DF/test/000040.mp4}
    \end{subfigure}
    \caption{Cross-modal cosine similarity on \method's representations of 8 real examples.}
    \label{fig:real_interpret}
\end{figure*}

\begin{figure*}[htbp]
    \centering
    \begin{subfigure}{0.5\textwidth}
        \centering
        \includegraphics[width=\textwidth]{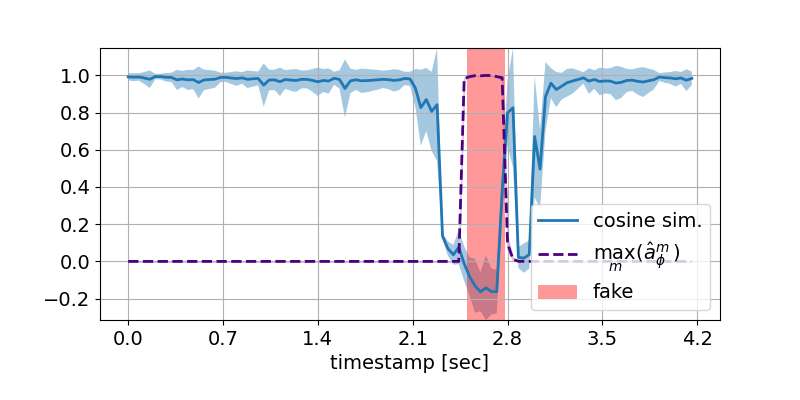}
        \caption{Fake \#1: LAV-DF/test/000003.mp4}
    \end{subfigure}\hfill
    \begin{subfigure}{0.5\textwidth}
        \centering
        \includegraphics[width=\textwidth]{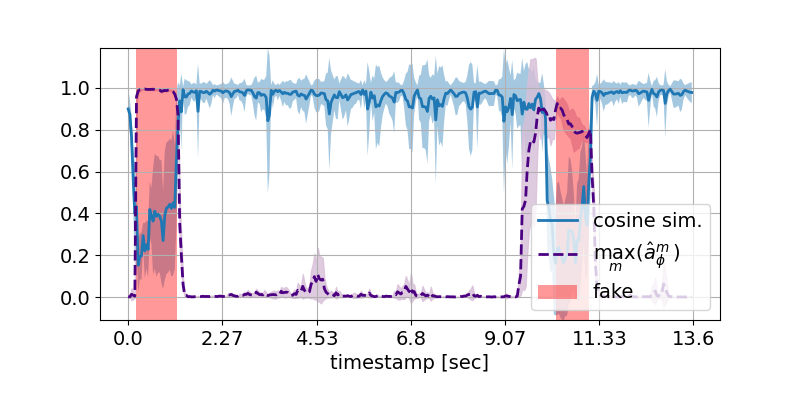}
        \caption{Fake \#2: LAV-DF/test/000008.mp4}
    \end{subfigure}\hfill
    \begin{subfigure}{0.5\textwidth}
        \centering
        \includegraphics[width=\textwidth]{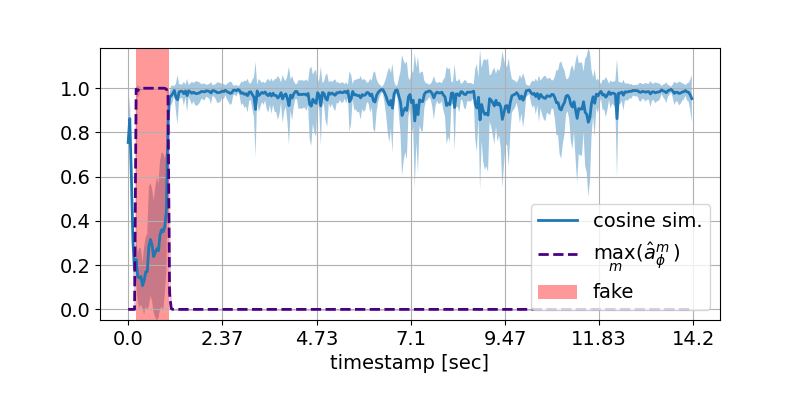}
        \caption{Fake \#3: LAV-DF/test/000011.mp4}
    \end{subfigure}\hfill
    \begin{subfigure}{0.5\textwidth}
        \centering
        \includegraphics[width=\textwidth]{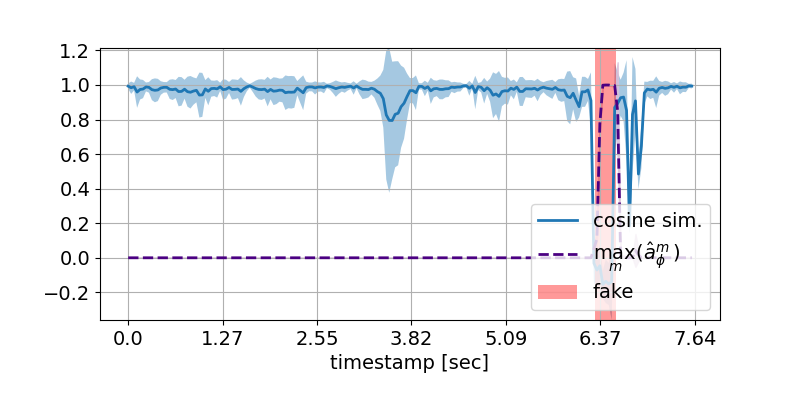}
        \caption{Fake \#4: LAV-DF/test/000014.mp4}
    \end{subfigure}\hfill
    \begin{subfigure}{0.5\textwidth}
        \centering
        \includegraphics[width=\textwidth]{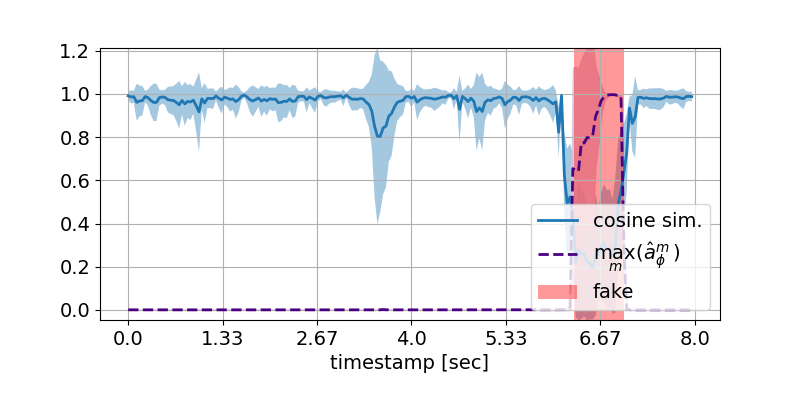}
        \caption{Fake \#5: LAV-DF/test/000015.mp4}
    \end{subfigure}\hfill
    \begin{subfigure}{0.5\textwidth}
        \centering
        \includegraphics[width=\textwidth]{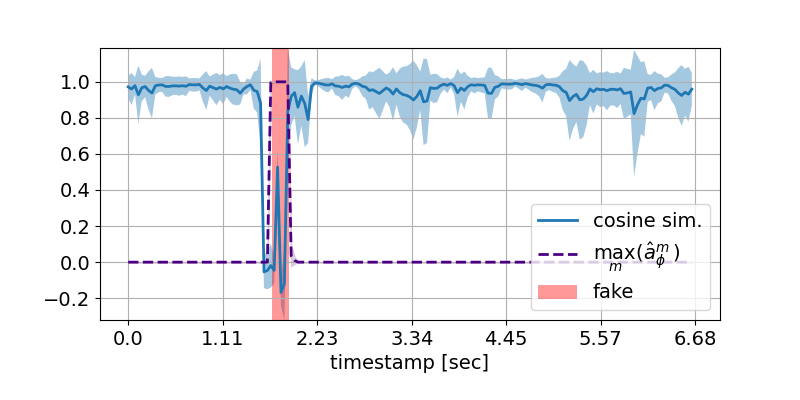}
        \caption{Fake \#6: LAV-DF/test/000018.mp4}
    \end{subfigure}\hfill
    \begin{subfigure}{0.5\textwidth}
        \centering
        \includegraphics[width=\textwidth]{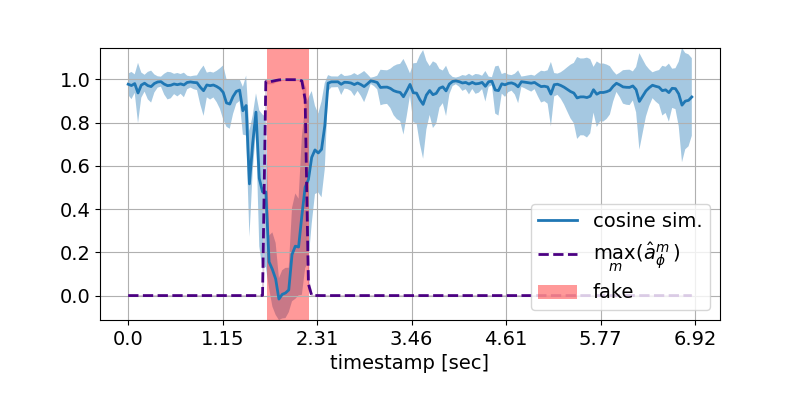}
        \caption{Fake \#7: LAV-DF/test/000019.mp4}
    \end{subfigure}\hfill
    \begin{subfigure}{0.5\textwidth}
        \centering
        \includegraphics[width=\textwidth]{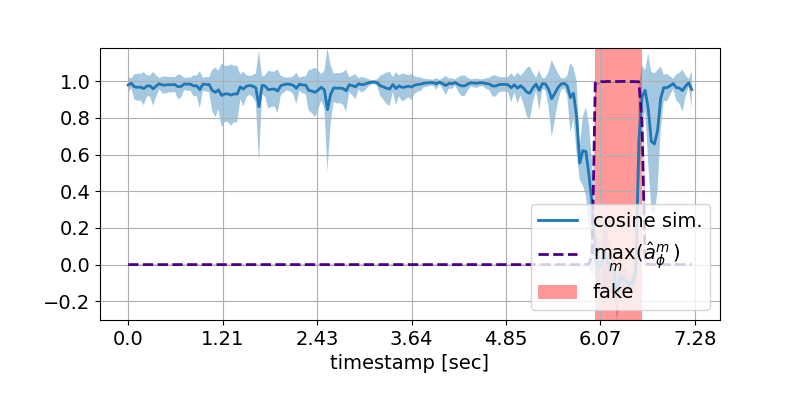}
        \caption{Fake \#8: LAV-DF/test/000030.mp4}
    \end{subfigure}
    \caption{Cross-modal cosine similarity on \method's representations of 8 fake examples.}
    \label{fig:fake_inerpret}
\end{figure*}

\begin{figure*}
    \centering
    \includegraphics[width=0.75\linewidth]{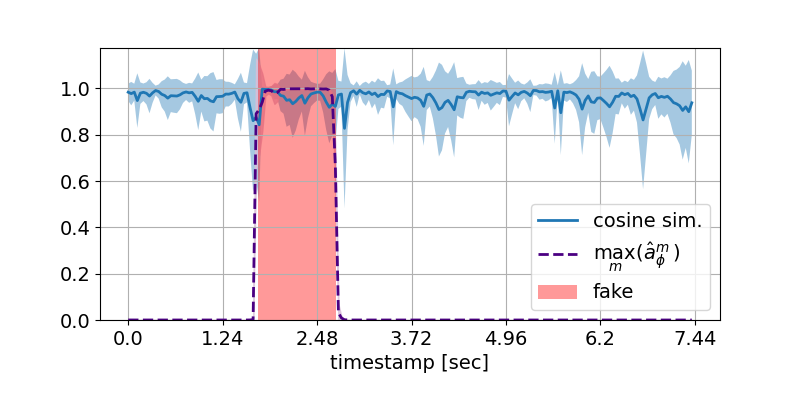}
    \caption{Non-interpretable fake example (LAV-DF/test/000017.mp4).}
    \label{fig:non_interpret}
\end{figure*}

\Cref{fig:real_interpret,fig:fake_inerpret} present further interpretability plots for 8 real and 8 fake video samples from LAV-DF. Cosine similarity at the frame-level reveals \method's decision reasoning in these cases. However, there are some rare cases in which \method\ correctly identified the modified parts although cross-modal similarity is high, such as in the example presented in \Cref{fig:non_interpret}. In such cases \method\ is forced to successfully identify alternative manipulation features sacrificing interpretability.

\section{Robustness}
\begin{figure*}
    \centering
    \includegraphics[width=0.95\textwidth]{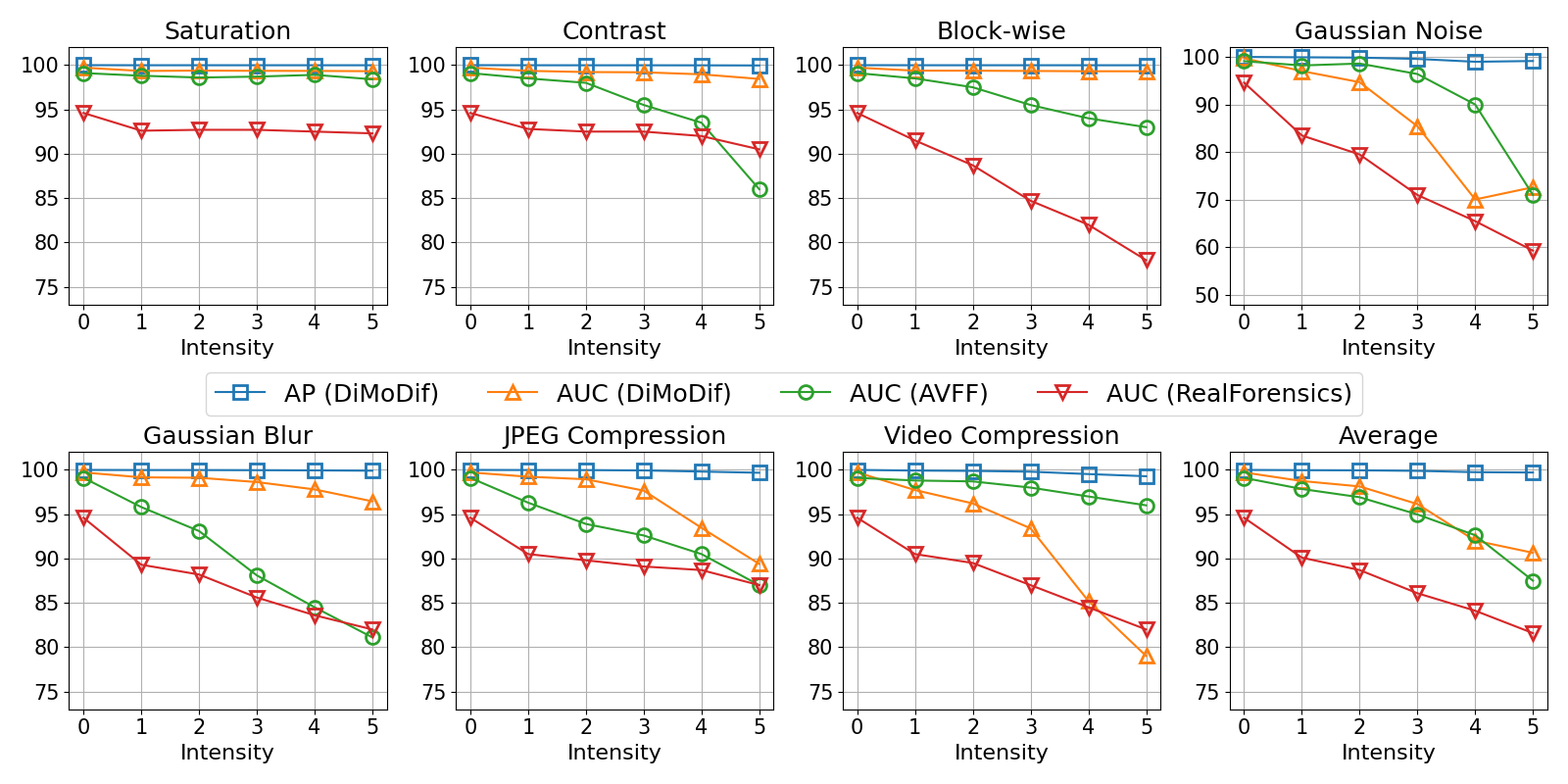}
    \caption{Robustness to unseen visual perturbations. AVFF and RealForensics performance taken from \cite{oorloff2024avff}'s supplementary material.}
    \label{fig:robustness_visual}
\end{figure*}

\textbf{Robustness to visual distortions}: \Cref{fig:robustness_visual} illustrates the performance of \method\ under unseen visual perturbations (our model has been trained without any visual augmentations) commonly present in real-world scenarios. We follow the evaluation protocol of \cite{oorloff2024avff,feng2023self,haliassos2021lips} which consider the FakeAVCeleb test set and directly source the implementations of visual perturbations from \url{https://github.com/EndlessSora/DeeperForensics-1.0}. Specifically, videos are subject to \texttt{saturation}, \texttt{contrast}, \texttt{block-wise}, \texttt{Gaussian noise}, \texttt{Gaussian blur}, \texttt{JPEG compression}, and \texttt{video compression}, while we also compute the average robustness across perturbations. \method\ clearly outperforms RealForensics \cite{haliassos2022leveraging} under all perturbation scenarios except \texttt{video compression} at intensity levels 4 and 5. Also, our method outperforms AVFF \cite{oorloff2024avff} under \texttt{saturation}, \texttt{contrast}, \texttt{block-wise}, \texttt{Gaussian blur}, and \texttt{JPEG compression} at all levels of intensity, and is outperformed by AVFF under \texttt{Gaussian noise} and \texttt{video compression} at all levels of intensity. Wrt average AUC \method\ slightly outperforms AVFF and is much more robust than RealForensics. Notably, \method\ exhibits exceptional robustness wrt AP on average achieving 99.7 at intensity level 5, indicating near-perfect detection of positive samples despite significant visual distortions. In approximately 0.06\% of cases, high-intensity \texttt{Gaussian noise} (level 5) prevented landmark detection, making lip-reading impossible and thus precluding the application of our method.

\begin{figure*}
    \centering
    \includegraphics[width=0.75\textwidth]{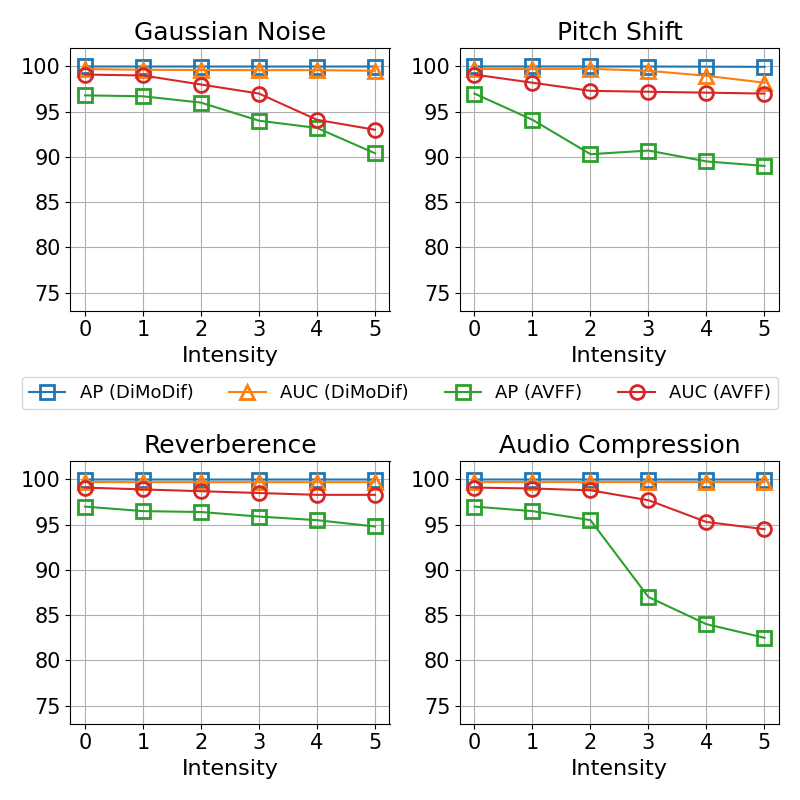}
    \caption{Robustness to unseen audio perturbations. AVFF performance taken from \cite{oorloff2024avff}'s supplementary material.}
    \label{fig:robustness_audio}
\end{figure*}

\textbf{Robustness to audio distortions}: \Cref{fig:robustness_audio} illustrates the performance of \method\ under unseen audio perturbations (our model has been trained without any audio augmentations) commonly present in real-world scenarios. We follow the evaluation protocol of \cite{oorloff2024avff} that considers the FakeAVCeleb test set and implement the audio perturbations using \textit{torchaudio}, \textit{pysndfx}, and \textit{pydub} Python libraries as indicated therein. Specifically, the videos are subject to \texttt{Gaussian noise}, \texttt{pitch shift}, \texttt{reverberence}, and \texttt{audio compression} at 5 levels of intensity. \method\ achieves unprecedented levels of  robustness to audio perturbations with 99.3 average AUC across all perturbations at intensity level 5. This fact indicates that audio speech recognition is not affected by such perturbations and the corresponding features are effective under any perturbation level scenario. In addition, \method\ outperforms AVFF under all perturbation scenarios and at all intensity levels.

\textbf{Why is \method\ highly robust to unseen content distortions?} \Cref{tab:transcript_robustness} demonstrates the resilience of \method's text prediction backbones to various content distortions. Notably, transcripts remained largely unaffected across most distortion-level combinations. Video distortions, specifically high-intensity \texttt{video compression}, \texttt{JPEG compression}, and \texttt{Gaussian noise}, had the most significant impact on text prediction, mirroring the performance declines observed in \Cref{fig:robustness_visual}. Similarly, only high-intensity \texttt{pitch shift} audio distortions reduced performance, consistent with the trends in \Cref{fig:robustness_audio}. This association underscores the critical role of the video and audio speech recognition representations in \method's overall robustness.

\section{Generalization}
We evaluate \method\ (trained on FakeAVCeleb, LAV-DF, and AV-Deepfake1M) on the test set of DFDC \cite{dolhansky2020deepfake} and a randomly selected 6K-sized sample from KoDF \cite{kwon2021kodf} proportionally spanning all synthesis methods and real samples.  

\Cref{tab:generalization_dfd} presents the corresponding results. Training on FakeAV-Celeb \cite{khalid2021fakeavceleb} leads to poor performance on DFDC but to high performance on KoDF with 98.8 AP. Training on LAV-DF \cite{cai2022you} leads to almost random performance on DFDC but high performance on KoDF with 97.9 AP. Training on AV-Deepfake1M \cite{cai2024av} dataset leads to random performance on DFDC but high performance on KoDF with 98.9 AP. In general, very high generalization performance is observed on KoDF, although Korean are not recognized by the (Ma 2022) backbones, indicating the learning of cross-language features by \method. Also, low generalization is observed on DFDC, which can be partly explained by the discussion in Section 2.2. Our models focus on cross-modal speech-related information differences, while many DFDC videos contain multiple talking persons\footnote{E.g., \texttt{DFDC/test/aalyqplqns.mp4}}, moving persons\footnote{E.g., \texttt{DFDC/test/chyfcacfjr.mp4}} or wide shots\footnote{E.g., \texttt{DFDC/test/ljnjhvpnhq.mp4"}} rendering lip-reading hard, do not contain talking\footnote{E.g., \texttt{DFDC/test/aarpyivfys.mp4}} or the talking person is different from the depicted one\footnote{E.g., \texttt{DFDC/test/mfjljhqhca.mp4}}; thus, visual-only face manipulation detectors lead to better generalization in this dataset.

\Cref{tab:generalization_tfl} presents cross-dataset performance on TFL wrt all AP@\{0.5, 0.75, 0.90, 0.95\} and AR@\{100, 50, 30, 20, 10, 5\} metrics. Cross-dataset performance on TFL is shown to be much more challenging than on DFD. Also, the results are even worse when training on LAV-DF in contrast to AV-Deepfake1M which contains half-sized fake parts, thus presenting a more challenging objective, while also being much larger. 

\Cref{tab:generalization_fakeavceleb_dfd} provides a more detailed generalization evaluation on the FakeAVCeleb dataset with scores per forgery type, indicating the effectiveness of training on AV-Deepfake1M reaching 99.2 AP \& 95.5 AUC (AVG-FV)  and 93.9 AP \& 95.5 AUC (RVFA). Training on LAV-DF also results in good generalization reaching 98.5 AP \& 91.2 AUC (AVG-FV) and 81.3 AP \& 76.8 AUC (RVFA). Note, that this is not a cross-manipulation analysis as the models trained on LAV-DF and AV-Deepfake1M datasets have seen all manipulation types (RVRA, RVFA, FVFA, FVRA). Thus, metrics of the 2nd and 3rd rows are comparable but the 1st row (showing cross-manipulation performance) is only provided for reference and is not comparable with the other two.

\section{Real-world analysis}
\Cref{tab:real_world_info} provides further information on the set of videos used for the real-world analysis. Specifically, we provide the public URL (if exists), the spoken language, the duration of the video, the processing (by \method) time, and the Coverage score $\mathtt{c}$. Language has been automatically retrieved through \url{https://speechbrain.github.io/}. Finally, 16 of the videos do not have public URLs.

\section{Error analysis}
\begin{figure*}[htbp]
    \centering
    \begin{subfigure}{0.5\textwidth}
        \centering
        \includegraphics[width=\textwidth]{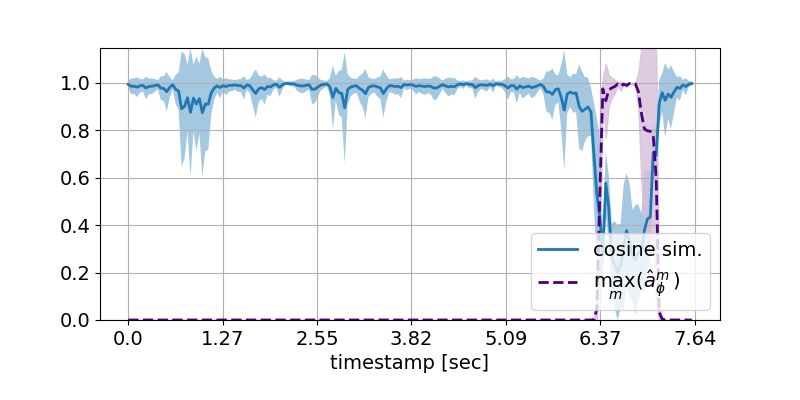}
        \caption{Real predicted as fake: LAV-DF/test/000020.mp4}
        \label{subfig:real}
    \end{subfigure}\hfill
    \begin{subfigure}{0.5\textwidth}
        \centering
        \includegraphics[width=\textwidth]{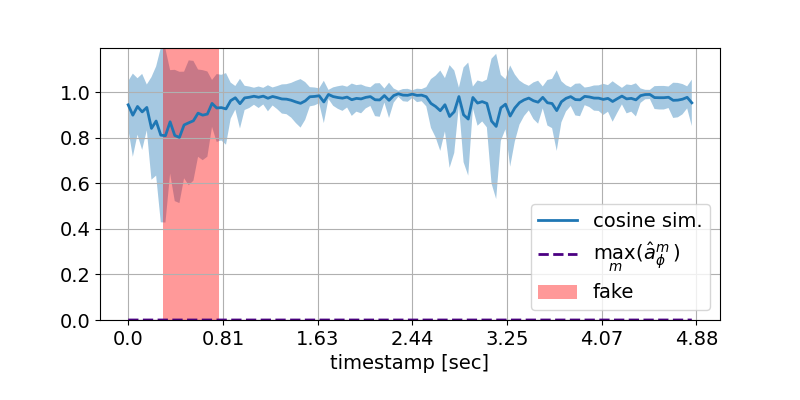}
        \caption{Fake predicted as real: LAV-DF/test/000038.mp4}
        \label{subfig:fake}
    \end{subfigure}
    \caption{Example videos that \method\ has misclassified.}
    \label{fig:errors}
\end{figure*}

\Cref{fig:errors} presents two examples of \method's misclassifications. Specifically, in the real video shown in \cref{subfig:real}, a non-verbal vocalization (panting imitation) between 6.3s and 7.3s is incorrectly interpreted as inconsistency, leading \method\ to assign a high fake probability to those frames. In contrast, \cref{subfig:fake} features a visual-only manipulated segment from 0.3s to 0.78s; here, despite a slight reduction in cross-modal inconsistency, the high fidelity of the utterance deceives the model.

\begin{table*}[th]
    \centering
     \resizebox{\textwidth}{!}{
    \begin{tabular}{l llll llll llll}
        \toprule
        &\multicolumn{4}{c}{FakeAVCeleb} & \multicolumn{4}{c}{LAV-DF} & \multicolumn{4}{c}{AV-Deepfake1M} \\
        \cmidrule(lr){2-5} \cmidrule(lr){6-9} \cmidrule(lr){10-13}
         & train & val & test &\cellcolor{gray!20} total & train & val & test &\cellcolor{gray!20} total & train & val & test &\cellcolor{gray!20} total \\
         \midrule
        FVFA & 6,996 & 561 & 3,278 &\cellcolor{gray!20} 10,835 & 19,090 & 7,701 & 6,369 &\cellcolor{gray!20} 33,160 & 186,344 & 14,515 & - &\cellcolor{gray!20} - \\
        FVRA & 6,363 & 470 & 2,876 &\cellcolor{gray!20} 9,709 & 19,271 & 7,820 & 6,452 &\cellcolor{gray!20} 33,543 & 186,597 & 14,304 & - &\cellcolor{gray!20} - \\
        RVFA & 326 & 24 & 150 &\cellcolor{gray!20} 500 & 19,088 & 7,709 & 6,373 &\cellcolor{gray!20} 33,170 & 186,573 & 14,286 & - &\cellcolor{gray!20} -\\
        RVRA & 13,689 & 22 & 160 &\cellcolor{gray!20} 13,871 & 21,254 & 8,271 & 6,906 &\cellcolor{gray!20} 36,431 & 186,666 & 14,235 & - &\cellcolor{gray!20} - \\
        \cellcolor{gray!20}total &\cellcolor{gray!20} 27,374 &\cellcolor{gray!20} 1,077 &\cellcolor{gray!20} 6,464 & \cellcolor{gray!40} 34,915 &\cellcolor{gray!20} 78,703 &\cellcolor{gray!20} 31,501 &\cellcolor{gray!20} 26,100 & \cellcolor{gray!40} 136,304 &\cellcolor{gray!20} 746,180 &\cellcolor{gray!20} 57,340 &\cellcolor{gray!20} 343,240 & \cellcolor{gray!40} 1,146,760 \\
        \bottomrule
    \end{tabular}
    }
    \caption{Dataset class and split sizes. F: fake, R: real, V: video, A: audio.}
    \label{tab:splits}
\end{table*}

\onecolumn
\twocolumn
\onecolumn
\begin{longtable}{p{0.6cm}p{0.5cm}p{15cm}}
\caption{Text prediction robustness of \method's feature extraction backbones (Ma 2022) \cite{ma2022visual}, to visual and audio distortions for the real video sample: \texttt{FakeAVCeleb/RealVideo-RealAudio/Caucasian (European)/men/id00055/00120.mp4}. First row presents predictions on the original video while the next ones use versions of it with visual or audio distortions as input. $\mathcal{V}$: transcript from visual input, $\mathcal{A}$: transcript from audio input.}
\label{tab:transcript_robustness}

\\
\hline
\textbf{Dist.} & \textbf{Lev.} & \textbf{Transcript} \\ \hline
\multirow{2}{*}{-} & \multirow{2}{*}{-} &$\mathcal{V}$: you know i played with growing up and always had great ability great talent from a young age \\ 
& &$\mathcal{A}$: know i played with growing up and he always had great ability great talent from a young age\\
\hline
\multicolumn{3}{c}{\textbf{Visual distortions}} \\ \hline
\multirow{5}{*}{\begin{sideways}Saturation\end{sideways}} & 1 &$\mathcal{V}$: you know i played with growing up and always had great ability great talent from a young age \\ \cline{2-3}
 & 2 &$\mathcal{V}$: how i played with growing up and always had great ability great talent from a young age \\ \cline{2-3}
 & 3 &$\mathcal{V}$: you know i played with growing up and always had great ability great talent from a young age \\ \cline{2-3}
 & 4 &$\mathcal{V}$: yeah i played with growing up and always had great ability great talent from a young age \\ \cline{2-3}
 & 5 &$\mathcal{V}$: yeah i played with growing up and always had great ability great talent from a young age \\ \hline
\multirow{6}{*}{\begin{sideways}Contrast\end{sideways}} & 1 &$\mathcal{V}$: yeah i played with growing up and always had a great ability great talent from a young age \\ \cline{2-3}
 & 2 &$\mathcal{V}$: how i played with growing up and always had great ability great talent from a young age \\ \cline{2-3}
 & 3 &$\mathcal{V}$: played with growing up and always had great ability great talent from a young age \\ \cline{2-3}
 & 4 &$\mathcal{V}$: you know i played with growing up and always had great ability great talent from a young age \\ \cline{2-3}
 & 5 &$\mathcal{V}$: yeah i played with harold growing up and he always had great ability great talent from a young age \\ \hline
\multirow{5}{*}{\begin{sideways}Block-wise\end{sideways}} & 1 &$\mathcal{V}$: yeah i played with growing up and always had great ability great talent from a young age \\ \cline{2-3}
 & 2 &$\mathcal{V}$: yeah i played with growing up and always had great ability great talent from a young age \\ \cline{2-3}
 & 3 &$\mathcal{V}$: yeah i played with growing up and always had great ability great talent from a young age \\ \cline{2-3}
 & 4 &$\mathcal{V}$: yeah i played with growing up and always had great ability great talent from a young age \\ \cline{2-3}
 & 5 &$\mathcal{V}$: yeah i played with growing up and always had great ability great talent from a young age \\ \hline
\multirow{5}{*}{\begin{sideways}Gauss. noise\end{sideways}} & 1 &$\mathcal{V}$: i played with carol growing up and he always had a great ability great talent from a young age \\ \cline{2-3}
 & 2 &$\mathcal{V}$: i played with her growing up and it was a great ability great talent from a young age \\ \cline{2-3}
 & 3 &$\mathcal{V}$: everybody was scared of growing up with this great ability great talent from a young age \\ \cline{2-3}
 & 4 &$\mathcal{V}$: our plan was to start growing up and you can see how i beat the grass down from the high \\ \cline{2-3}
 & 5 &$\mathcal{V}$: - \\ \hline
\multirow{7}{*}{\begin{sideways}Gauss. blur\end{sideways}} & 1 &$\mathcal{V}$: i played with growing up and always had great ability great talent from a young age \\ \cline{2-3}
 & 2 &$\mathcal{V}$: how i played with carol growing up and he always had a great ability great talent from a young age \\ \cline{2-3}
 & 3 &$\mathcal{V}$: how i played with carol growing up and he always has a great ability great talent from a young age \\ \cline{2-3}
 & 4 &$\mathcal{V}$: how i played with carol growing up he always has great ability great talent from a young age \\ \cline{2-3}
 & 5 &$\mathcal{V}$: now i played with carol growing up he always had great ability great talent from a young age \\ \hline
\multirow{6}{*}{\begin{sideways}JPEG compr.\end{sideways}} & 1 &$\mathcal{V}$: you know i played with growing up and always had great ability great talent from a young age \\ \cline{2-3}
 & 2 &$\mathcal{V}$: i played with growing up and always had great ability great talent from a young age \\ \cline{2-3}
 & 3 &$\mathcal{V}$: now i'm here with the care of growing up and i always like to see the great ability to raise hands for without hands \\ \cline{2-3}
 & 4 &$\mathcal{V}$: i will play with you the character i'll be happy once i turn to you i'll play you as i grew up in high \\ \cline{2-3}
 & 5 &$\mathcal{V}$: there were pictures of his car flipped up so he said to the red bricks at the right angle that's \\ \hline
\multirow{9}{*}{\begin{sideways}Video compr.\end{sideways}} & 1 &$\mathcal{V}$: yeah i played with growing up and always had great ability great talent from a young age \\ \cline{2-3}
 & 2 &$\mathcal{V}$: how i played with how i've grown up and always had the right ability the right talent from a young age \\ \cline{2-3}
 & 3 &$\mathcal{V}$: a little bit we've now grown up and we're starting to create building a great town from a young age \\ \cline{2-3}
 & 4 &$\mathcal{V}$: now i'll be able to fire from the camera you know adjust it to the front but the right hand will be high \\ \cline{2-3}
 & 5 &$\mathcal{V}$: i'll make sure it's just a narrow volume count it will take you to the right place and the right hand will be high \\ \hline
\multicolumn{3}{c}{\textbf{Audio distortions}} \\ \hline
\multirow{5}{*}{\begin{sideways}Gauss. noise\end{sideways}} & 1 &$\mathcal{A}$: know i played with growing up and he always had great ability great talent from a young age \\ \cline{2-3}
 & 2 &$\mathcal{A}$: know i played with growing up and he always had great ability great talent from a young age \\ \cline{2-3}
 & 3 &$\mathcal{A}$: know i played with growing up and he always had great ability great talent from a young age \\ \cline{2-3}
 & 4 &$\mathcal{A}$: know i played with growing up and he always had great ability great talent from a young age \\ \cline{2-3}
 & 5 &$\mathcal{A}$: know i played with growing up and he always had great ability great talent from a young age \\ \hline
\multirow{5}{*}{\begin{sideways}Pitch shift\end{sideways}} & 1 &$\mathcal{A}$: know i played with growing up and he always had great ability great talent from a young age \\ \cline{2-3}
 & 2 &$\mathcal{A}$: yeah i played with growing up and he always had great ability great talent from a young age \\ \cline{2-3}
 & 3 &$\mathcal{A}$: you know papers cars growing up and it was a great time for the young age \\ \cline{2-3}
 & 4 &$\mathcal{A}$: yeah played of gaff growing up and he was a great and it's a great kind of for a young age \\ \cline{2-3}
 & 5 &$\mathcal{A}$: - \\ \hline
\multirow{5}{*}{\begin{sideways}Reverberence\end{sideways}} & 1 &$\mathcal{A}$: know i played with growing up and he always had great ability great talent from a young age \\ \cline{2-3}
 & 2 &$\mathcal{A}$: know i played with growing up and he always had great ability great talent from a young age \\ \cline{2-3}
 & 3 &$\mathcal{A}$: know i played with growing up and he always had great ability great talent from a young age \\ \cline{2-3}
 & 4 &$\mathcal{A}$: know i played with growing up and he always had great ability great talent from a young age \\ \cline{2-3}
 & 5 &$\mathcal{A}$: know i played with growing up and he always had great ability great talent from a young age \\ \hline
\multirow{5}{*}{\begin{sideways}Audio compr.\end{sideways}} & 1 &$\mathcal{A}$: know i played with growing up and he always had great ability great talent from a young age \\ \cline{2-3}
 & 2 &$\mathcal{A}$: know i played with growing up and he always had great ability great talent from a young age \\ \cline{2-3}
 & 3 &$\mathcal{A}$: know i played with growing up and he always had great ability great talent from a young age \\ \cline{2-3}
 & 4 &$\mathcal{A}$: know i played with growing up and he always had great ability great talent from a young age \\ \cline{2-3}
 & 5 &$\mathcal{A}$: know i played with growing up and he always had great ability great talent from a young age \\ \hline
\end{longtable}
\twocolumn

\begin{table*}[t]
    \centering
    % \resizebox{0.45\textwidth}{!}{
    \begin{tabular}{llllll}
    \toprule
    &&\multicolumn{4}{c}{Test dataset}\\
    \cmidrule{3-6}
    &&\multicolumn{2}{c}{DFDC \cite{dolhansky2020deepfake}}&\multicolumn{2}{c}{KoDF \cite{kwon2021kodf}}\\
    \cmidrule(lr){3-4}\cmidrule(lr){5-6}
    &&AP&AUC&AP&AUC\\
    \midrule
\multirow{3}{*}{\rotatebox[origin=c]{90}{\shortstack[c]{Training\\dataset}}}&FakeAVCeleb \cite{khalid2021fakeavceleb}&60.8&58.9&98.8&69.9\\[2pt]
&LAV-DF \cite{cai2022you}&52.9&54.5&97.9&56.7\\[2pt]
&AV-Deepfake1M \cite{cai2024av}&50.5&49.8&98.9&72.8\\
\bottomrule
    \end{tabular}
    % }
    \caption{Generalization on Deepfake detection task.}
    \label{tab:generalization_dfd}
\end{table*}

\begin{table*}[t]
\centering
\resizebox{\textwidth}{!}{
\begin{tabular}{llllllllllll}
\toprule
&&AP@0.5&AP@0.75&AP@0.9&AP@0.95&AR@100&AR@50&AR@30&AR@20&AR@10&AR@5\\
\cmidrule{1-12}
&&\multicolumn{10}{c}{Test dataset: LAV-DF}\\
\cmidrule{2-12}
\multirow{5}{*}{\rotatebox[origin=c]{90}{\shortstack[c]{Training dataset}}}&\textcolor{gray}{LAV-DF} \cite{cai2022you}&\textcolor{gray}{95.5}&\textcolor{gray}{87.9}&\textcolor{gray}{54.6}&\textcolor{gray}{20.6}&\textcolor{gray}{94.2}&\textcolor{gray}{93.7}&\textcolor{gray}{93.2}&\textcolor{gray}{92.7}&\textcolor{gray}{91.4}&\textcolor{gray}{89.2}\\[2pt]
&AV-Deepfake1M \cite{cai2024av}&59.8&26.2&2.3&0.2&75.3&72.3&69.7&67.3&62.5&56.2\\
\cmidrule{2-12}
&&\multicolumn{10}{c}{Test dataset: AV-Deepfake1M}\\
\cmidrule{2-12}
&LAV-DF \cite{cai2022you}&16.8&4.6&0.3&0.0&31.9&30.1&28.6&27.0&23.6&19.9\\[2pt]
&\textcolor{gray}{AV-Deepfake1M} \cite{cai2024av}&\textcolor{gray}{96.0}&\textcolor{gray}{86.2}&\textcolor{gray}{34.5}&\textcolor{gray}{6.9}&\textcolor{gray}{89.2}&\textcolor{gray}{88.7}&\textcolor{gray}{88.2}&\textcolor{gray}{87.7}&\textcolor{gray}{86.4}&\textcolor{gray}{84.3}\\
\bottomrule
\end{tabular}
}
\caption{Generalization performance on Temporal Forgery Localization task.}
\label{tab:generalization_tfl}
\end{table*}

\begin{table*}[]
    \centering
    % \resizebox{\textwidth}{!}{
    \begin{tabular}{llllllllllllll}
    \toprule
    &\multicolumn{12}{c}{Category}\\
        \cmidrule{2-13}
        Training dataset&  \multicolumn{2}{c}{RVFA}&  \multicolumn{2}{c}{FVRA-WL}&  \multicolumn{2}{c}{FVFA-WL}&  \multicolumn{2}{c}{FVFA-FS}&  \multicolumn{2}{c}{FVFA-GAN}&\multicolumn{2}{c}{AVG-FV}\\
        \cmidrule(lr){2-3}\cmidrule(lr){4-5}\cmidrule(lr){6-7}\cmidrule(lr){8-9}\cmidrule(lr){10-11}\cmidrule(lr){12-13}
        & AP&AUC& AP&AUC& AP&AUC& AP&AUC& AP&AUC& AP&AUC\\
    \midrule
     \textcolor{gray}{FakeAVCeleb} \cite{khalid2021fakeavceleb}&\textcolor{gray}{66.4}&\textcolor{gray}{51.6}&\textcolor{gray}{100.}&\textcolor{gray}{99.8}&\textcolor{gray}{100.}&\textcolor{gray}{100.}&\textcolor{gray}{100.}&\textcolor{gray}{100.}&\textcolor{gray}{100.}&\textcolor{gray}{100.}&\textcolor{gray}{100.}&\textcolor{gray}{99.9}\\
    LAV-DF \cite{cai2022you}& 81.3 & 76.8  & 98.8 & 90.8  & 98.8 & 91.6  & 98.2 & 91.8  & 98.3 & 90.4  & 98.5 & 91.2\\
    AV-Deepfake1M \cite{cai2024av}& 93.9 & 95.5  & 98.2 & 88.6  & 99.7 & 98.1  & 99.4 & 97.8  & 99.4 & 97.5  & 99.2 & 95.5 \\
    \bottomrule
    \end{tabular}
    % }
    \caption{Detailed generalization performance for each forgery type of the FakeAVCeleb dataset.}
    \label{tab:generalization_fakeavceleb_dfd}
\end{table*}

\onecolumn
\begin{longtable}{p{0.1\textwidth}p{0.4\textwidth}p{0.1\textwidth}p{0.1\textwidth}p{0.1\textwidth}p{0.1\textwidth}}
\caption{Real world analysis dataset details.}\label{tab:real_world_info} \\
\toprule
Class & URL & Language & Video-Duration & Processing-Time & Coverage c \\
\midrule
\endhead
\midrule
\multicolumn{6}{r}{\textit{Continued on next page}} \\
\endfoot
\bottomrule
\endlastfoot
\RaggedRight real \#1 & \RaggedRight \url{https://www.youtube.com/watch?v=bej61P8YaXo} & \RaggedRight Russian & \RaggedRight 00:00:47 & \RaggedRight 0:01:31 & \RaggedRight 0.198 \\
\RaggedRight real \#2 & \RaggedRight \url{https://www.youtube.com/watch?v=wgx5neIx6qc} & \RaggedRight Russian & \RaggedRight 00:01:16 & \RaggedRight 0:01:03 & \RaggedRight 0.059 \\
\RaggedRight real \#3 & \RaggedRight \url{-} & \RaggedRight Russian & \RaggedRight 00:02:00 & \RaggedRight 0:01:36 & \RaggedRight 0.177 \\
\RaggedRight real \#4 & \RaggedRight \url{https://www.youtube.com/watch?v=_4PKe8WGCPg} & \RaggedRight English & \RaggedRight 00:14:27 & \RaggedRight 0:22:50 & \RaggedRight 0.105 \\
\RaggedRight real \#5 & \RaggedRight \url{https://www.youtube.com/watch?v=MBFnZhUq-Es} & \RaggedRight Ukrainian & \RaggedRight 00:01:04 & \RaggedRight 0:00:44 & \RaggedRight 0.119 \\
\RaggedRight real \#6 & \RaggedRight \url{https://www.instagram.com/zelenskyy_official/p/CaaFzibgLES/} & \RaggedRight Russian & \RaggedRight 00:00:32 & \RaggedRight 0:00:27 & \RaggedRight 0.063 \\
\RaggedRight real \#7 & \RaggedRight \url{https://www.youtube.com/watch?v=NrDtNOiP65U} & \RaggedRight Russian & \RaggedRight 00:01:08 & \RaggedRight 0:00:57 & \RaggedRight 0.081 \\
\RaggedRight real \#8 & \RaggedRight \url{https://www.instagram.com/zelenskyy_official/p/CZKXF33KqEl/} & \RaggedRight Ukrainian & \RaggedRight 00:04:05 & \RaggedRight 0:03:18 & \RaggedRight 0.041 \\
\RaggedRight real \#9 & \RaggedRight \url{https://www.youtube.com/watch?v=Cz2mFOJLaRI} & \RaggedRight Russian & \RaggedRight 00:01:34 & \RaggedRight 0:03:10 & \RaggedRight 0.118 \\
\RaggedRight real \#10 & \RaggedRight \url{https://www.youtube.com/watch?v=CInwveNLUmA} & \RaggedRight English & \RaggedRight 00:01:34 & \RaggedRight 0:02:53 & \RaggedRight 0.073 \\
\RaggedRight real \#11 & \RaggedRight \url{https://www.instagram.com/zelenskyy_official/p/CaBw7eNIDKp/} & \RaggedRight Ukrainian & \RaggedRight 00:01:07 & \RaggedRight 0:00:55 & \RaggedRight 0.028 \\
\RaggedRight real \#12 & \RaggedRight \url{https://www.youtube.com/watch?v=GdxofSvTYUI} & \RaggedRight English & \RaggedRight 00:03:05 & \RaggedRight 0:05:40 & \RaggedRight 0.031 \\
\RaggedRight real \#13 & \RaggedRight \url{https://www.youtube.com/watch?v=4WpjgtK2ryo} & \RaggedRight Russian & \RaggedRight 00:01:00 & \RaggedRight 0:01:51 & \RaggedRight 0.051 \\
\RaggedRight real \#14 & \RaggedRight \url{-} & \RaggedRight Italian & \RaggedRight 00:01:03 & \RaggedRight 0:00:51 & \RaggedRight 0.295 \\
\RaggedRight real \#15 & \RaggedRight \url{https://www.youtube.com/watch?v=qPVVAy9df_Y} & \RaggedRight Russian & \RaggedRight 00:11:13 & \RaggedRight 0:09:27 & \RaggedRight 0.415 \\
\RaggedRight real \#16 & \RaggedRight \url{https://www.youtube.com/watch?v=s9s-4b5hYpw} & \RaggedRight Russian & \RaggedRight 00:01:06 & \RaggedRight 0:00:59 & \RaggedRight 0.094 \\
\RaggedRight real \#17 & \RaggedRight \url{-} & \RaggedRight Italian & \RaggedRight 00:01:13 & \RaggedRight 0:00:59 & \RaggedRight 0.061 \\
\RaggedRight real \#18 & \RaggedRight \url{https://www.youtube.com/watch?v=xVwU-YMgHvo} & \RaggedRight English & \RaggedRight 00:07:01 & \RaggedRight 0:10:50 & \RaggedRight 0.197 \\
\RaggedRight real \#19 & \RaggedRight \url{https://www.youtube.com/watch?v=rWi3T9okyvA} & \RaggedRight Russian & \RaggedRight 00:02:19 & \RaggedRight 0:02:23 & \RaggedRight 0.053 \\
\RaggedRight real \#20 & \RaggedRight \url{https://www.instagram.com/zelenskyy_official/p/CVvedjElHwX/} & \RaggedRight Russian & \RaggedRight 00:08:29 & \RaggedRight 0:07:00 & \RaggedRight 0.159 \\
\RaggedRight real \#21 & \RaggedRight \url{https://www.youtube.com/watch?v=hHV4DEvattM} & \RaggedRight Russian & \RaggedRight 00:00:56 & \RaggedRight 0:01:00 & \RaggedRight 0.156 \\
\RaggedRight real \#22 & \RaggedRight \url{https://www.youtube.com/watch?v=HU2wY17Umzg} & \RaggedRight English & \RaggedRight 00:01:03 & \RaggedRight 0:02:49 & \RaggedRight 0.204 \\
\RaggedRight real \#23 & \RaggedRight \url{-} & \RaggedRight Italian & \RaggedRight 00:02:00 & \RaggedRight 0:01:24 & \RaggedRight 0.025 \\
\RaggedRight real \#24 & \RaggedRight \url{https://www.youtube.com/watch?v=taykmAyaw-Q} & \RaggedRight Russian & \RaggedRight 00:02:19 & \RaggedRight 0:01:35 & \RaggedRight 0.165 \\
\RaggedRight real \#25 & \RaggedRight \url{https://www.youtube.com/watch?v=Qc5hwRAXVXs} & \RaggedRight Russian & \RaggedRight 00:03:54 & \RaggedRight 0:04:03 & \RaggedRight 0.1 \\
\RaggedRight real \#26 & \RaggedRight \url{-} & \RaggedRight Russian & \RaggedRight 00:02:30 & \RaggedRight 0:01:42 & \RaggedRight 0.343 \\
\RaggedRight real \#27 & \RaggedRight \url{https://www.youtube.com/watch?v=zdH0FhnBWwY} & \RaggedRight Russian & \RaggedRight 00:00:51 & \RaggedRight 0:01:43 & \RaggedRight 0.0 \\
\RaggedRight real \#28 & \RaggedRight \url{https://www.instagram.com/zelenskyy_official/p/CakB_4yAUsa/} & \RaggedRight Ukrainian & \RaggedRight 00:06:33 & \RaggedRight 0:05:21 & \RaggedRight 0.611 \\
\RaggedRight real \#29 & \RaggedRight \url{https://www.youtube.com/watch?v=75g7cA-_SEI} & \RaggedRight Russian & \RaggedRight 00:02:12 & \RaggedRight 0:04:47 & \RaggedRight 0.095 \\
\RaggedRight real \#30 & \RaggedRight \url{https://www.youtube.com/watch?v=Hl9Sp5xJxwk} & \RaggedRight German & \RaggedRight 00:03:34 & \RaggedRight 0:02:30 & \RaggedRight 0.163 \\
\RaggedRight real \#31 & \RaggedRight \url{https://www.youtube.com/watch?v=caxMBk1__-Y} & \RaggedRight English & \RaggedRight 00:03:41 & \RaggedRight 0:02:52 & \RaggedRight 0.105 \\
\RaggedRight real \#32 & \RaggedRight \url{-} & \RaggedRight Italian & \RaggedRight 00:02:00 & \RaggedRight 0:01:26 & \RaggedRight 0.0 \\
\RaggedRight real \#33 & \RaggedRight \url{https://www.youtube.com/watch?v=hQ58Yv6kP44} & \RaggedRight Russian & \RaggedRight 00:30:46 & \RaggedRight 0:37:10 & \RaggedRight 0.391 \\
\RaggedRight real \#34 & \RaggedRight \url{https://www.youtube.com/watch?v=nDzX13RHTXI} & \RaggedRight Arabic & \RaggedRight 00:46:26 & \RaggedRight 0:53:14 & \RaggedRight 0.094 \\
\RaggedRight real \#35 & \RaggedRight \url{https://www.instagram.com/zelenskyy_official/p/CUHvAc9AFoX/} & \RaggedRight Ukrainian & \RaggedRight 00:01:42 & \RaggedRight 0:01:20 & \RaggedRight 0.13 \\
\RaggedRight real \#36 & \RaggedRight \url{https://www.youtube.com/watch?v=7WUyEPkGnAc} & \RaggedRight Russian & \RaggedRight 00:00:37 & \RaggedRight 0:00:38 & \RaggedRight 0.104 \\
\RaggedRight real \#37 & \RaggedRight \url{https://www.youtube.com/watch?v=ujCFOPfqJbQ} & \RaggedRight Russian & \RaggedRight 00:08:13 & \RaggedRight 0:08:51 & \RaggedRight 0.116 \\
\RaggedRight real \#38 & \RaggedRight \url{https://www.youtube.com/watch?v=BEh1Fr-BaN8} & \RaggedRight English & \RaggedRight 00:01:46 & \RaggedRight 0:01:17 & \RaggedRight 0.031 \\
\RaggedRight real \#39 & \RaggedRight \url{https://www.instagram.com/zelenskyy_official/p/CaklNXjAYz3/} & \RaggedRight Russian & \RaggedRight 00:14:02 & \RaggedRight 0:12:36 & \RaggedRight 0.085 \\
\RaggedRight real \#40 & \RaggedRight \url{https://www.youtube.com/watch?v=_g_AP4hhGQM} & \RaggedRight Russian & \RaggedRight 00:02:04 & \RaggedRight 0:02:08 & \RaggedRight 0.025 \\
\RaggedRight real \#41 & \RaggedRight \url{https://www.youtube.com/watch?v=DPIR4EmaDDE} & \RaggedRight Russian & \RaggedRight 00:01:16 & \RaggedRight 0:01:00 & \RaggedRight 0.033 \\
\RaggedRight real \#42 & \RaggedRight \url{https://www.instagram.com/zelenskyy_official/p/CaokxUWARj7/} & \RaggedRight Ukrainian & \RaggedRight 00:08:25 & \RaggedRight 0:13:16 & \RaggedRight 0.066 \\
\RaggedRight real \#43 & \RaggedRight \url{https://www.youtube.com/watch?v=rZ2J4b-dkUs} & \RaggedRight English & \RaggedRight 00:00:29 & \RaggedRight 0:00:58 & \RaggedRight 0.535 \\
\RaggedRight real \#44 & \RaggedRight \url{https://www.instagram.com/zelenskyy_official/p/CaorkbdgSv4/} & \RaggedRight Ukrainian & \RaggedRight 00:07:04 & \RaggedRight 0:11:44 & \RaggedRight 0.05 \\
\RaggedRight real \#45 & \RaggedRight \url{https://www.instagram.com/zelenskyy_official/p/Cab8euXgtyF/} & \RaggedRight Russian & \RaggedRight 00:05:16 & \RaggedRight 0:04:22 & \RaggedRight 0.046 \\
\RaggedRight real \#46 & \RaggedRight \url{https://www.youtube.com/watch?v=2Vmx47wsOBQ} & \RaggedRight Russian & \RaggedRight 00:00:53 & \RaggedRight 0:00:36 & \RaggedRight 0.034 \\
\RaggedRight real \#47 & \RaggedRight \url{-} & \RaggedRight Italian & \RaggedRight 00:02:00 & \RaggedRight 0:01:28 & \RaggedRight 0.024 \\
\RaggedRight real \#48 & \RaggedRight \url{-} & \RaggedRight Italian & \RaggedRight 00:02:00 & \RaggedRight 0:01:34 & \RaggedRight 0.107 \\
\RaggedRight real \#49 & \RaggedRight \url{https://www.instagram.com/zelenskyy_official/p/CaUvEpaAKWj/} & \RaggedRight Ukrainian & \RaggedRight 00:00:41 & \RaggedRight 0:01:11 & \RaggedRight 0.026 \\
\RaggedRight real \#50 & \RaggedRight \url{https://www.youtube.com/watch?v=cxKn2sv2cgQ} & \RaggedRight Ukrainian & \RaggedRight 00:05:34 & \RaggedRight 0:10:49 & \RaggedRight 0.048 \\
\RaggedRight real \#51 & \RaggedRight \url{https://www.youtube.com/watch?v=zHdj1HW4cQw} & \RaggedRight Russian & \RaggedRight 00:02:06 & \RaggedRight 0:04:59 & \RaggedRight 0.17 \\
\RaggedRight real \#52 & \RaggedRight \url{-} & \RaggedRight Russian & \RaggedRight 00:02:30 & \RaggedRight 0:01:42 & \RaggedRight 0.268 \\
\RaggedRight real \#53 & \RaggedRight \url{https://www.youtube.com/watch?v=36_Vj3N3oIk} & \RaggedRight Ukrainian & \RaggedRight 00:06:02 & \RaggedRight 0:13:33 & \RaggedRight 0.035 \\
\RaggedRight real \#54 & \RaggedRight \url{https://www.youtube.com/watch?v=prfaWHQoxVg} & \RaggedRight Russian & \RaggedRight 00:01:25 & \RaggedRight 0:01:17 & \RaggedRight 0.037 \\
\RaggedRight real \#55 & \RaggedRight \url{https://www.youtube.com/watch?v=M0iV5vIABX0} & \RaggedRight English & \RaggedRight 00:01:36 & \RaggedRight 0:01:20 & \RaggedRight 0.076 \\
\RaggedRight real \#56 & \RaggedRight \url{https://www.instagram.com/zelenskyy_official/p/CWh0pPuFOTA/} & \RaggedRight Ukrainian & \RaggedRight 00:07:10 & \RaggedRight 0:13:56 & \RaggedRight 0.09 \\
\RaggedRight real \#57 & \RaggedRight \url{https://www.youtube.com/watch?v=hLCGKRF3YAI} & \RaggedRight Russian & \RaggedRight 00:01:47 & \RaggedRight 0:01:45 & \RaggedRight 0.162 \\
\RaggedRight real \#58 & \RaggedRight \url{-} & \RaggedRight Russian & \RaggedRight 00:02:24 & \RaggedRight 0:04:14 & \RaggedRight 0.349 \\
\RaggedRight real \#59 & \RaggedRight \url{-} & \RaggedRight Italian & \RaggedRight 00:01:00 & \RaggedRight 0:00:51 & \RaggedRight 0.378 \\
\RaggedRight real \#60 & \RaggedRight \url{https://www.youtube.com/watch?v=1CnyqLogH0Y} & \RaggedRight Russian & \RaggedRight 00:01:40 & \RaggedRight 0:01:29 & \RaggedRight 0.127 \\
\RaggedRight real \#61 & \RaggedRight \url{https://www.youtube.com/watch?v=WhqVwPYOY7M} & \RaggedRight Ukrainian & \RaggedRight 00:03:45 & \RaggedRight 0:03:00 & \RaggedRight 0.14 \\
\RaggedRight real \#62 & \RaggedRight \url{-} & \RaggedRight Russian & \RaggedRight 00:02:27 & \RaggedRight 0:04:28 & \RaggedRight 0.056 \\
\RaggedRight real \#63 & \RaggedRight \url{https://www.youtube.com/watch?v=TFl2nfPoxVk} & \RaggedRight Russian & \RaggedRight 00:00:41 & \RaggedRight 0:00:35 & \RaggedRight 0.049 \\
\RaggedRight real \#64 & \RaggedRight \url{https://www.instagram.com/zelenskyy_official/p/CZEnvm0KdQv/} & \RaggedRight Russian & \RaggedRight 00:01:06 & \RaggedRight 0:00:55 & \RaggedRight 0.025 \\
\RaggedRight real \#65 & \RaggedRight \url{https://www.youtube.com/watch?v=2ze4gImZADM} & \RaggedRight Russian & \RaggedRight 00:01:46 & \RaggedRight 0:01:36 & \RaggedRight 0.03 \\
\RaggedRight real \#66 & \RaggedRight \url{https://www.youtube.com/watch?v=TccwMWVtmj0} & \RaggedRight English & \RaggedRight 00:08:02 & \RaggedRight 0:07:17 & \RaggedRight 0.03 \\
\RaggedRight real \#67 & \RaggedRight \url{https://www.youtube.com/watch?v=J55cSx8Rz_g} & \RaggedRight Russian & \RaggedRight 00:00:34 & \RaggedRight 0:01:17 & \RaggedRight 0.144 \\
\RaggedRight real \#68 & \RaggedRight \url{https://www.instagram.com/zelenskyy_official/p/Cal8YaIFOqM/} & \RaggedRight Ukrainian & \RaggedRight 00:08:04 & \RaggedRight 0:06:47 & \RaggedRight 0.048 \\
\RaggedRight real \#69 & \RaggedRight \url{-} & \RaggedRight Russian & \RaggedRight 00:01:18 & \RaggedRight 0:02:16 & \RaggedRight 0.286 \\
\RaggedRight real \#70 & \RaggedRight \url{-} & \RaggedRight Russian & \RaggedRight 00:02:09 & \RaggedRight 0:01:26 & \RaggedRight 0.155 \\
\RaggedRight real \#71 & \RaggedRight \url{https://www.instagram.com/zelenskyy_official/p/CYmeFavK7wC/} & \RaggedRight English & \RaggedRight 00:02:14 & \RaggedRight 0:04:04 & \RaggedRight 0.032 \\
\RaggedRight real \#72 & \RaggedRight \url{https://www.instagram.com/zelenskyy_official/p/CaKjFkXAhts/} & \RaggedRight Ukrainian & \RaggedRight 00:19:01 & \RaggedRight 0:15:29 & \RaggedRight 0.336 \\
\RaggedRight real \#73 & \RaggedRight \url{https://www.youtube.com/watch?v=ndNyE7f0NVs} & \RaggedRight Russian & \RaggedRight 00:02:28 & \RaggedRight 0:01:42 & \RaggedRight 0.0 \\
\RaggedRight real \#74 & \RaggedRight \url{https://www.youtube.com/watch?v=iCOwqJ8kLCM} & \RaggedRight Russian & \RaggedRight 00:00:58 & \RaggedRight 0:00:47 & \RaggedRight 0.051 \\
\RaggedRight real \#75 & \RaggedRight \url{https://www.youtube.com/watch?v=NdUkaFnunKw} & \RaggedRight Russian & \RaggedRight 00:01:23 & \RaggedRight 0:01:08 & \RaggedRight 0.04 \\
\RaggedRight fake \#1 & \RaggedRight \url{https://twitter.com/i/status/1174316798700924929
} & \RaggedRight English & \RaggedRight 00:02:07 & \RaggedRight 0:01:58 & \RaggedRight 0.339 \\
\RaggedRight fake \#2 & \RaggedRight \url{https://www.threads.net/@cj_wirenut/post/C8atfbOtQ4U} & \RaggedRight English & \RaggedRight 00:00:23 & \RaggedRight 0:00:19 & \RaggedRight 0.342 \\
\RaggedRight fake \#3 & \RaggedRight \url{https://www.dropbox.com/s/o7ndovcsycfchl3/HaoLi-UNIDIR3.mp4
} & \RaggedRight English & \RaggedRight 00:00:59 & \RaggedRight 0:00:47 & \RaggedRight 0.415 \\
\RaggedRight fake \#4 & \RaggedRight \url{https://twitter.com/reface_app/status/1352274324040908801
} & \RaggedRight French & \RaggedRight 00:00:15 & \RaggedRight 0:00:11 & \RaggedRight 0.0 \\
\RaggedRight fake \#5 & \RaggedRight \url{https://twitter.com/kaikertv/status/1284053913516179457
} & \RaggedRight French & \RaggedRight 00:00:32 & \RaggedRight 0:00:31 & \RaggedRight 0.0 \\
\RaggedRight fake \#6 & \RaggedRight \url{https://twitter.com/PhilEhr/status/1311667726742560769
} & \RaggedRight English & \RaggedRight 00:00:59 & \RaggedRight 0:00:49 & \RaggedRight 0.119 \\
\RaggedRight fake \#7 & \RaggedRight \url{https://twitter.com/facemagic_app/status/1404091187728896000
} & \RaggedRight Hakha-Chin & \RaggedRight 00:00:23 & \RaggedRight 0:00:18 & \RaggedRight 0.0 \\
\RaggedRight fake \#8 & \RaggedRight \url{https://twitter.com/ThiudaDeepfakes/status/1429455322389827592
} & \RaggedRight Breton & \RaggedRight 00:00:13 & \RaggedRight 0:00:18 & \RaggedRight 0.849 \\
\RaggedRight fake \#9 & \RaggedRight \url{https://www.instagram.com/p/C3z9bv_u2KH/} & \RaggedRight English & \RaggedRight 00:02:17 & \RaggedRight 0:01:52 & \RaggedRight 0.336 \\
\RaggedRight fake \#10 & \RaggedRight \url{https://www.dropbox.com/s/ej7swf300aynlsz/HaoLi-UNIDIR7.mp4
} & \RaggedRight English & \RaggedRight 00:00:59 & \RaggedRight 0:00:48 & \RaggedRight 0.426 \\
\RaggedRight fake \#11 & \RaggedRight \url{https://twitter.com/dw_sports/status/1405602671525040142
} & \RaggedRight French & \RaggedRight 00:00:44 & \RaggedRight 0:00:57 & \RaggedRight 0.0 \\
\RaggedRight fake \#12 & \RaggedRight \url{https://twitter.com/facemagic_app/status/1426257244728004611
} & \RaggedRight English & \RaggedRight 00:00:21 & \RaggedRight 0:00:18 & \RaggedRight 0.0 \\
\RaggedRight fake \#13 & \RaggedRight \url{https://twitter.com/larrykim/status/1228979054579073024
} & \RaggedRight French & \RaggedRight 00:01:26 & \RaggedRight 0:01:12 & \RaggedRight 0.22 \\
\RaggedRight fake \#14 & \RaggedRight \url{https://twitter.com/EddiePozos_/status/1365420001327337474
} & \RaggedRight Persian & \RaggedRight 00:00:11 & \RaggedRight 0:00:10 & \RaggedRight 0.603 \\
\RaggedRight fake \#15 & \RaggedRight \url{https://twitter.com/sumitsaurabh/status/1324424493717049344
} & \RaggedRight Tamil & \RaggedRight 00:00:09 & \RaggedRight 0:00:09 & \RaggedRight 0.643 \\
\RaggedRight fake \#16 & \RaggedRight \url{https://twitter.com/CestCanteloup/status/1315576171803938817
} & \RaggedRight French & \RaggedRight 00:00:41 & \RaggedRight 0:00:35 & \RaggedRight 0.041 \\
\RaggedRight fake \#17 & \RaggedRight \url{https://twitter.com/welcomeai/status/1090256517666885632
} & \RaggedRight Esperanto & \RaggedRight 00:00:55 & \RaggedRight 0:00:38 & \RaggedRight 0.4 \\
\RaggedRight fake \#18 & \RaggedRight \url{https://www.facebook.com/plugins/video.php?height=476&href=https%3A%2F%2Fwww.facebook.com%2F61551008896469%2Fvideos%2F878366723849002%2F&show_text=true&width=476&t=0} & \RaggedRight English & \RaggedRight 00:01:49 & \RaggedRight 0:01:36 & \RaggedRight 0.279 \\
\RaggedRight fake \#19 & \RaggedRight \url{https://www.youtube.com/watch?v=9ccd3LMNMl8} & \RaggedRight English & \RaggedRight 00:33:15 & \RaggedRight 0:24:34 & \RaggedRight 0.21 \\
\RaggedRight fake \#20 & \RaggedRight \url{https://twitter.com/facemagic_app/status/1419311479065518083
} & \RaggedRight Persian & \RaggedRight 00:00:10 & \RaggedRight 0:00:10 & \RaggedRight 0.19 \\
\RaggedRight fake \#21 & \RaggedRight \url{https://fb.watch/vEHrmibPaG/} & \RaggedRight English & \RaggedRight 00:01:18 & \RaggedRight 0:01:08 & \RaggedRight 0.306 \\
\RaggedRight fake \#22 & \RaggedRight \url{https://twitter.com/facemagic_app/status/1431270277506166796
} & \RaggedRight English & \RaggedRight 00:00:11 & \RaggedRight 0:00:11 & \RaggedRight 0.233 \\
\RaggedRight fake \#23 & \RaggedRight \url{https://packaged-media.redd.it/e4bntb4wn74a1/pb/m2-res_720p.mp4?m=DASHPlaylist.mpd&v=1&e=1743004800&s=c47b62e83cdf59e76e9a1e26a302a02091577c5c} & \RaggedRight English & \RaggedRight 00:00:16 & \RaggedRight 0:00:16 & \RaggedRight 0.546 \\
\RaggedRight fake \#24 & \RaggedRight \url{https://twitter.com/facemagic_app/status/1408439846474072064
} & \RaggedRight Persian & \RaggedRight 00:00:12 & \RaggedRight 0:00:11 & \RaggedRight 0.0 \\
\RaggedRight fake \#25 & \RaggedRight \url{https://twitter.com/facemagic_app/status/1421485812609323011
} & \RaggedRight Spanish & \RaggedRight 00:00:12 & \RaggedRight 0:00:12 & \RaggedRight 0.223 \\
\RaggedRight fake \#26 & \RaggedRight \url{https://twitter.com/facemagic_app/status/1430545513955221508
} & \RaggedRight Hakha-Chin & \RaggedRight 00:00:11 & \RaggedRight 0:00:12 & \RaggedRight 0.463 \\
\RaggedRight fake \#27 & \RaggedRight \url{https://www.facebook.com/plugins/video.php?height=476&href=https%3A%2F%2Fwww.facebook.com%2F61552447227376%2Fvideos%2F870526964468908%2F&show_text=false&width=476&t=0} & \RaggedRight English & \RaggedRight 00:01:42 & \RaggedRight 0:01:30 & \RaggedRight 0.159 \\
\RaggedRight fake \#28 & \RaggedRight \url{https://twitter.com/hytel/status/1425631238208081924
} & \RaggedRight English & \RaggedRight 00:00:43 & \RaggedRight 0:00:38 & \RaggedRight 0.275 \\
\RaggedRight fake \#29 & \RaggedRight \url{https://twitter.com/Happy_Finish/status/1091007206944919559
} & \RaggedRight English & \RaggedRight 00:01:16 & \RaggedRight 0:01:08 & \RaggedRight 0.264 \\
\RaggedRight fake \#30 & \RaggedRight \url{https://twitter.com/FutureAdvocacy/status/1194148111121354752
} & \RaggedRight Welsh & \RaggedRight 00:01:11 & \RaggedRight 0:00:59 & \RaggedRight 0.045 \\
\RaggedRight fake \#31 & \RaggedRight \url{https://twitter.com/facemagic_app/status/1404453586382188550
} & \RaggedRight English & \RaggedRight 00:00:14 & \RaggedRight 0:00:14 & \RaggedRight 0.0 \\
\RaggedRight fake \#32 & \RaggedRight \url{https://twitter.com/facemagic_app/status/1426559247043674112
} & \RaggedRight Persian & \RaggedRight 00:00:15 & \RaggedRight 0:00:13 & \RaggedRight 0.0 \\
\RaggedRight fake \#33 & \RaggedRight \url{https://twitter.com/facemagic_app/status/1428371186468098064
} & \RaggedRight Breton & \RaggedRight 00:00:44 & \RaggedRight 0:00:36 & \RaggedRight 0.427 \\
\RaggedRight fake \#34 & \RaggedRight \url{https://twitter.com/brianmonarch/status/1307037446056677382
} & \RaggedRight English & \RaggedRight 00:00:59 & \RaggedRight 0:00:54 & \RaggedRight 0.446 \\
\RaggedRight fake \#35 & \RaggedRight \url{https://www.bitchute.com/video/8ZGoBmGmKtgi} & \RaggedRight English & \RaggedRight 00:00:37 & \RaggedRight 0:00:32 & \RaggedRight 0.369 \\
\RaggedRight fake \#36 & \RaggedRight \url{https://twitter.com/facemagic_app/status/1424384912723824641
} & \RaggedRight French & \RaggedRight 00:00:17 & \RaggedRight 0:00:14 & \RaggedRight 0.0 \\
\RaggedRight fake \#37 & \RaggedRight \url{https://www.bitchute.com/video/ChybNNbjEEOd} & \RaggedRight English & \RaggedRight 00:02:16 & \RaggedRight 0:01:55 & \RaggedRight 0.114 \\
\RaggedRight fake \#38 & \RaggedRight \url{https://twitter.com/CestCanteloup/status/1277896113618669569
} & \RaggedRight French & \RaggedRight 00:00:51 & \RaggedRight 0:00:39 & \RaggedRight 0.153 \\
\RaggedRight fake \#39 & \RaggedRight \url{https://www.dropbox.com/s/5ll22su3gzc4qyn/HaoLi-UNIDIR4.mp4
} & \RaggedRight English & \RaggedRight 00:00:59 & \RaggedRight 0:00:47 & \RaggedRight 0.201 \\
\RaggedRight fake \#40 & \RaggedRight \url{https://twitter.com/facemagic_app/status/1406627903249518596
} & \RaggedRight English & \RaggedRight 00:00:23 & \RaggedRight 0:00:18 & \RaggedRight 0.683 \\
\RaggedRight fake \#41 & \RaggedRight \url{https://www.dropbox.com/s/b9o21z834kwi59s/HaoLi-UNIDIR1.mp4
} & \RaggedRight English & \RaggedRight 00:00:59 & \RaggedRight 0:00:46 & \RaggedRight 0.504 \\
\RaggedRight fake \#42 & \RaggedRight \url{https://twitter.com/insanedio/status/1308620241732157440
} & \RaggedRight Spanish & \RaggedRight 00:00:44 & \RaggedRight 0:00:35 & \RaggedRight 0.391 \\
\RaggedRight fake \#43 & \RaggedRight \url{https://www.youtube.com/watch?v=J7WqjrtJHkk} & \RaggedRight Welsh & \RaggedRight 00:00:50 & \RaggedRight 0:00:42 & \RaggedRight 0.07 \\
\RaggedRight fake \#44 & \RaggedRight \url{https://www.facebook.com/100092555835479/posts/1389999691891907/} & \RaggedRight Romanian & \RaggedRight 00:00:15 & \RaggedRight 0:00:14 & \RaggedRight 0.115 \\
\RaggedRight fake \#45 & \RaggedRight \url{https://twitter.com/facemagic_app/status/1410251795348271126
} & \RaggedRight English & \RaggedRight 00:00:19 & \RaggedRight 0:00:16 & \RaggedRight 0.058 \\
\RaggedRight fake \#46 & \RaggedRight \url{https://www.facebook.com/100095605808597/posts/3495941347387764/} & \RaggedRight Romanian & \RaggedRight 00:01:04 & \RaggedRight 0:00:45 & \RaggedRight 0.79 \\
\RaggedRight fake \#47 & \RaggedRight \url{https://twitter.com/stephaniemain2/status/1315238754148212736
} & \RaggedRight Esperanto & \RaggedRight 00:00:15 & \RaggedRight 0:00:10 & \RaggedRight 0.322 \\
\RaggedRight fake \#48 & \RaggedRight \url{https://twitter.com/Seekthetruth101/status/1308135272706695169
} & \RaggedRight English & \RaggedRight 00:02:09 & \RaggedRight 0:01:48 & \RaggedRight 0.174 \\
\RaggedRight fake \#49 & \RaggedRight \url{https://twitter.com/facemagic_app/status/1416774762441830406
} & \RaggedRight English & \RaggedRight 00:00:16 & \RaggedRight 0:00:13 & \RaggedRight 0.594 \\
\RaggedRight fake \#50 & \RaggedRight \url{https://twitter.com/facemagic_app/status/1426921628462567424
} & \RaggedRight Dhivehi & \RaggedRight 00:00:13 & \RaggedRight 0:00:13 & \RaggedRight 0.752 \\
\RaggedRight fake \#51 & \RaggedRight \url{https://twitter.com/EvanKirstel/status/1268984667153276928
} & \RaggedRight English & \RaggedRight 00:01:01 & \RaggedRight 0:00:51 & \RaggedRight 0.174 \\
\RaggedRight fake \#52 & \RaggedRight \url{https://twitter.com/drfakenstein/status/1259838123468632065
} & \RaggedRight English & \RaggedRight 00:00:57 & \RaggedRight 0:00:47 & \RaggedRight 0.431 \\
\RaggedRight fake \#53 & \RaggedRight \url{https://twitter.com/facemagic_app/status/1429095960093790210
} & \RaggedRight English & \RaggedRight 00:00:24 & \RaggedRight 0:00:20 & \RaggedRight 0.259 \\
\RaggedRight fake \#54 & \RaggedRight \url{https://www.bitchute.com/video/sq78wmU60uHY} & \RaggedRight German & \RaggedRight 00:00:44 & \RaggedRight 0:01:08 & \RaggedRight 0.365 \\
\RaggedRight fake \#55 & \RaggedRight \url{https://twitter.com/KRMTRoySuryo2/status/1327286419857915910
} & \RaggedRight Spanish & \RaggedRight 00:00:50 & \RaggedRight 0:00:37 & \RaggedRight 0.801 \\
\RaggedRight fake \#56 & \RaggedRight \url{https://media2.ellinikahoaxes.gr/uploads/2024/10/video-isxurismou-deepfake.mp4} & \RaggedRight English & \RaggedRight 00:00:09 & \RaggedRight 0:00:09 & \RaggedRight 0.04 \\
\RaggedRight fake \#57 & \RaggedRight \url{https://www.facebook.com/EtzrodtJaimeeShuler/posts/7939864682747869/} & \RaggedRight English & \RaggedRight 00:02:03 & \RaggedRight 0:01:55 & \RaggedRight 0.322 \\
\RaggedRight fake \#58 & \RaggedRight \url{https://twitter.com/sp_a/status/998089909369016325
} & \RaggedRight English & \RaggedRight 00:00:54 & \RaggedRight 0:00:39 & \RaggedRight 0.269 \\
\RaggedRight fake \#59 & \RaggedRight \url{https://twitter.com/drfakenstein/status/1171638420822736897
} & \RaggedRight Chinese-Taiwan & \RaggedRight 00:00:59 & \RaggedRight 0:00:49 & \RaggedRight 0.017 \\
\RaggedRight fake \#60 & \RaggedRight \url{https://twitter.com/facemagic_app/status/1420761028468649985
} & \RaggedRight English & \RaggedRight 00:00:16 & \RaggedRight 0:00:15 & \RaggedRight 0.032 \\
\RaggedRight fake \#61 & \RaggedRight \url{https://www.facebook.com/plugins/video.php?height=476&href=https%3A%2F%2Fwww.facebook.com%2F100089533433509%2Fvideos%2F897022265405097%2F&show_text=false&width=476&t=0} & \RaggedRight English & \RaggedRight 00:01:40 & \RaggedRight 0:01:27 & \RaggedRight 0.292 \\
\RaggedRight fake \#62 & \RaggedRight \url{https://twitter.com/hwingo/status/1090809957342154752
} & \RaggedRight English & \RaggedRight 00:01:35 & \RaggedRight 0:01:09 & \RaggedRight 0.089 \\
\RaggedRight fake \#63 & \RaggedRight \url{https://twitter.com/facemagic_app/status/1426196858418843655
} & \RaggedRight Chinese-Taiwan & \RaggedRight 00:00:22 & \RaggedRight 0:00:17 & \RaggedRight 0.253 \\
\RaggedRight fake \#64 & \RaggedRight \url{https://www.facebook.com/plugins/video.php?height=476&href=https%3A%2F%2Fwww.facebook.com%2F61550777281885%2Fvideos%2F1992291657812664%2F&show_text=false&width=476&t=0} & \RaggedRight English & \RaggedRight 00:01:40 & \RaggedRight 0:01:26 & \RaggedRight 0.323 \\
\RaggedRight fake \#65 & \RaggedRight \url{https://web.archive.org/web/20240816152954if_/https://video.xx.fbcdn.net/v/t42.1790-2/453528538_3676577589299215_6162862663998579296_n.mp4?_nc_cat=106&ccb=1-7&_nc_sid=55d0d3&efg=eyJ2ZW5jb2RlX3RhZyI6InN2ZV9zZCIsInZpZGVvX2lkIjo1MDc0Nzc3NTQ5OTkxNjF9&_nc_ohc=fI8yJJw2qCkQ7kNvgG5DX0W&_nc_ht=scontent.fprg1-1.fna&oh=00_AYD-yhNK-2jlWPV8DVBgUL2KpMUOMiMbue1ss52pCISILg&oe=66C548DA} & \RaggedRight Czech & \RaggedRight 00:01:03 & \RaggedRight 0:00:50 & \RaggedRight 0.661 \\
\RaggedRight fake \#66 & \RaggedRight \url{https://twitter.com/facemagic_app/status/1401916866038169601
} & \RaggedRight English & \RaggedRight 00:00:25 & \RaggedRight 0:00:20 & \RaggedRight 0.809 \\
\RaggedRight fake \#67 & \RaggedRight \url{https://video.twimg.com/ext_tw_video/1407579338531291138/pu/vid/320x568/nNPZ0rn3C9kFsr_P.mp4
} & \RaggedRight Russian & \RaggedRight 00:00:19 & \RaggedRight 0:00:17 & \RaggedRight 0.242 \\
\RaggedRight fake \#68 & \RaggedRight \url{https://twitter.com/facemagic_app/status/1409889406404726784
} & \RaggedRight English & \RaggedRight 00:00:16 & \RaggedRight 0:00:15 & \RaggedRight 0.0 \\
\RaggedRight fake \#69 & \RaggedRight \url{https://twitter.com/adrlenard/status/1172567432239751168
} & \RaggedRight English & \RaggedRight 00:00:21 & \RaggedRight 0:00:18 & \RaggedRight 0.264 \\
\RaggedRight fake \#70 & \RaggedRight \url{https://twitter.com/facemagic_app/status/1428008798317387777
} & \RaggedRight Esperanto & \RaggedRight 00:00:27 & \RaggedRight 0:00:19 & \RaggedRight 0.0 \\
\RaggedRight fake \#71 & \RaggedRight \url{https://twitter.com/i/status/1095597694712799232
} & \RaggedRight French & \RaggedRight 00:00:16 & \RaggedRight 0:00:14 & \RaggedRight 0.0 \\
\RaggedRight fake \#72 & \RaggedRight \url{https://twitter.com/PapillonFlacko/status/1282885977518809089
} & \RaggedRight Kabyle & \RaggedRight 00:02:18 & \RaggedRight 0:01:40 & \RaggedRight 0.326 \\
\RaggedRight fake \#73 & \RaggedRight \url{https://twitter.com/facemagic_app/status/1417861927808606212
} & \RaggedRight French & \RaggedRight 00:00:10 & \RaggedRight 0:00:11 & \RaggedRight 0.72 \\
\RaggedRight fake \#74 & \RaggedRight \url{https://twitter.com/facemagic_app/status/1406265519234224132
} & \RaggedRight Hakha-Chin & \RaggedRight 00:00:19 & \RaggedRight 0:00:16 & \RaggedRight 0.424 \\
\RaggedRight fake \#75 & \RaggedRight \url{https://www.bitchute.com/video/bvJsG8fyeCTt} & \RaggedRight English & \RaggedRight 00:00:26 & \RaggedRight 0:00:23 & \RaggedRight 0.192 \\
\RaggedRight fake \#76 & \RaggedRight \url{-} & \RaggedRight English & \RaggedRight 00:00:27 & \RaggedRight 0:00:22 & \RaggedRight 0.355 \\
\RaggedRight fake \#77 & \RaggedRight \url{https://www.tiktok.com/@jbviet/video/6994005244309032197
} & \RaggedRight French & \RaggedRight 00:00:18 & \RaggedRight 0:00:16 & \RaggedRight 0.136 \\
\RaggedRight fake \#78 & \RaggedRight \url{https://twitter.com/facemagic_app/status/1424747301277364229
} & \RaggedRight Romansh-Sursilvan & \RaggedRight 00:00:18 & \RaggedRight 0:00:14 & \RaggedRight 0.613 \\
\RaggedRight fake \#79 & \RaggedRight \url{https://www.facebook.com/Jmvgames/videos/821549489686229} & \RaggedRight English & \RaggedRight 00:01:16 & \RaggedRight 0:01:05 & \RaggedRight 0.447 \\
\RaggedRight fake \#80 & \RaggedRight \url{https://twitter.com/facemagic_app/status/1430183118808387585
} & \RaggedRight Welsh & \RaggedRight 00:00:11 & \RaggedRight 0:00:12 & \RaggedRight 0.038 \\
\RaggedRight fake \#81 & \RaggedRight \url{https://twitter.com/CestCanteloup/status/1276794899355049984
} & \RaggedRight French & \RaggedRight 00:00:46 & \RaggedRight 0:00:37 & \RaggedRight 0.107 \\
\RaggedRight fake \#82 & \RaggedRight \url{https://twitter.com/Nosetwittears/status/1425532656532475919
} & \RaggedRight English & \RaggedRight 00:00:17 & \RaggedRight 0:00:14 & \RaggedRight 0.055 \\
\RaggedRight fake \#83 & \RaggedRight \url{https://x.com/TheBabylonBee/status/1836518555648692570} & \RaggedRight English & \RaggedRight 00:01:55 & \RaggedRight 0:01:36 & \RaggedRight 0.151 \\
\RaggedRight fake \#84 & \RaggedRight \url{https://twitter.com/GACrewFans/status/1308559503462297600
} & \RaggedRight English & \RaggedRight 00:00:14 & \RaggedRight 0:00:11 & \RaggedRight 0.0 \\
\RaggedRight fake \#85 & \RaggedRight \url{https://www.facebook.com/plugins/video.php?height=476&href=https%3A%2F%2Fwww.facebook.com%2FNwadshairnnails%2Fvideos%2F203824829410862%2F&show_text=false&width=476&t=0} & \RaggedRight English & \RaggedRight 00:01:42 & \RaggedRight 0:01:27 & \RaggedRight 0.303 \\
\RaggedRight fake \#86 & \RaggedRight \url{https://twitter.com/RT_com/status/1308410711027089411
} & \RaggedRight Dhivehi & \RaggedRight 00:02:02 & \RaggedRight 0:01:42 & \RaggedRight 0.191 \\
\RaggedRight fake \#87 & \RaggedRight \url{https://twitter.com/Knuckle_HeadTV/status/1426780954056658947
} & \RaggedRight Romansh-Sursilvan & \RaggedRight 00:01:00 & \RaggedRight 0:00:50 & \RaggedRight 0.216 \\
\RaggedRight fake \#88 & \RaggedRight \url{https://twitter.com/GeorgeTakei/status/1269459672714694661
} & \RaggedRight English & \RaggedRight 00:01:22 & \RaggedRight 0:02:02 & \RaggedRight 0.025 \\
\RaggedRight fake \#89 & \RaggedRight \url{https://www.dropbox.com/s/d0qfadmo05ckdoj/HaoLi-UNIDIR8.mp4
} & \RaggedRight English & \RaggedRight 00:00:59 & \RaggedRight 0:00:50 & \RaggedRight 0.508 \\
\RaggedRight fake \#90 & \RaggedRight \url{https://twitter.com/nuweverse/status/1320405519438405634
} & \RaggedRight English & \RaggedRight 00:00:10 & \RaggedRight 0:00:10 & \RaggedRight 0.35 \\
\RaggedRight fake \#91 & \RaggedRight \url{https://twitter.com/NuritBen/status/1284084875897778177
} & \RaggedRight Welsh & \RaggedRight 00:01:09 & \RaggedRight 0:01:02 & \RaggedRight 0.107 \\
\RaggedRight fake \#92 & \RaggedRight \url{https://media2.ellinikahoaxes.gr/uploads/2024/06/447088377_986072086506420_2750102588021047120_n.mp4} & \RaggedRight Greek & \RaggedRight 00:03:57 & \RaggedRight 0:03:08 & \RaggedRight 0.671 \\
\RaggedRight fake \#93 & \RaggedRight \url{https://twitter.com/simerazzi/status/1333843243516223488
} & \RaggedRight Kabyle & \RaggedRight 00:00:15 & \RaggedRight 0:00:12 & \RaggedRight 0.907 \\
\RaggedRight fake \#94 & \RaggedRight \url{https://twitter.com/facemagic_app/status/1411701337981661192
} & \RaggedRight Hakha-Chin & \RaggedRight 00:00:08 & \RaggedRight 0:00:09 & \RaggedRight 0.236 \\
\RaggedRight fake \#95 & \RaggedRight \url{https://twitter.com/facemagic_app/status/1427284018979090433
} & \RaggedRight English & \RaggedRight 00:00:33 & \RaggedRight 0:00:26 & \RaggedRight 0.436 \\
\RaggedRight fake \#96 & \RaggedRight \url{https://twitter.com/callmeuschi/status/1377599278252707845?s=21
} & \RaggedRight English & \RaggedRight 00:00:17 & \RaggedRight 0:00:14 & \RaggedRight 0.231 \\
\RaggedRight fake \#97 & \RaggedRight \url{https://twitter.com/facemagic_app/status/1427646404344700928
} & \RaggedRight English & \RaggedRight 00:00:16 & \RaggedRight 0:00:15 & \RaggedRight 0.033 \\
\RaggedRight fake \#98 & \RaggedRight \url{https://twitter.com/facemagic_app/status/1411338949977591812
} & \RaggedRight English & \RaggedRight 00:00:18 & \RaggedRight 0:00:15 & \RaggedRight 0.303 \\
\RaggedRight fake \#99 & \RaggedRight \url{https://twitter.com/facemagic_app/status/1430183114337361931
} & \RaggedRight English & \RaggedRight 00:00:28 & \RaggedRight 0:00:21 & \RaggedRight 0.131 \\
\RaggedRight fake \#100 & \RaggedRight \url{-} & \RaggedRight English & \RaggedRight 00:00:28 & \RaggedRight 0:00:26 & \RaggedRight 0.079 \\
\RaggedRight fake \#101 & \RaggedRight \url{https://www.dailymotion.com/video/x7mf9hf
} & \RaggedRight English & \RaggedRight 00:03:02 & \RaggedRight 0:02:11 & \RaggedRight 0.122 \\
\RaggedRight fake \#102 & \RaggedRight \url{https://www.dropbox.com/s/6v58miw1o7c015z/HaoLi-UNIDIR2.mp4
} & \RaggedRight English & \RaggedRight 00:00:59 & \RaggedRight 0:00:49 & \RaggedRight 0.491 \\
\RaggedRight fake \#103 & \RaggedRight \url{https://www.facebook.com/100001934236498/videos/594209412579621/} & \RaggedRight English & \RaggedRight 00:01:39 & \RaggedRight 0:01:23 & \RaggedRight 0.252 \\
\RaggedRight fake \#104 & \RaggedRight \url{https://www.dropbox.com/s/8n7o5ajkllw4wgm/HaoLi-UNIDIR6.mp4
} & \RaggedRight English & \RaggedRight 00:00:59 & \RaggedRight 0:00:47 & \RaggedRight 0.444 \\
\RaggedRight fake \#105 & \RaggedRight \url{https://web.archive.org/web/20240821053915if_/https://video-lga3-2.xx.fbcdn.net/o1/v/t2/f2/m69/An862HN9EEl8l00qEY6Y09SfPqP08f9VqMtlf4F9ZkQ8cY4H5jrsoEyvCLx4MDKQy0EqvFeuKwZ_1zS4s4fmXTNa.mp4?efg=eyJ2ZW5jb2RlX3RhZyI6Im9lcF9oZCJ9&_nc_ht=video-lga3-2.xx.fbcdn.net&_nc_cat=111&strext=1&vs=fb8e3bd95d6a8122&_nc_vs=HBksFQIYOnBhc3N0aHJvdWdoX2V2ZXJzdG9yZS9HQzgtTHh2U1c0aTF5UkVEQUxrWm1xc25HWEV6Ym1kakFBQUYVAALIAQAVAhg6cGFzc3Rocm91Z2hfZXZlcnN0b3JlL0dLLUJNUnVjQmhNdkxfb0ZBSG5CR21icEl0Wi1ickZxQUFBRhUCAsgBAEsHiBJwcm9ncmVzc2l2ZV9yZWNpcGUBMQ1zdWJzYW1wbGVfZnBzABB2bWFmX2VuYWJsZV9uc3ViACBtZWFzdXJlX29yaWdpbmFsX3Jlc29sdXRpb25fc3NpbQAoY29tcHV0ZV9zc2ltX29ubHlfYXRfb3JpZ2luYWxfcmVzb2x1dGlvbgAddXNlX2xhbmN6b3NfZm9yX3ZxbV91cHNjYWxpbmcAEWRpc2FibGVfcG9zdF9wdnFzABUAJQAcjBdAAAAAAAAAABERAAAAJqC436e6mMwDFQIoAkMzGAt2dHNfcHJldmlldxwXQErGyLQ5WBAYGWRhc2hfaDI2NC1iYXNpYy1nZW4yXzcyMHASABgYdmlkZW9zLnZ0cy5jYWxsYmFjay5wcm9kOBJWSURFT19WSUVXX1JFUVVFU1QbCogVb2VtX3RhcmdldF9lbmNvZGVfdGFnBm9lcF9oZBNvZW1fcmVxdWVzdF90aW1lX21zATAMb2VtX2NmZ19ydWxlB3VubXV0ZWQTb2VtX3JvaV9yZWFjaF9jb3VudAM0MzYRb2VtX2lzX2V4cGVyaW1lbnQADG9lbV92aWRlb19pZBAxNjk1Njk0ODU0NTIyNzEzEm9lbV92aWRlb19hc3NldF9pZBAyMjA1Mjg1NTkzMTgxODE4FW9lbV92aWRlb19yZXNvdXJjZV9pZBAxMDExOTcwODQwNzE4ODY0HG9lbV9zb3VyY2VfdmlkZW9fZW5jb2RpbmdfaWQPODIxMTU3NjM5OTk4NjI5DnZ0c19yZXF1ZXN0X2lkACUCHAAljgIbCIgBcwQ0NTg2AmNkCjIwMjQtMDgtMTkDcmNiAzQwMANhcHAFVmlkZW8CY3QZQ09OVEFJTkVEX1BPU1RfQVRUQUNITUVOVBNvcmlnaW5hbF9kdXJhdGlvbl9zBjUzLjU2OAFmAmFkAnRzFXByb2dyZXNzaXZlX2VuY29kaW5ncwA%3D&ccb=9-4&oh=00_AYByVnpiEhIPvm2WhRSvuqB-gWs_wYumgmOnAvtFGL4NfA&oe=66C748CE&_nc_sid=1d576d&_nc_rid=779600729108004&_nc_store_type=} & \RaggedRight Czech & \RaggedRight 00:00:53 & \RaggedRight 0:00:46 & \RaggedRight 0.242 \\
\RaggedRight fake \#106 & \RaggedRight \url{https://media2.ellinikahoaxes.gr/uploads/2024/06/447071239_3323218401145262_6755718274625284919_n.mp4} & \RaggedRight Greek & \RaggedRight 00:03:58 & \RaggedRight 0:03:15 & \RaggedRight 0.673 \\
\RaggedRight fake \#107 & \RaggedRight \url{https://twitter.com/facemagic_app/status/1415325224468369413
} & \RaggedRight English & \RaggedRight 00:00:12 & \RaggedRight 0:00:12 & \RaggedRight 0.542 \\
\RaggedRight fake \#108 & \RaggedRight \url{https://twitter.com/facemagic_app/status/1431995054181208066
} & \RaggedRight Hakha-Chin & \RaggedRight 00:00:12 & \RaggedRight 0:00:13 & \RaggedRight 0.0 \\
\RaggedRight fake \#109 & \RaggedRight \url{https://twitter.com/DBTagsen_Ofc/status/1312217715050979328
} & \RaggedRight English & \RaggedRight 00:00:16 & \RaggedRight 0:00:16 & \RaggedRight 0.0 \\
\RaggedRight fake \#110 & \RaggedRight \url{https://twitter.com/facemagic_app/status/1430907889305194496
} & \RaggedRight English & \RaggedRight 00:00:10 & \RaggedRight 0:00:11 & \RaggedRight 0.894 \\
\RaggedRight fake \#111 & \RaggedRight \url{https://twitter.com/SuperHeroAcade/status/1308803743421468677
} & \RaggedRight English & \RaggedRight 00:00:14 & \RaggedRight 0:00:14 & \RaggedRight 0.329 \\
\RaggedRight fake \#112 & \RaggedRight \url{https://twitter.com/DrDMemes1/status/1306246535160754177
} & \RaggedRight English & \RaggedRight 00:01:44 & \RaggedRight 0:01:37 & \RaggedRight 0.186 \\
\RaggedRight fake \#113 & \RaggedRight \url{https://twitter.com/facemagic_app/status/1406990302083272704
} & \RaggedRight Dhivehi & \RaggedRight 00:00:15 & \RaggedRight 0:00:14 & \RaggedRight 0.0 \\
\RaggedRight fake \#114 & \RaggedRight \url{https://www.tiktok.com/@jbviet/video/6995159085595331846
} & \RaggedRight Chinese-China & \RaggedRight 00:00:11 & \RaggedRight 0:00:11 & \RaggedRight 0.12 \\
\RaggedRight fake \#115 & \RaggedRight \url{https://twitter.com/AllanXia/status/1168521996168097793
} & \RaggedRight Persian & \RaggedRight 00:00:26 & \RaggedRight 0:00:35 & \RaggedRight 0.373 \\
\RaggedRight fake \#116 & \RaggedRight \url{https://twitter.com/facemagic_app/status/1414600441049010182
} & \RaggedRight English & \RaggedRight 00:00:21 & \RaggedRight 0:00:16 & \RaggedRight 0.247 \\
\RaggedRight fake \#117 & \RaggedRight \url{https://twitter.com/wanmafiq/status/1330742746869420032
} & \RaggedRight Spanish & \RaggedRight 00:00:09 & \RaggedRight 0:00:08 & \RaggedRight 0.0 \\
\RaggedRight fake \#118 & \RaggedRight \url{https://twitter.com/versl9/status/1281132787089711104
} & \RaggedRight Russian & \RaggedRight 00:00:42 & \RaggedRight 0:00:36 & \RaggedRight 0.446 \\
\RaggedRight fake \#119 & \RaggedRight \url{https://twitter.com/firejapan/status/1305133253225922560
} & \RaggedRight Breton & \RaggedRight 00:00:07 & \RaggedRight 0:00:08 & \RaggedRight 0.542 \\
\RaggedRight fake \#120 & \RaggedRight \url{https://twitter.com/stephaniemain2/status/1326246214078238720
} & \RaggedRight Chinese-Taiwan & \RaggedRight 00:00:11 & \RaggedRight 0:00:10 & \RaggedRight 0.542 \\
\RaggedRight fake \#121 & \RaggedRight \url{https://twitter.com/drfakenstein/status/1171182601563820038
} & \RaggedRight English & \RaggedRight 00:00:27 & \RaggedRight 0:00:25 & \RaggedRight 0.602 \\
\RaggedRight fake \#122 & \RaggedRight \url{https://twitter.com/rschu/status/1182372560236888065
} & \RaggedRight English & \RaggedRight 00:02:19 & \RaggedRight 0:02:08 & \RaggedRight 0.218 \\
\RaggedRight fake \#123 & \RaggedRight \url{https://www.facebook.com/MISTERKOCTWO/videos/949168468846018/
} & \RaggedRight French & \RaggedRight 00:01:00 & \RaggedRight 0:00:39 & \RaggedRight 0.0 \\
\RaggedRight fake \#124 & \RaggedRight \url{https://twitter.com/facemagic_app/status/1432357446450749446
} & \RaggedRight Spanish & \RaggedRight 00:00:12 & \RaggedRight 0:00:12 & \RaggedRight 0.0 \\
\RaggedRight fake \#125 & \RaggedRight \url{https://twitter.com/facemagic_app/status/1421848193671712770
} & \RaggedRight English & \RaggedRight 00:00:13 & \RaggedRight 0:00:12 & \RaggedRight 0.087 \\
\RaggedRight fake \#126 & \RaggedRight \url{https://twitter.com/CestCanteloup/status/1275394966022471685
} & \RaggedRight French & \RaggedRight 00:00:37 & \RaggedRight 0:00:33 & \RaggedRight 0.0 \\
\RaggedRight fake \#127 & \RaggedRight \url{https://twitter.com/lachoseparis/status/1181168724629475328
} & \RaggedRight English & \RaggedRight 00:00:32 & \RaggedRight 0:00:24 & \RaggedRight 0.054 \\
\RaggedRight fake \#128 & \RaggedRight \url{https://twitter.com/Startup_Nerd/status/1283695456992862208
} & \RaggedRight English & \RaggedRight 00:00:49 & \RaggedRight 0:00:38 & \RaggedRight 0.094 \\
\RaggedRight fake \#129 & \RaggedRight \url{https://twitter.com/MikaelThalen/status/1311696656061009920
} & \RaggedRight English & \RaggedRight 00:00:49 & \RaggedRight 0:00:37 & \RaggedRight 0.166 \\
\end{longtable}
\twocolumn

\end{document}